\documentclass{article}

    \usepackage[preprint]{neurips_2021}

\usepackage[utf8]{inputenc} %
\usepackage[T1]{fontenc}    %
\usepackage{hyperref}       %
\usepackage{url}            %
\usepackage{booktabs}       %
\usepackage{amsfonts}       %
\usepackage{nicefrac}       %
\usepackage{microtype}      %
\usepackage{xcolor}         %
\newcommand{\matr}[1]{\mathbf{#1}}
\usepackage{graphicx}
\usepackage{booktabs}
\usepackage{makecell}
\usepackage{caption}
\usepackage{subcaption}
\usepackage{float}
\usepackage{textcomp,gensymb}
\usepackage{enumitem}
\usepackage[olditem]{paralist}

\title{Grasping the Arrow of Time from the Singularity: Decoding Micromotion in Low-dimensional Latent Spaces from StyleGAN}

\author{%
  Qiucheng Wu$^1$\thanks{Equal Contribution.} , Yifan Jiang$^{2*}$, Junru Wu$^{3*}$, Kai Wang$^5$, Gong Zhang$^5$, \\\textbf{Humphrey Shi$^{4,5,6}$, Zhangyang Wang$^2$, Shiyu Chang$^1$}\\
  $^1$University of California, Santa Barbara, $^2$The University of Texas at Austin, \\$^3$Texas A\&M University, $^4$UIUC, $^5$University of Oregon, $^6$Picsart AI Research
}

\begin{document}

\maketitle

\begin{abstract}
The disentanglement of StyleGAN latent space has paved the way for realistic and controllable image editing, but does StyleGAN know anything about temporal motion, as it was only trained on static images? To study the motion features in the latent space of StyleGAN, in this paper, we hypothesize and demonstrate that a series of meaningful, natural, and versatile small, local movements (referred to as ``micromotion'', such as expression, head movement, and aging effect) can be represented in low-rank spaces extracted from the latent space of a conventionally pre-trained StyleGAN-v2 model for face generation, with the guidance of proper ``anchors'' in the form of either short text or video clips. 
Starting from one target face image, with the editing direction decoded from the low-rank space, its micromotion features can be represented as simple as an affine transformation over its latent feature. Perhaps more surprisingly, such micromotion subspace, even learned from just single target face, can be painlessly transferred to other unseen face images, even those from vastly different domains (such as oil painting, cartoon, and sculpture faces). It demonstrates that the local feature geometry corresponding to one type of micromotion is aligned across different face subjects, and hence that StyleGAN-v2 is indeed ``secretly'' aware of the subject-disentangled feature variations caused by that micromotion. We present various successful examples of applying our low-dimensional micromotion subspace technique to directly and effortlessly manipulate faces, showing high robustness, low computational overhead, and impressive domain transferability. Our codes are available at \url{https://github.com/wuqiuche/micromotion-StyleGAN}.
\end{abstract}

\begin{figure*}
\centering

\begin{subfigure}{.18\textwidth}
  \centering
  \includegraphics[width=\textwidth]{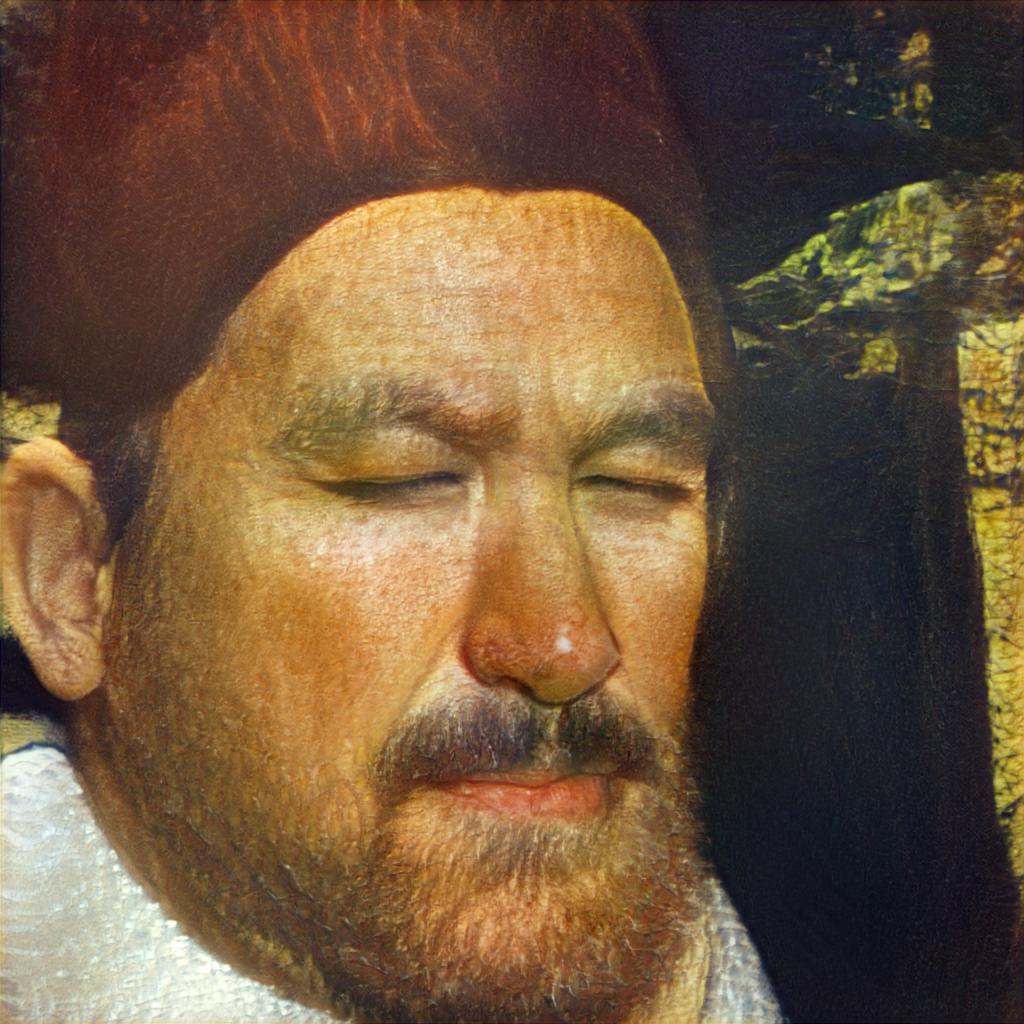}
\end{subfigure}
\begin{subfigure}{.18\textwidth}
  \centering
  \includegraphics[width=\textwidth]{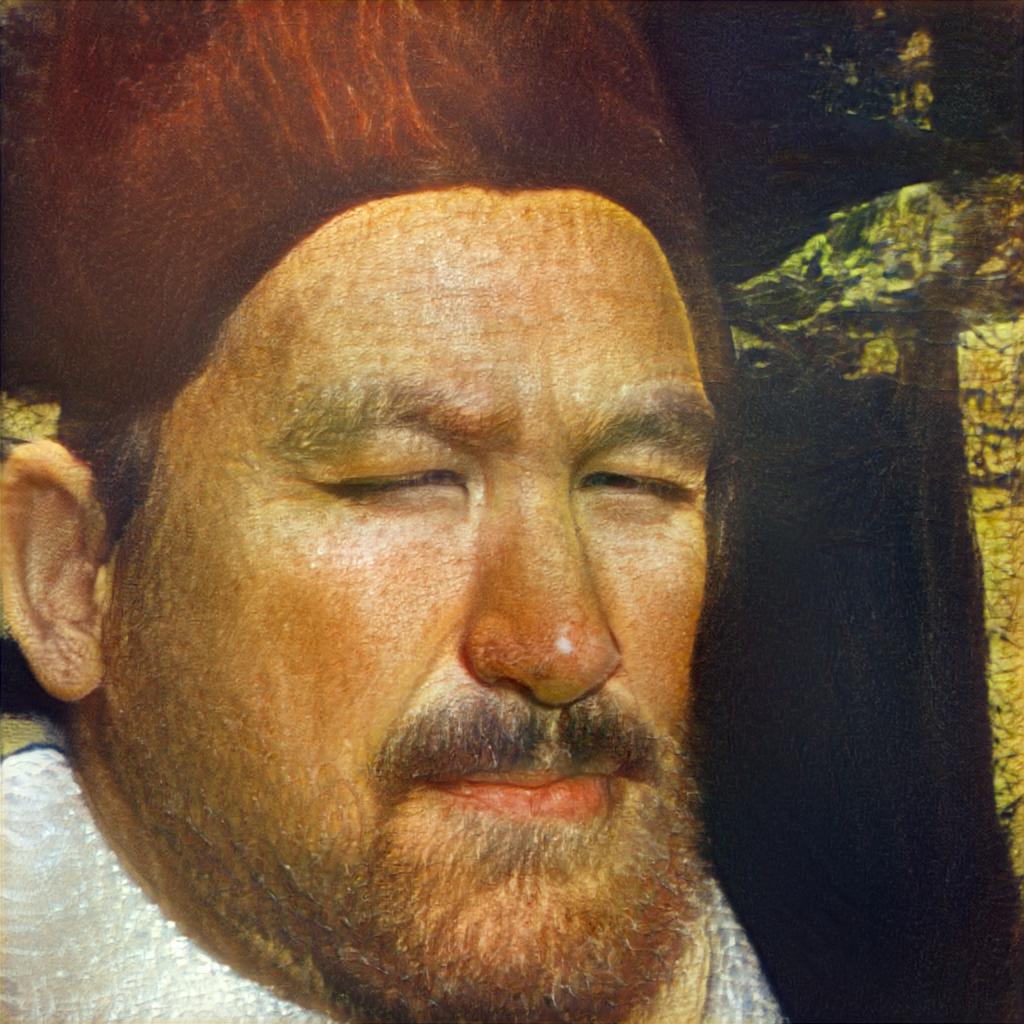}
\end{subfigure}
\begin{subfigure}{.18\textwidth}
  \centering
  \includegraphics[width=\textwidth]{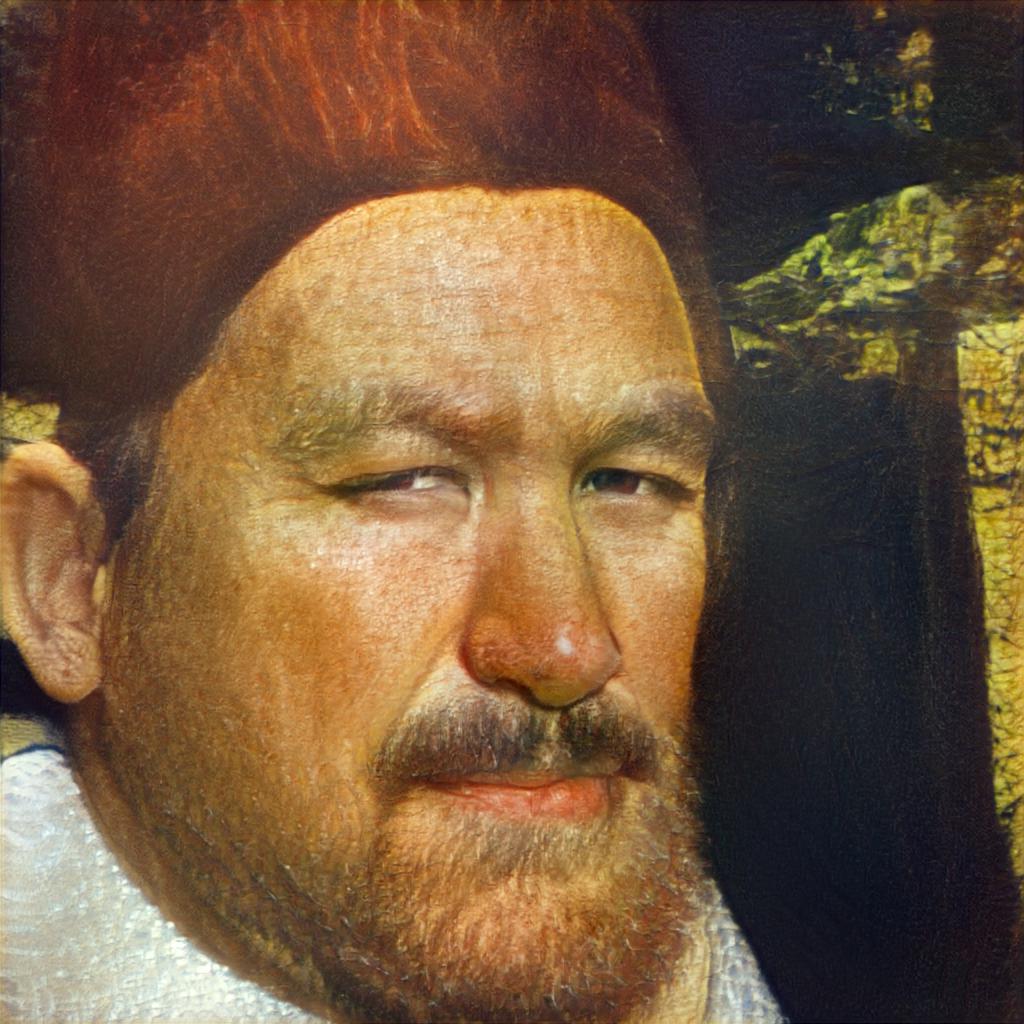}
\end{subfigure}
\begin{subfigure}{.18\textwidth}
  \centering
  \includegraphics[width=\textwidth]{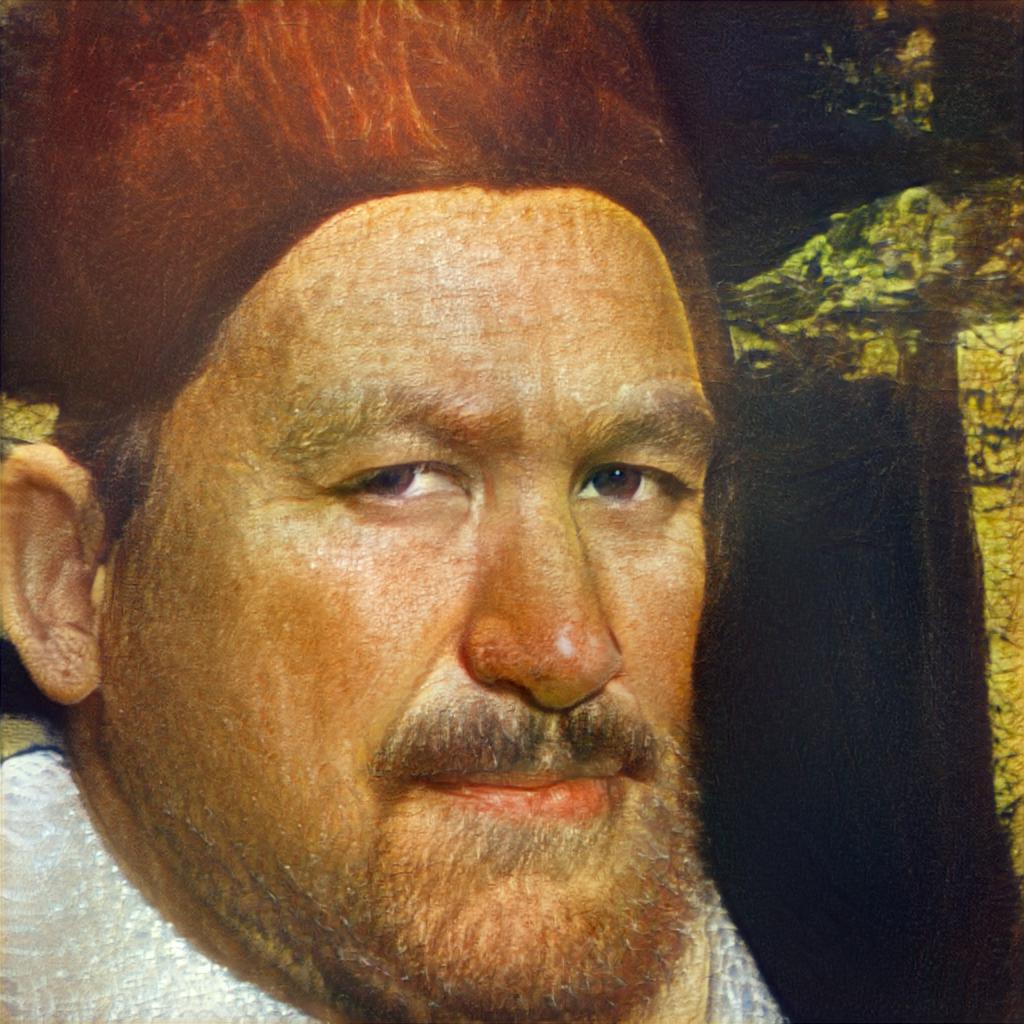}
\end{subfigure}
\begin{subfigure}{.18\textwidth}
  \centering
  \includegraphics[width=\textwidth]{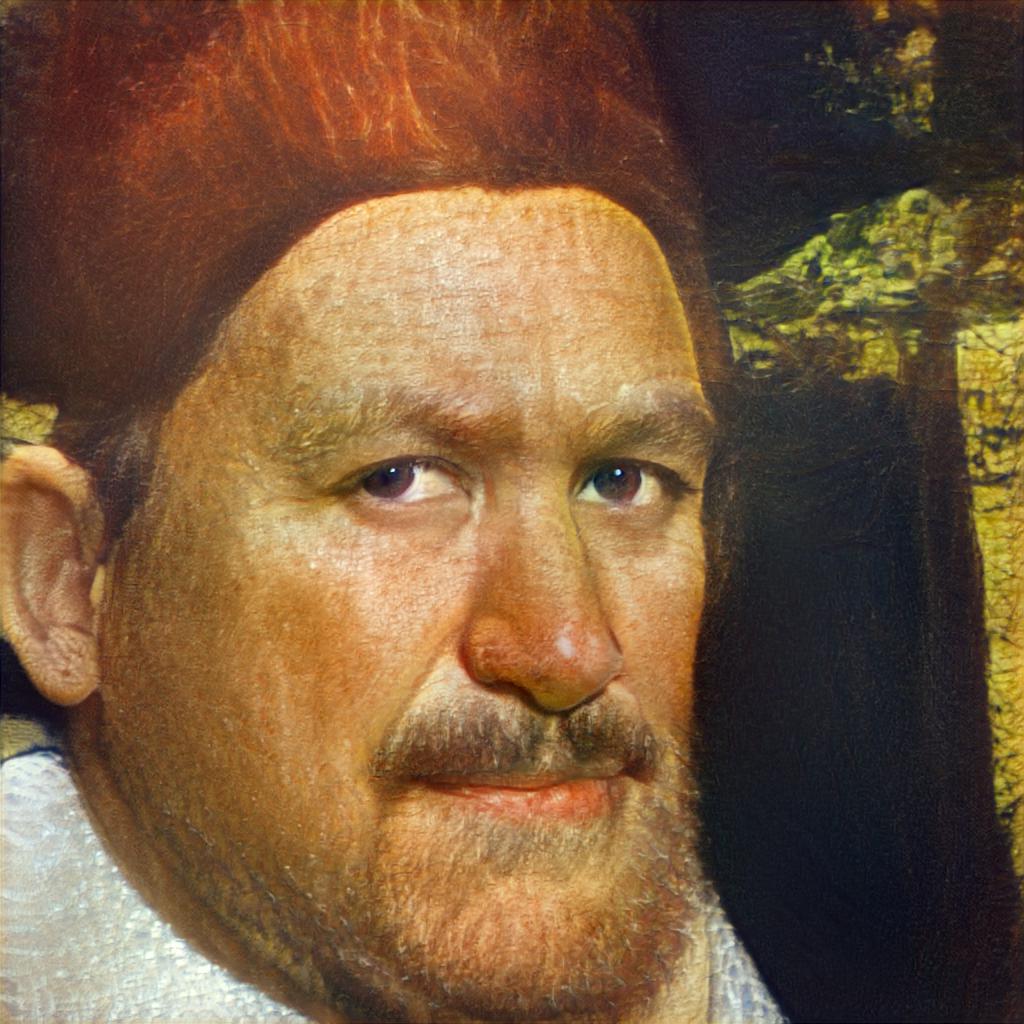}
\end{subfigure}

\begin{subfigure}{.18\textwidth}
  \centering
  \includegraphics[width=\textwidth]{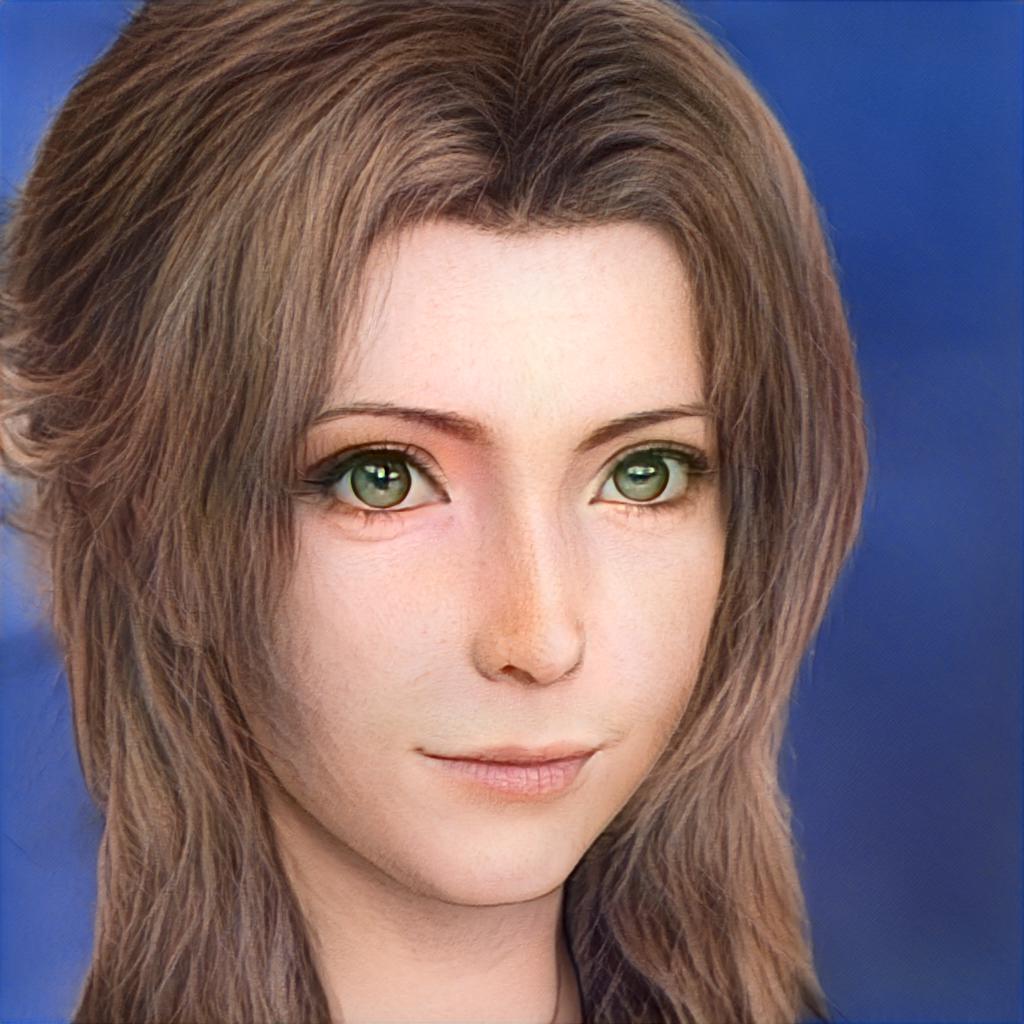}
\end{subfigure}
\begin{subfigure}{.18\textwidth}
  \centering
  \includegraphics[width=\textwidth]{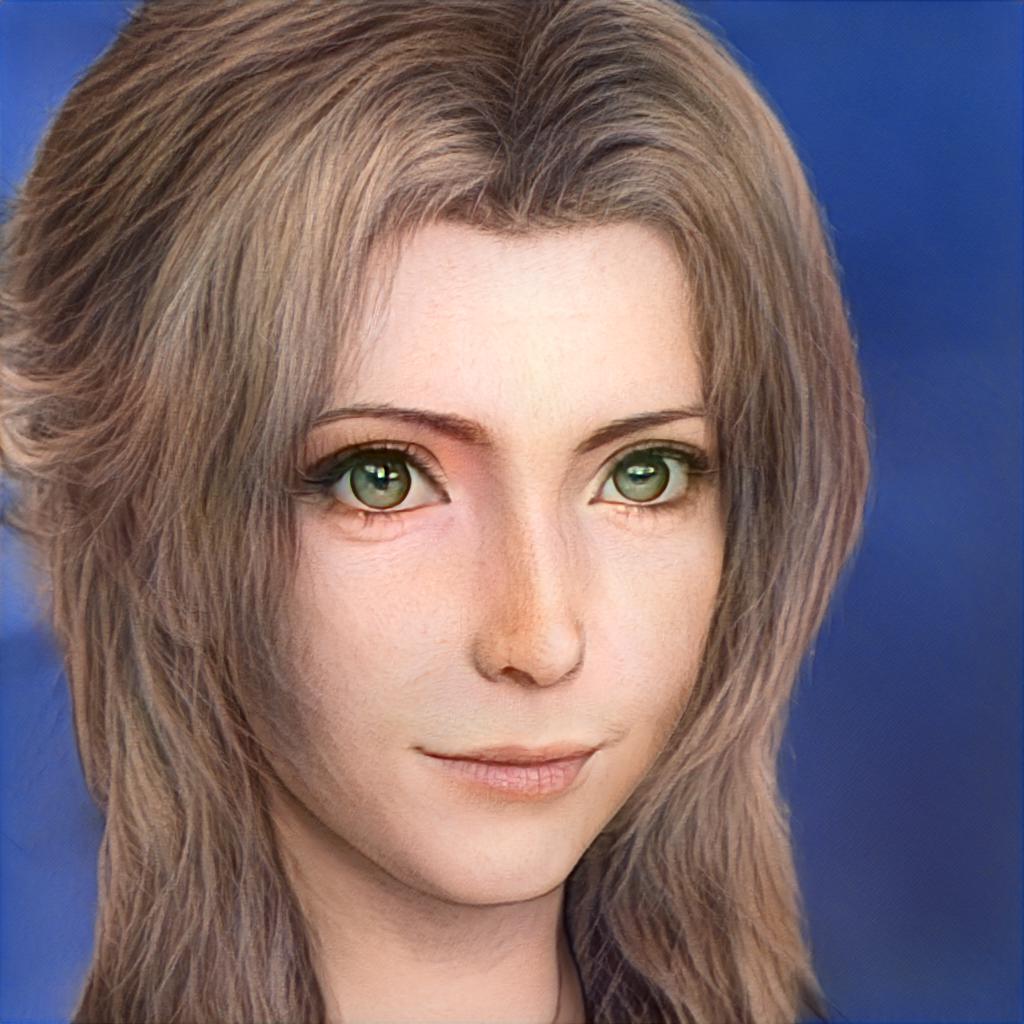}
\end{subfigure}
\begin{subfigure}{.18\textwidth}
  \centering
  \includegraphics[width=\textwidth]{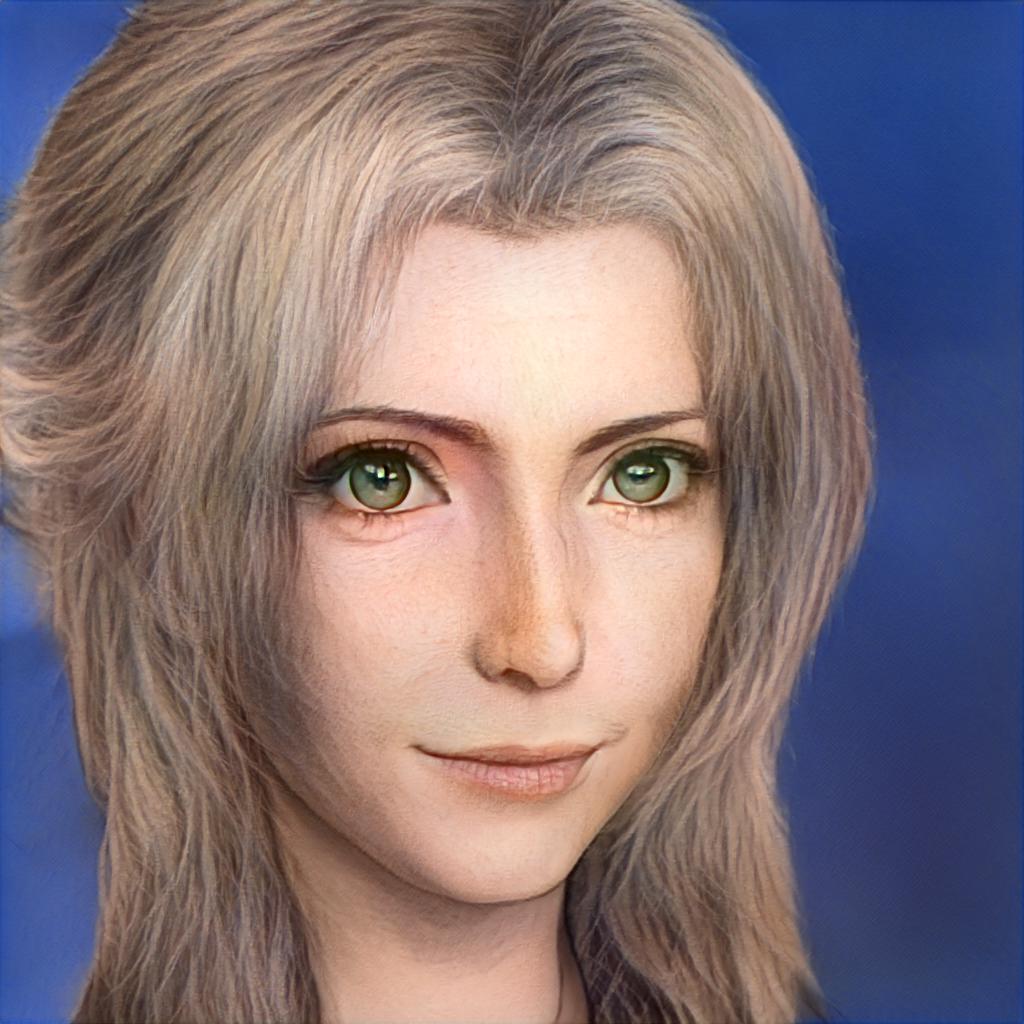}
\end{subfigure}
\begin{subfigure}{.18\textwidth}
  \centering
  \includegraphics[width=\textwidth]{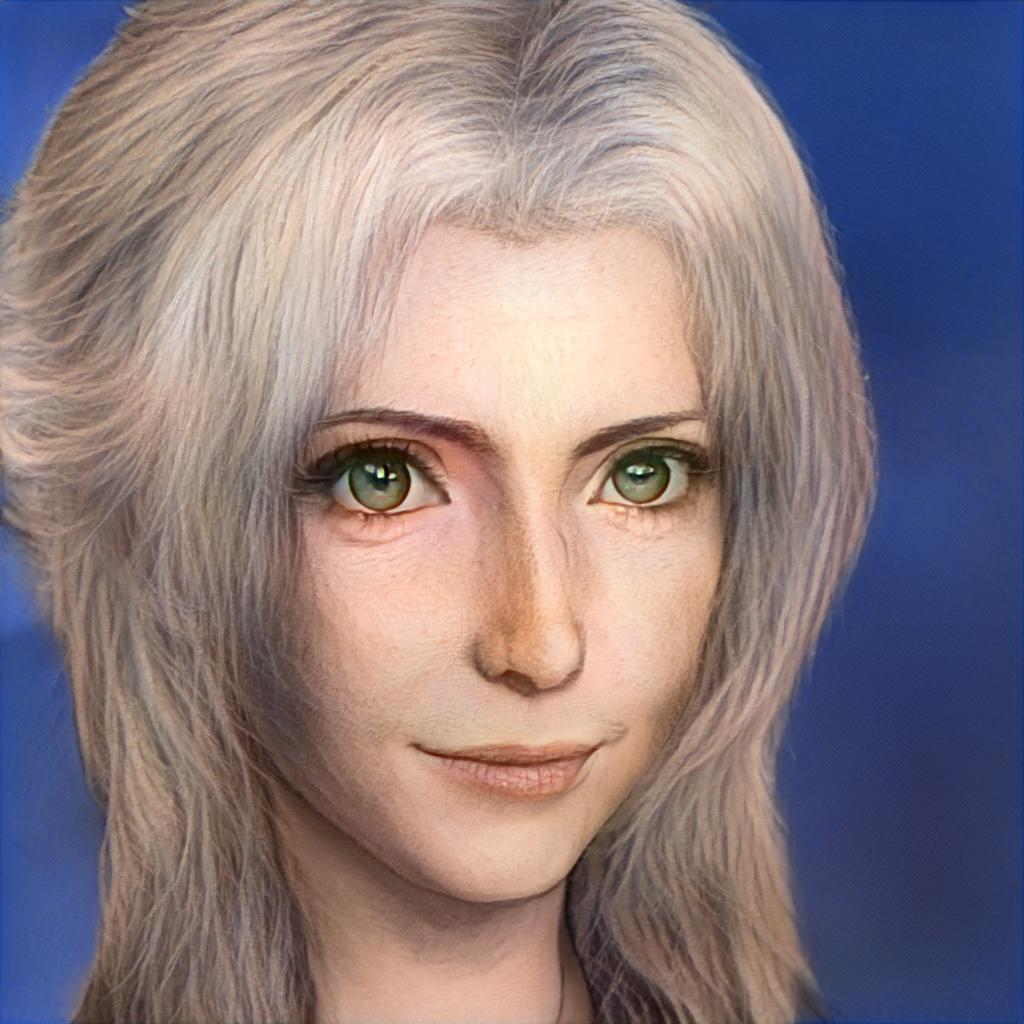}
\end{subfigure}
\begin{subfigure}{.18\textwidth}
  \centering
  \includegraphics[width=\textwidth]{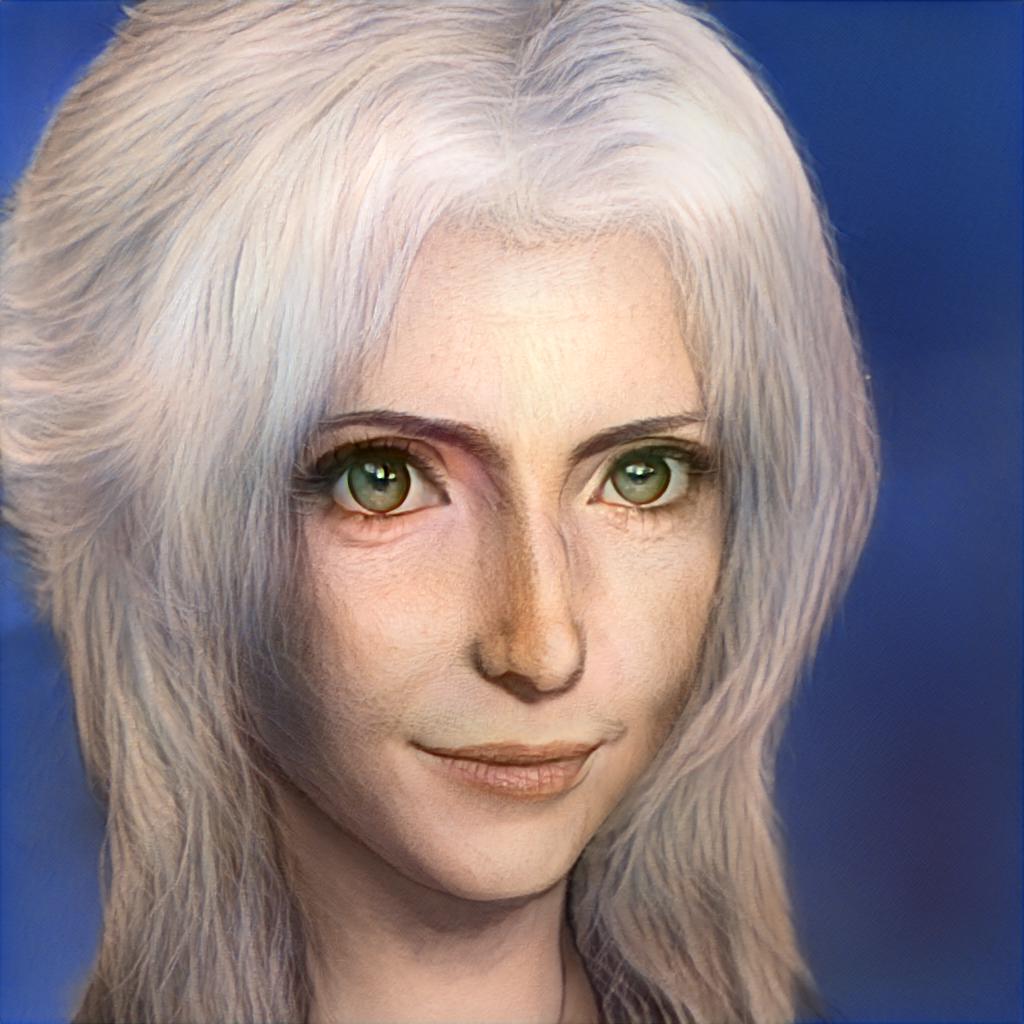}
\end{subfigure}

\begin{subfigure}{.18\textwidth}
  \centering
  \includegraphics[width=\textwidth]{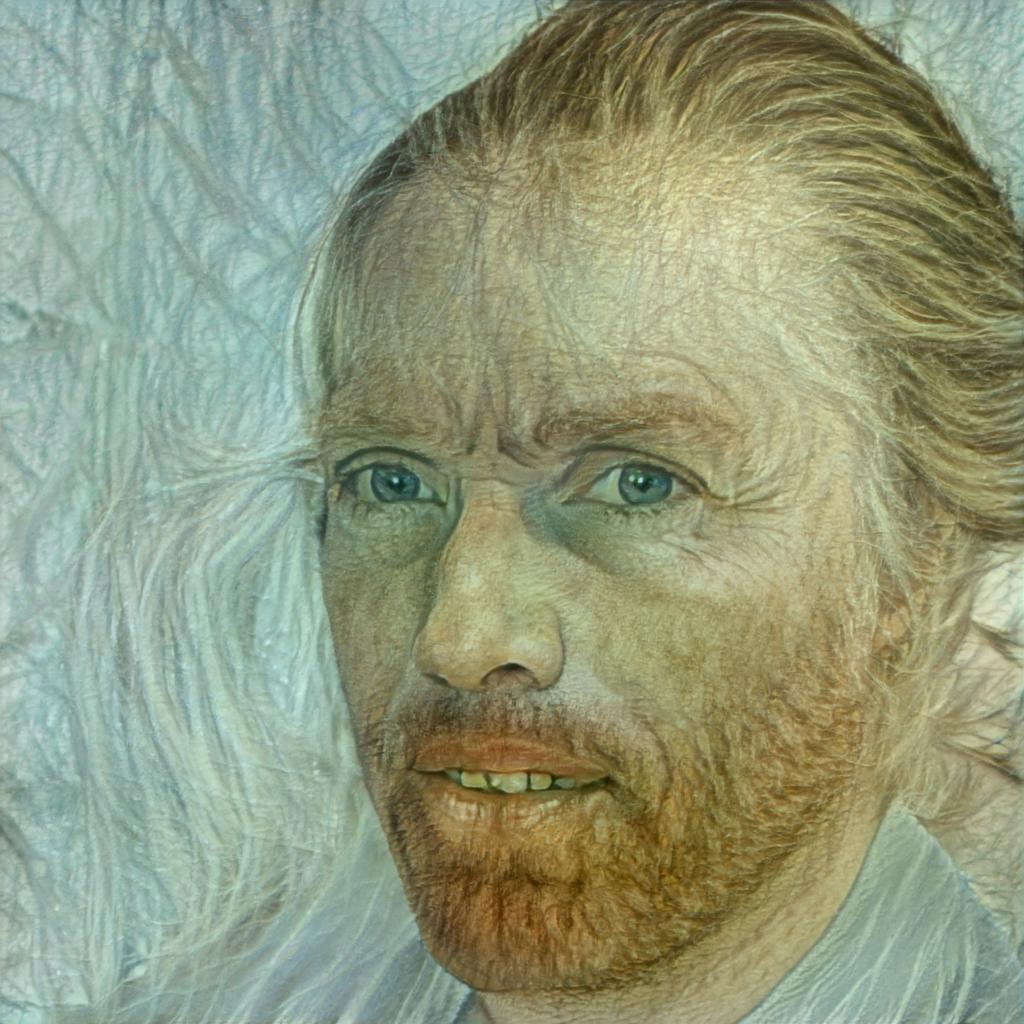}
\end{subfigure}
\begin{subfigure}{.18\textwidth}
  \centering
  \includegraphics[width=\textwidth]{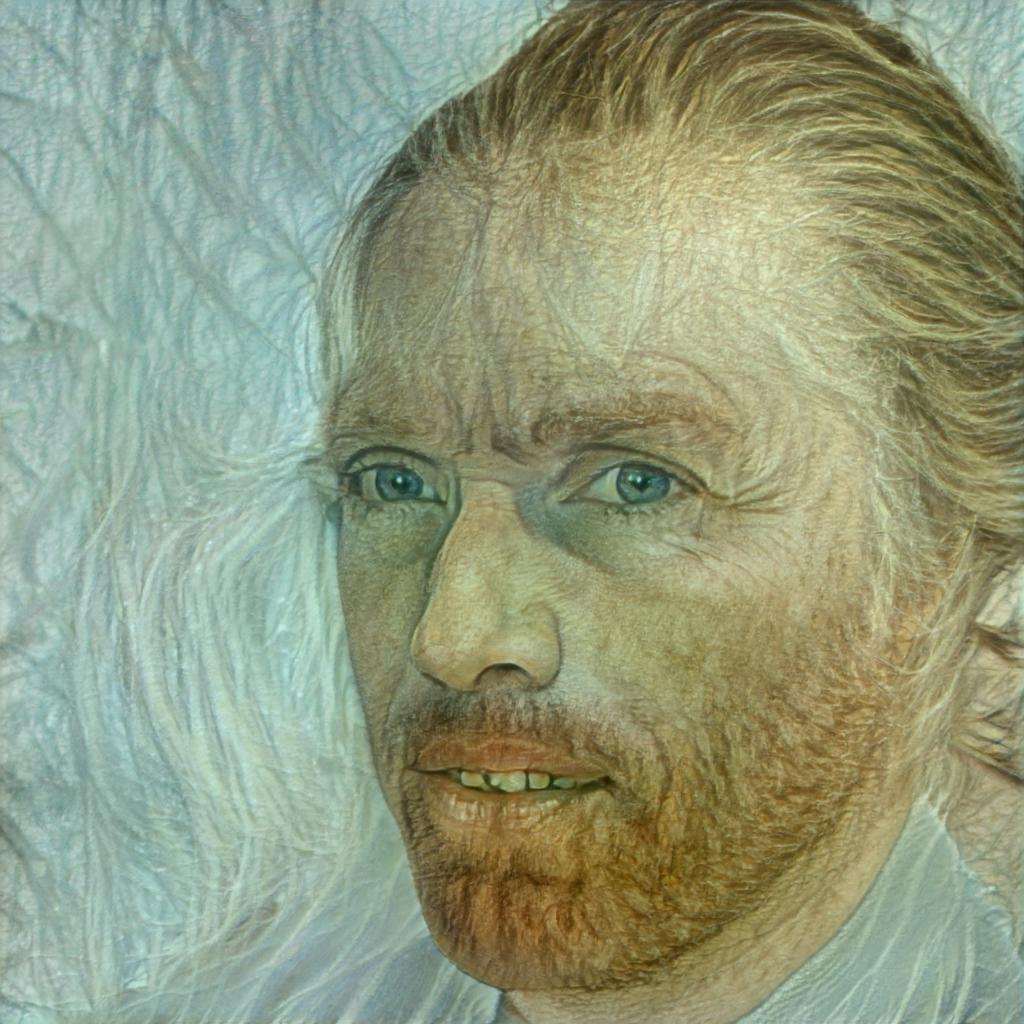}
\end{subfigure}
\begin{subfigure}{.18\textwidth}
  \centering
  \includegraphics[width=\textwidth]{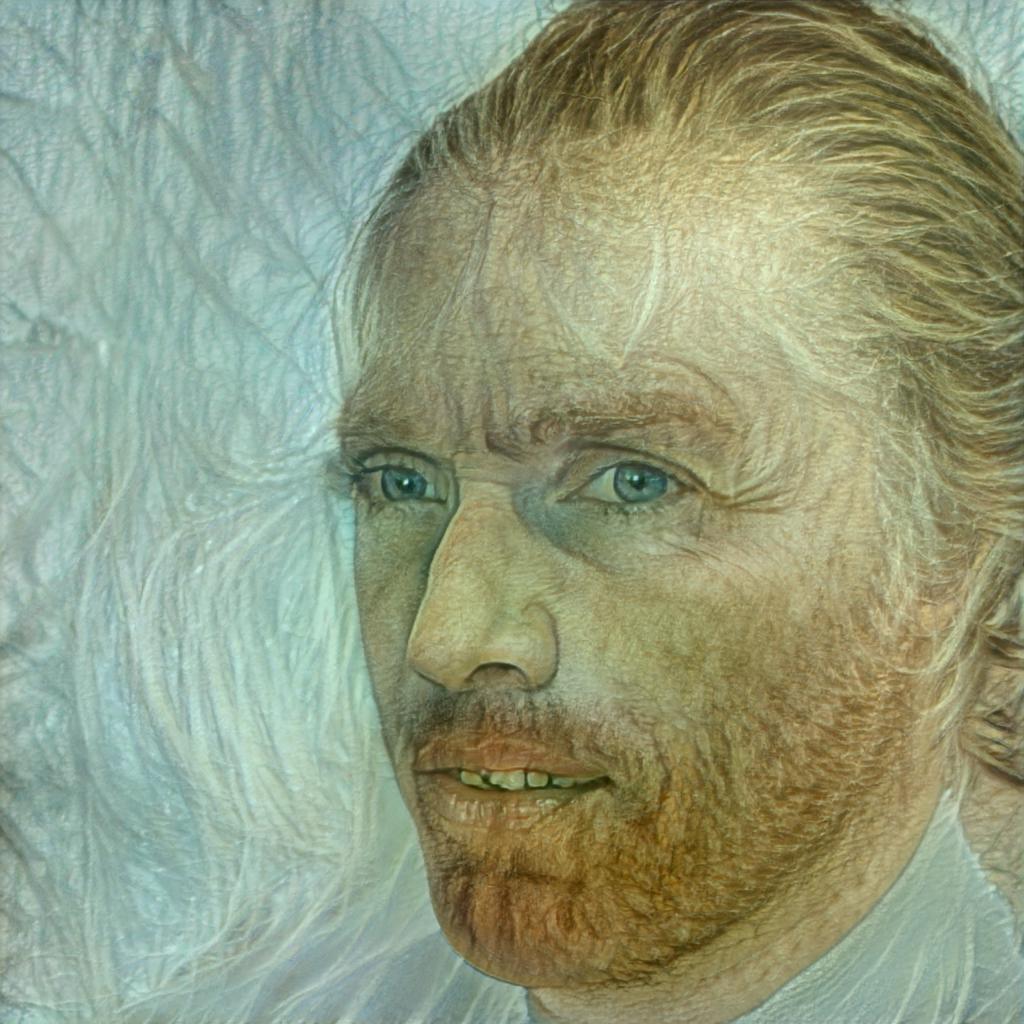}
\end{subfigure}
\begin{subfigure}{.18\textwidth}
  \centering
  \includegraphics[width=\textwidth]{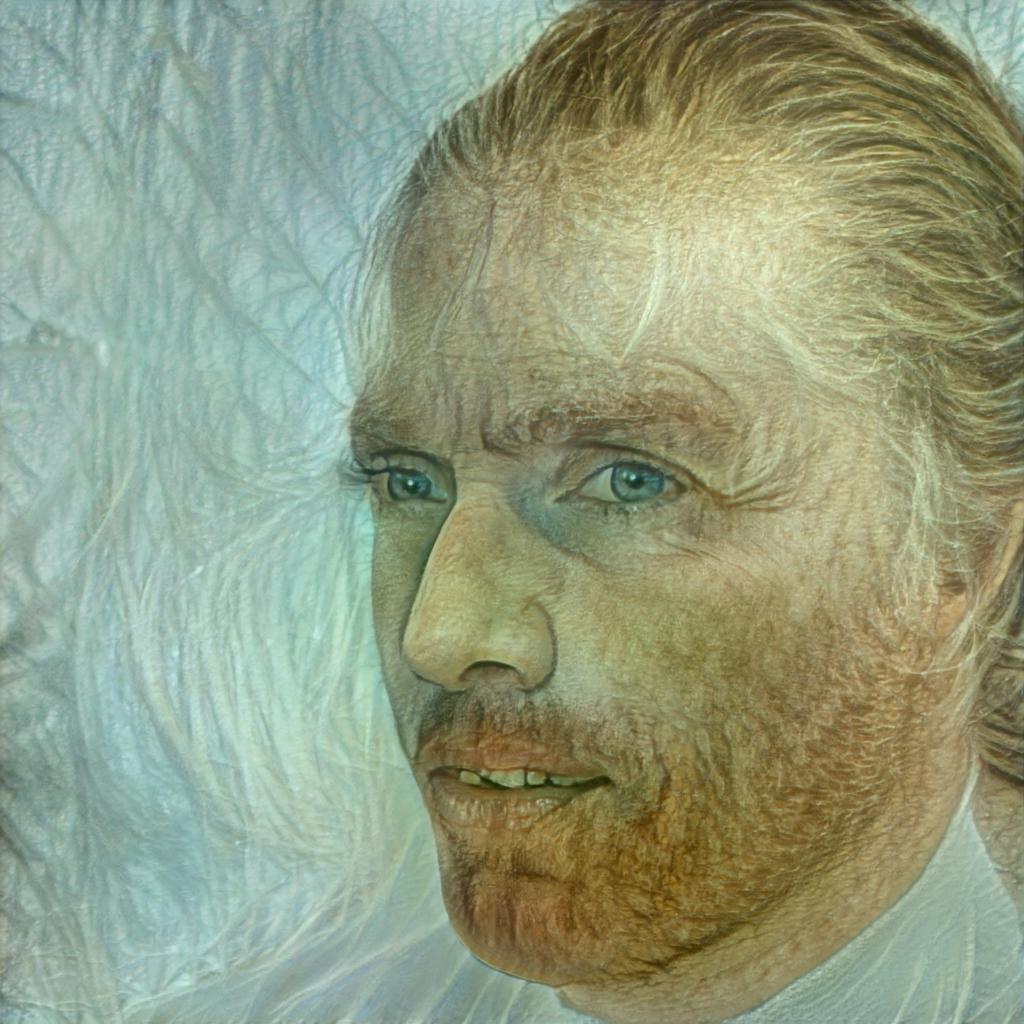}
\end{subfigure}
\begin{subfigure}{.18\textwidth}
  \centering
  \includegraphics[width=\textwidth]{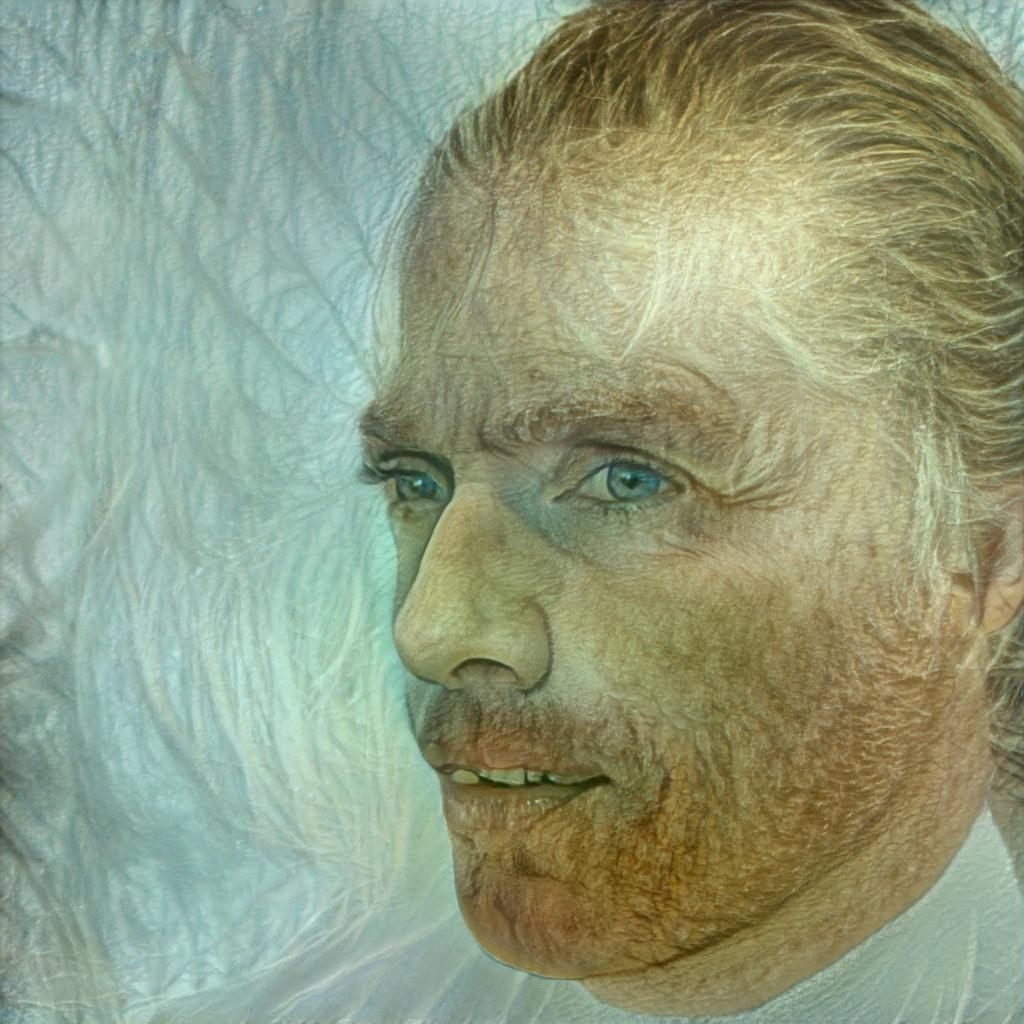}
\end{subfigure}
  \caption{\textbf{Representative examples created by the proposed method.} The original images are edited using a simple linear scaling with the discovered universal editing directions on various transformations. These three rows correspond to \textit{eye-opening, aging, and head rotation}. }
  \vspace{-1em}
  \label{fig:teaser}
\end{figure*}

\section{Introduction}
In recent years, the StyleGAN and its variants \cite{brock2018large,karras2017progressive,karras2021alias,karras2019style,karras2020analyzing,sauer2022stylegan} have achieved state-of-the-art performance in controllable image synthesis.  It has been shown \cite{karras2019style} that by latent feature manipulations and interpolations, the Style-based GANs can generate a variety of intriguing images,  which made them widely applied to many downstream applications such as image editing~\cite{alaluf2021only,gu2020image,park2020swapping,suzuki2018spatially} and video generation~\cite{chu2020learning,fox2021stylevideogan,skorokhodov2021stylegan,zhang2021facial}.  Specifically, the early attempts focus on searching in StyleGAN latent space to find edit directions corresponding to semantic meaningful manipulations \cite{hou2022guidedstyle,shen2020interpreting,tewari2020pie,tewari2020stylerig,wu2021stylespace}.  More recently, a few approaches try to simplify the searching process and enable more fine-grained controls using text-driven latent manipulation \cite{abdal2021clip2stylegan,gal2021stylegan,patashnik2021styleclip}.   Furthermore, reference images/videos have also been considered \cite{chefer2021image,kim2021exploiting,lewis2021vogue} to pinpoint the generation process.

Given these phenomenal results, many try to understand the foundations of the synthesis process and the properties of the latent space of StyleGAN. People have found that the learned latent space of StyleGAN can be smooth, interpretable, and partially disentangled in various dimensions \cite{abdal2019image2stylegan,abdal2020image2stylegan++,wu2021stylespace,zhu2020improved}. These pleasing properties allow images to be editable in the latent space and the interpolation of latent space vectors to yield a smooth transition. However, plenty of these previous probing analyses are mainly on a per-example basis, and to change various input images to the same style (\emph{e.g.,} make different people smile), one needs to find such edits differently and individually. Therefore, a more universal, identity-agnostic edit is highly desirable for the ease of editing controls. A few works \cite{abdal2019image2stylegan,patashnik2021styleclip,shen2020interpreting} started to explore this sample-agnostic editing method, while these works usually require additional models with a large number of samples and fine-tuning, which introduces new challenges to high-quality editing. Nevertheless, whether there exist ubiquitous and sample-agnostic feature transformations in the latent space that can be easily found to manipulate any inputs in the same fashion remains an interesting yet unknown question.

To answer the question, in this paper, we propose in-depth investigations on the StyleGAN-v2's latent space trained on face generations. In particular, we hypothesize that from the StyleGAN's high dimensional latent space, a low-rank feature space can be extracted where universal editing directions can be reconstructed for various facial style transformations including changes in expressions/emotions, heads movements, and aging effects. In other words, for any given input, linear scaling along the same found direction will make the image change its style in a smooth fashion.  Furthermore, to find such a directional vector we leverage the guidance of proper “anchors” in the form of either short texts or a reference video clip and show the directional vector can be efficiently found via simple subtractions using a robustly learned linear subspace projection.  Surprisingly, such latent subspace can be extracted using only a single query image, and then the resulting editing direction can be used to any unseen face image, even for those from vastly different domains including oil painting, cartoon, sculpture, \emph{etc}.  Figure \ref{fig:teaser} shows the generated images for multiple style transformations and face types.  The contributions of our paper are three-fold:

\begin{itemize}[leftmargin=1.5em]\vspace{-0.5em}
\item Differing from former per-sample-based editing and analyses, we conduct the first pilot study to understand the properties of StyleGAN's latent space from a global and universal viewpoint, using ``micromotions'' as the subject.
\item We demonstrate that by using text/video-based anchors, low-dimensional micromotion subspace along with universal editing directions can be consistently discovered using the same robust subspace projection technique for a large range of micromotion-style facial transformations.  
\item We show the editing direction can be found using a single query face input and then directly applied to other faces, even from vastly different domains (\emph{e.g.}, oil painting, cartoon, and sculpture faces), in an easily controllable way as simple as linear scaling along the discovered subspace.
\end{itemize}

\section{Related Works}
\subsection{StyleGAN: Models and Characteristics}
The StyleGAN ~\cite{karras2021alias,karras2019style,karras2020analyzing} is a style-based generator architecture targeting on image synthesis task. With the help of a mapping network and affine transformation to render abstract style information, the StyleGAN is able to control the image synthesis in a scale-specific fashion. Particularly, by augmenting the learned feature space and hierarchically feeding latent codes at each layer of the generator architecture, the StyleGAN has demonstrated surprising image synthesis performance with controls from coarse properties to fine-grained characteristics~\cite{karras2019style}. Also, when trained on a high-resolution facial dataset (\emph{e.g.,} FFHQ~\cite{karras2019style}), the StyleGAN is able to generate high-quality human faces with good fidelity.

\subsection{StyleGAN-based Editing}
Leveraging the expressive and disentangled latent space by StyleGAN, recent studies consider interpolating and mixing the latent style codes to achieve specific attribute editing without impairing other attributes (e.g. person identity). \cite{hou2022guidedstyle,shen2020interpreting,tewari2020pie,tewari2020stylerig,wu2021stylespace} focus on searching latent space to find latent codes corresponding to global meaningful manipulations, while \cite{chong2021retrieve} utilizes semantic segmentation maps to locate and mix certain positions of style codes to achieve editing goals.

To achieve zero-shot and open-vocabulary editing, latest works set their sights on using pretrained multi-modality models as guidance. With the aligned image-text representation learned by CLIP, a few works~\cite{wei2021hairclip,patashnik2021styleclip} use text to extract the latent edit directions with textual defined semantic meanings for separate input images. These works focus on extracting latent directions using contrastive CLIP loss to conduct image manipulation tasks such as face editing~\cite{patashnik2021styleclip,wei2021hairclip}, cars editing~\cite{abdal2021clip2stylegan}. On the other hand, rather than editing the latent code, in observance of the smoothness of the StyleGAN feature space, Gal \textit{et al.}~\cite{gal2021stylegan} focus on fine-tuning the latent domain of the generator to transfer the feature domain. As the result of domain adaptation, the fine-tuned generator synthesizes images alleviated from the original domain. Besides, a few recent works manipulate the images with visual guidance ~\cite{lewis2021vogue,kim2021exploiting}. In these works, image editing is done by inverting the referential images into corresponding latent codes, and interpolating the latent codes to generate mixed-style images. However, most of the previous works focus on a per-example basis, with only a few exceptions ~\cite{abdal2019image2stylegan,patashnik2021styleclip,shen2020interpreting}. Therefore, a universal and sample-agnostic feature transformation in the latent space is highly desirable.

\subsection{Feature Disentanglement in StyleGAN Latent Space}
The natural and smooth performance of StyleGAN-based image editing largely credits to its disentangled feature space. Many works ~\cite{abdal2019image2stylegan,abdal2020image2stylegan++,wu2021stylespace,zhu2021barbershop} study on the disentangle properties of StyleGAN, comparing and contrasting on its various latent space including $\mathcal{Z}$ space, $\mathcal{W}$ space, and $\mathcal{S}$ space. These studies have revealed that the latent space is disentangled in different degrees, and therefore is suitable in various tasks. Due to the disentangle property in $\mathcal{W}$ and $\mathcal{S}$ spaces, large number of works ~\cite{chefer2021image,patashnik2021styleclip,roich2021pivotal} edit images on the $\mathcal{W}$ and $\mathcal{S}$ spaces, and the task of image inversions with StyleGAN encoders ~\cite{alaluf2021restyle,tov2021designing,richardson2021encoding} are mainly conducted in $\mathcal{W^+}$ space, an augmented latent space from $\mathcal{W}$ with more degree of freedom. To leverage the powerful image inversions techniques along with disentanglement properties in latent space, this work focuses on the $\mathcal{W^+}$ latent space, where we further study the existence of locally low-rank micromotion subspace.

\section{Method}
In this section, we first present the problem of decoding micromotion in a pre-trained StyleGAN latent space, and we define the notations involved in this paper.
We then articulate the low-rank micromotion subspace hypothesis in Sec. \ref{keyHypothesis}, proposing that the locally low-dimensional geometry corresponding to one type of micromotion is consistently aligned across different face subjects, which serves as the key to decode universal micromotion from even a single identity.
Finally, based on the hypothesis, we demonstrate a simple workflow to decode micromotions and seamlessly apply them to various in-domain and out-domain identities, incurring clear desired facial micromotions.

\subsection{Problem Setting}
Micromotions are reflected as smooth transitions in continuous video frames. In a general facial-style micromotion synthesis problem, given an arbitrary input image $I_0$ and a desired micromotion (\emph{e.g.} smile), the goal is to design an identity-agnostic workflow to synthesize temporal frames \{$I_1$, $I_2$, …, $I_t$\}, which constitute a consecutive video with the desired micromotion.

Synthesizing images with StyleGAN requires finding proper latent codes in its feature space. We use $G$ and $E$ to denote the pre-trained StyleGAN synthesis network and StyleGAN encoder respectively. Given a latent code $\matr{V} \in \mathcal{W}^+$, the pre-trained generator $G$ maps it to the image space by $I = G(\matr{V})$. Inversely, the encoder maps the image $I$ back to the latent space $\mathcal{W}^+$, or $\hat{\matr{V}} = E(I)$. Leveraging the StyleGAN latent space, finding consecutive video frames turns out to be a task of finding a series of latent codes \{$\matr{V}_1$, $\matr{V}_2$, …, $\matr{V}_t$\} corresponding to the micromotion.

\subsection{Key Hypothesis: The Low-rank Micromotion Subspace}\label{keyHypothesis}
\begin{figure}
\vspace{-1em}
    \centering
    \includegraphics[width=0.98\textwidth]{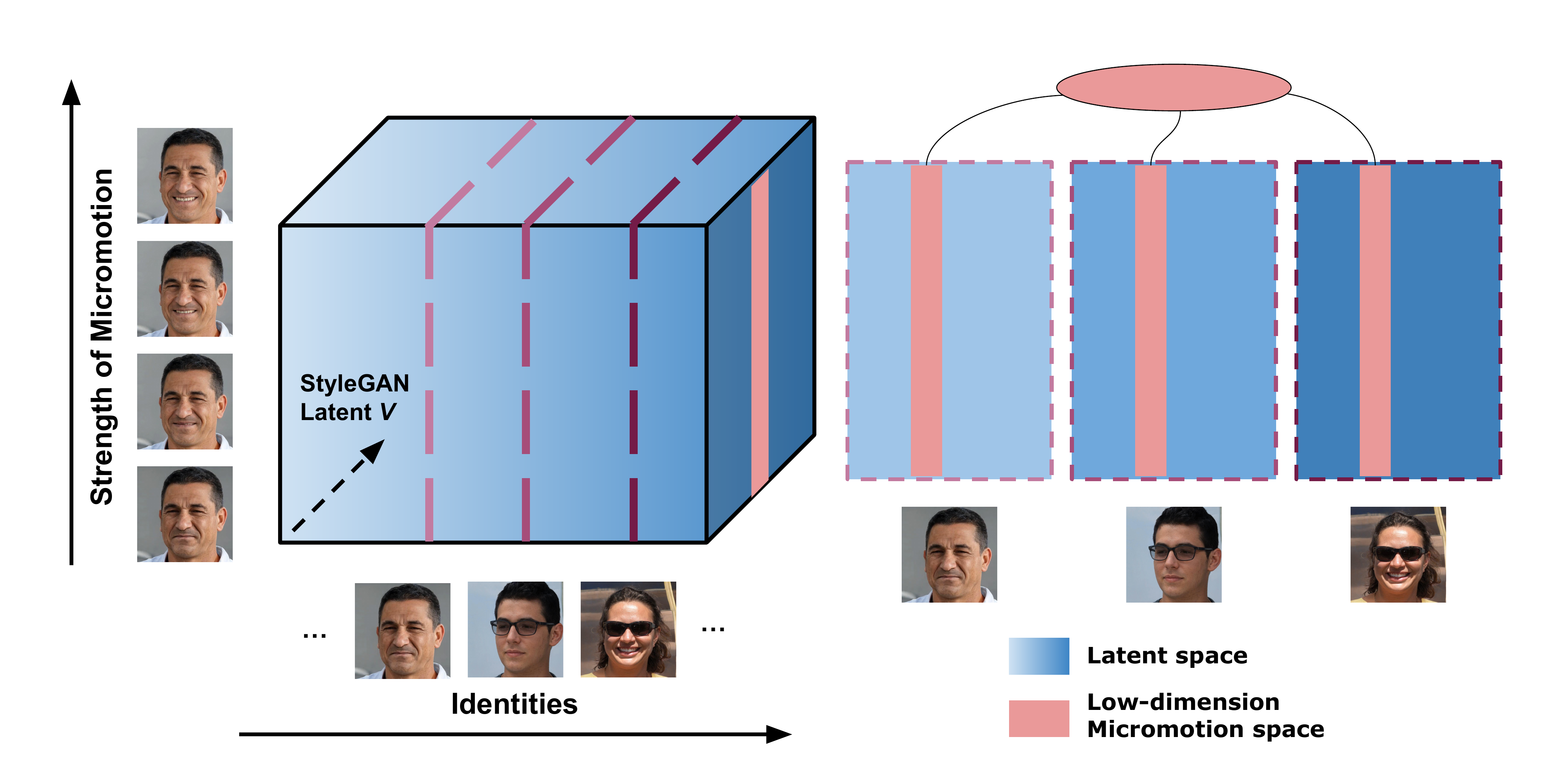}
    \vspace{-1em}
    \caption{\textbf{A tensor illustration of our low-rank micromotion subspace hypothesis.} In the StyleGAN latent space, we hypothesize the same type of micromotion, \textit{at different quantitative levels but for the same identity}, can be approximated by a low-rank subspace. We further hypothesize that subspaces \textit{for the same type of micromotion found at different identities} are extremely similar to each other, and can hence be transferred across identities.}
    \label{fig:key_hypothesis}
\end{figure}

To generate semantically meaningful and correct micromotions using StyleGAN, the key objective is to find proper latent code series in its feature space. We hypothesize that those latent codes can be decoded by a low-rank micromotion subspace. Specifically, we articulate the key hypothesis in this work, stated as: \textit{The versatile facial style micromotions can be represented as low-rank subspaces within the StyleGAN latent space, and such subspaces are subject-agnostic.} 

To give a concrete illustration of the hypothesis, we plot a tensor-view illustration of a micromotion subspace, smile, in Figure~\ref{fig:key_hypothesis}. The horizontal axis encodes the different face identities, and each perpendicular slice of the vertical plane represents all variations embedded in the StyleGAN latent space for a specific identity. We use the vertical axis to indicate the quantitative strength for a micromotion (\emph{e.g.}, smile from mild to wild). Given a sampled set of images in which a subject face changes from the beginning (e.g., neutral) to the terminal state of a micromotion, each image can be synthesized using a latent code $\matr{V}$. Aligning these latent codes for one single subject formulates a \textit{micromotion matrix} with dimension $V\times M$, where $V$ is the dimension of the latent codes and $M$ is the total number of images. Eventually, different subjects could all formulate their micromotion matrices in the same way, yielding a \textit{micromotion tensor}, with dimension $P\times V\times M$ assuming a total of $P$ identities. Our hypothesis is then stated in two folds:
\begin{itemize}[leftmargin=2.0em]
    \item Each subject's micromotion matrix can be approximated by a simple linear ``micromotion subspace'' and it is inherently low-rank. Representing micromotion ``strengths'' can be reduced to linearly scaling along the subspace.
    \item The micromotion subspaces found at different subjects are substantially similar and even mutually transferable. In other words, different subjects (approximately) share the common micromotion subspace. That implies the existence of universal edit direction for one specific micromotion type, regardless of subject identities. 
\end{itemize}

If the hypothesis can be proven true, it would be immediately appealing for both understanding the latent space of StyleGAN, and for practical applications in image and video manipulations. First, micromotion can be represented in low-dimensional disentangled spaces, and the dynamic edit direction can be reconstructed once the space is anchored. Second, when the low-dimensional space is found, it can immediately be applied to multiple other identities with extremely low overhead, and is highly controllable through interpolation and extrapolation by as simple as linear scaling. 

\subsection{Our Workflow}
\begin{figure}
\centering
\includegraphics[width=0.95\textwidth]{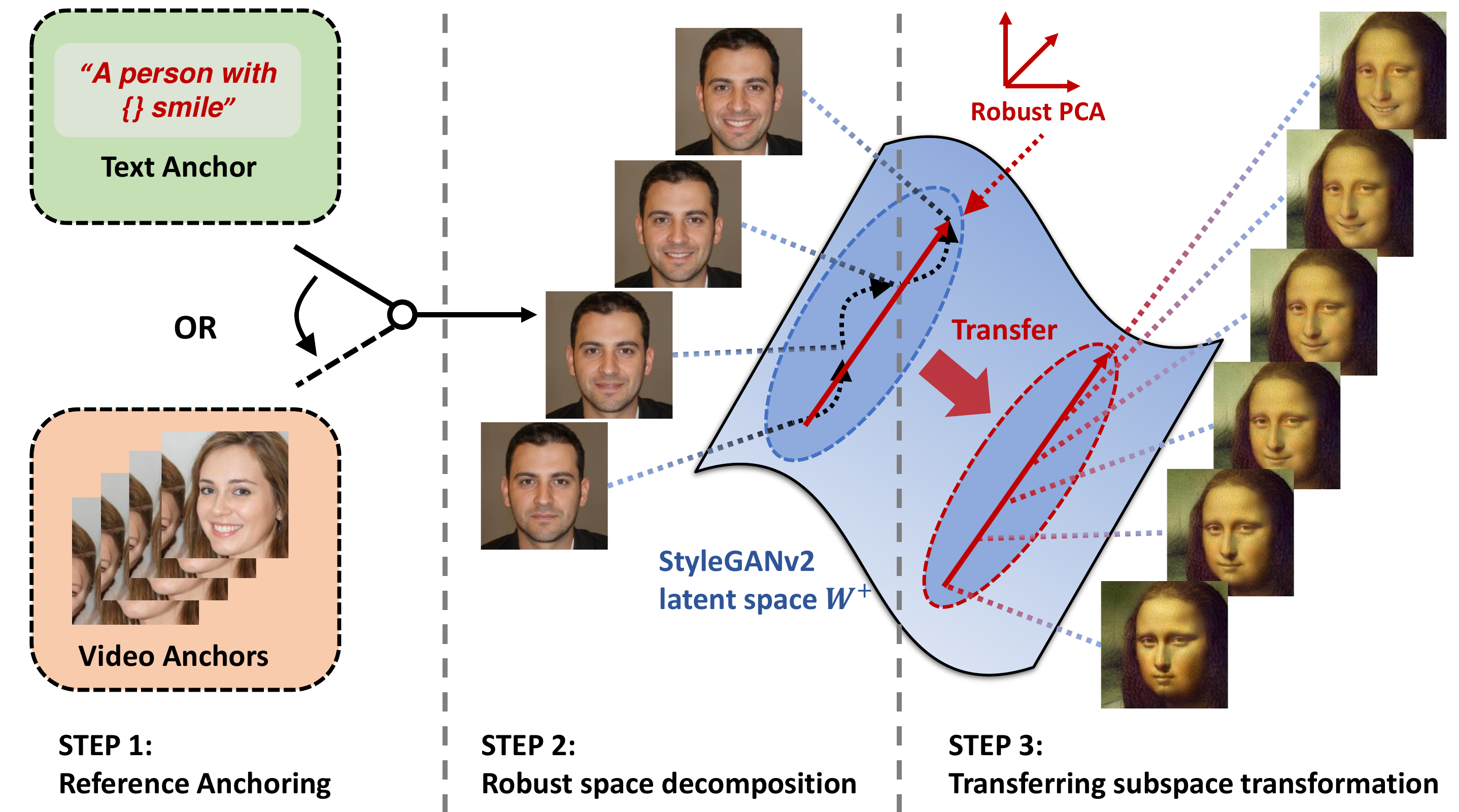}
\caption{\textbf{Our workflow illustration.} In our workflow, we first extract a low-dimensional micromotion subspace from one identity, and then transfer it to a novel identity ``Mona-lisa''.}
\label{fig:prelim_methods}
\vspace{-1em}
\end{figure}

With this hypothesis, we design a workflow to extract the edit direction from decomposed low-dimensional micromotion subspace, illustrated in Figure~\ref{fig:prelim_methods}. Our complete workflow can be distilled down to three simple steps: (a) collecting anchor latent codes from a single identity; (b) enforcing robustness linear decomposition to obtain a noise-free low-dimensional space; (c) applying the extracted edit direction from low-dimensional space to arbitrary input identities.

\paragraph{Step 1: Reference Anchoring.} To find the edit direction of a specific micromotion, we first acquire a set of latent codes corresponding to the desired action performed by the same person. Serving as anchors, these latent codes help to disentangle desired micromotions in later steps. Here, we consider two approaches, text-anchored and video-anchored methods, respectively.

\underline{Text-anchored Reference Generation:} Recent work, StyleCLIP~\cite{patashnik2021styleclip}, has shown that expressive phrases can successfully manipulate the properties of the synthesized images. In this method, we plan to leverage the StyleCLIP latent optimization pipeline to generate the anchoring latent codes for desired micromotions. The main-idea to optimize these latent codes is to minimize the contrastive loss between the designed input texts and the images rendered by the codes with a few regularizations. Here, one major question is how to design the most appropriate text template to guide the optimization. To generate images with only variance in degrees of micromotions, a natural method is to specify the degrees in the text, where we concatenate a series of adjectives or percentages with the micromotion description text to indicate the various strength and the stage of the current micromotion. For example, for the micromotion ``eyes closed'', we use both percentages and adjectives to modify the micromotion by specifying ``eyes \textit{greatly/slightly} closed'' and ``eyes \textit{10\%/30\%} closed''.  Here, we emphasize that this is just one of the possible text prompts design options. We compare various choices of text prompts, and the experiments of the text prompt choices will be covered in the ablation study.

\underline{Video-anchored Reference Generation:} StyleCLIP relies on text guidance to optimize the latent codes, while for abstract and complicated motions, such as a gradual head movement with various head postures, the text might not be able to express the micromotion concisely. To overcome this issue, we leverage a reference video demonstration to anchor the micromotion subspace instead. In the reference video-based anchoring methods, we use frames of reference videos to decode the desired micromotions. Specifically, given a reference video that consists of continuous frames, we invert these frames with a pre-trained StyleGAN encoder to obtain the reference latent codes. We emphasize that different from the per-frame editing method, the goal of using reference video frames is to anchor the low-dimensional micromotion subspace. Thus, we use significantly fewer frames than per-frame editing methods, and no further video frames are required once we extract such space.

After applying either anchoring method, we obtain a set of $t_n$ referential latent codes denoted as \{$\matr{V}_{t1}$, $\matr{V}_{t2}$, …, $\matr{V}_{tn}$\} from only a single identity. These codes will be the keys to obtain a low-dimensional micromotion space in later steps.

\paragraph{Step 2: Robust space decomposition.}
Due to the randomness of the optimization and the complicacy of image contents (e.g., background distractors), the latent codes from the previous step could be viewed as ``noisy samples'' from the underlying low-dimensional space. 
Therefore, based on our low-rank hypothesis, we leverage further  decomposition methods to robustify the latent codes and their shared micromotion subspace.

The first simple decomposition method we adopt is the principal component analysis (PCA), where each anchoring latent code serves as the row vector of the data matrix. Unfortunately, merely using PCA is insufficient for a noise-free micromotion subspace, since the outliers in latent codes degrade the quality of the extracted space.
As such, we further turn to a classical technique called \textit{robust PCA}~\cite{wright2009robust}, which can recover the underlining low-rank space from the latent codes with sparse gross corruptions. It can be formulated as a convex minimization of a nuclear norm plus an $\ell_1$ norm and solved efficiently with alternating directions optimization~\cite{candes2011robust}. That yields a more robust micromotion subspace to determine the micromotion edit direction $\Delta \matr{V}$.

\paragraph{Step 3: Applying the subspace transformation.}
Once the edit direction $\Delta \matr{V}$ is obtained, we could edit any arbitrary input faces for the micromotion. Specifically, the editing is conducted simply through interpolation and extrapolation along this latent direction to obtain the intermediate frames. For an arbitrary input image $I_0^{\prime}$, we find its latent code $\matr{V}_0^{\prime}= E(I_0^{\prime})$, and the videos can be synthesized through 
\begin{equation}
I_t = G(\matr{V}_t) = G(\matr{V}_0 + \alpha t\Delta \matr{V}),
\end{equation}
where $\alpha$ is a parameter controlling the degree of interpolation and extrapolation, $t$ corresponds to the index of the frame, and the resulting set of frames $\{I_t\}$ collectively construct the desired micromotion such as ``smiling'', ``eyes opening''. Combining these synthesized frames, we obtain a complete video corresponding to the desired micromotion.

This general pipeline can be applied to arbitrary micromotions. Once the latent micromotion subspace is found, this space can be applied to both in-domain and out-domain identities with no further cost.

\section{Experiments}
\begin{figure*}[!htb]
\centering
\begin{subfigure}{\textwidth}
\centering
\begin{tabular}{ccccc}
\includegraphics[width=0.18\textwidth]{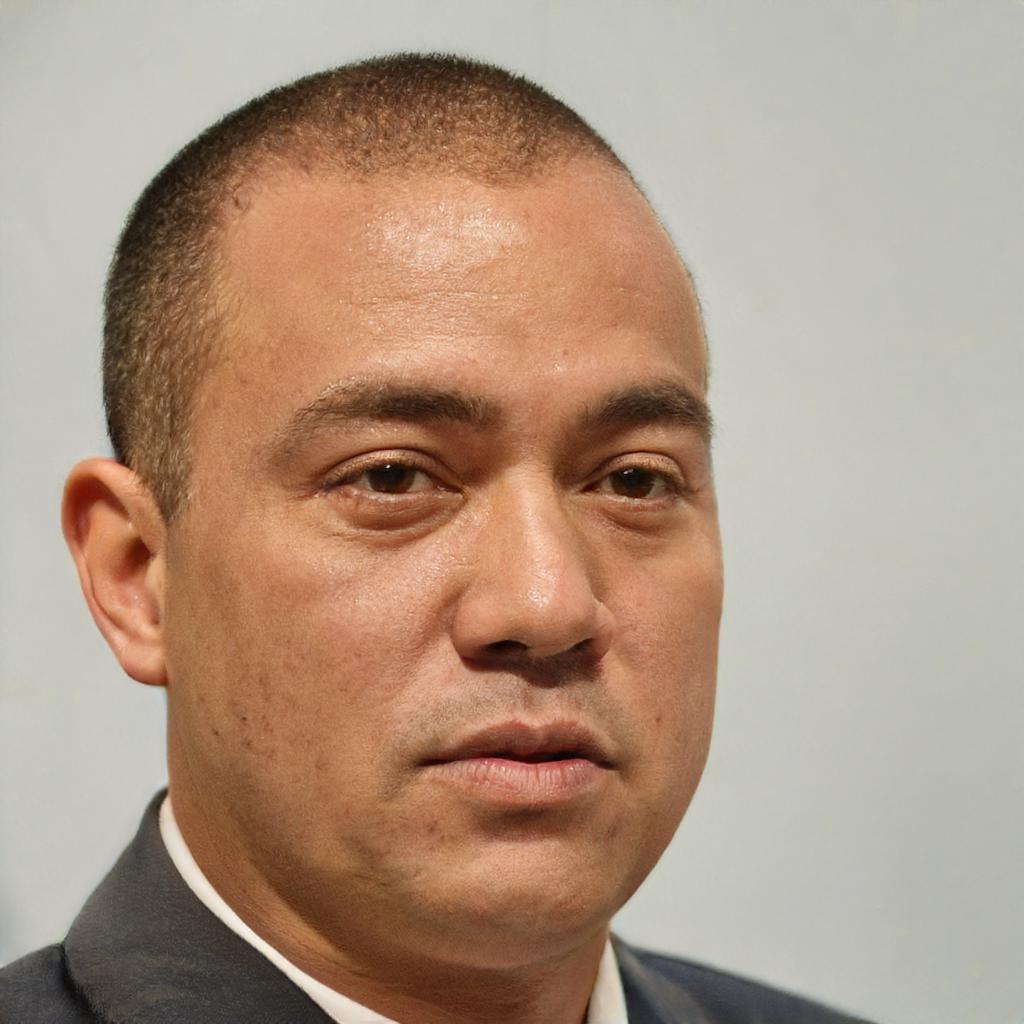}
\includegraphics[width=0.18\textwidth]{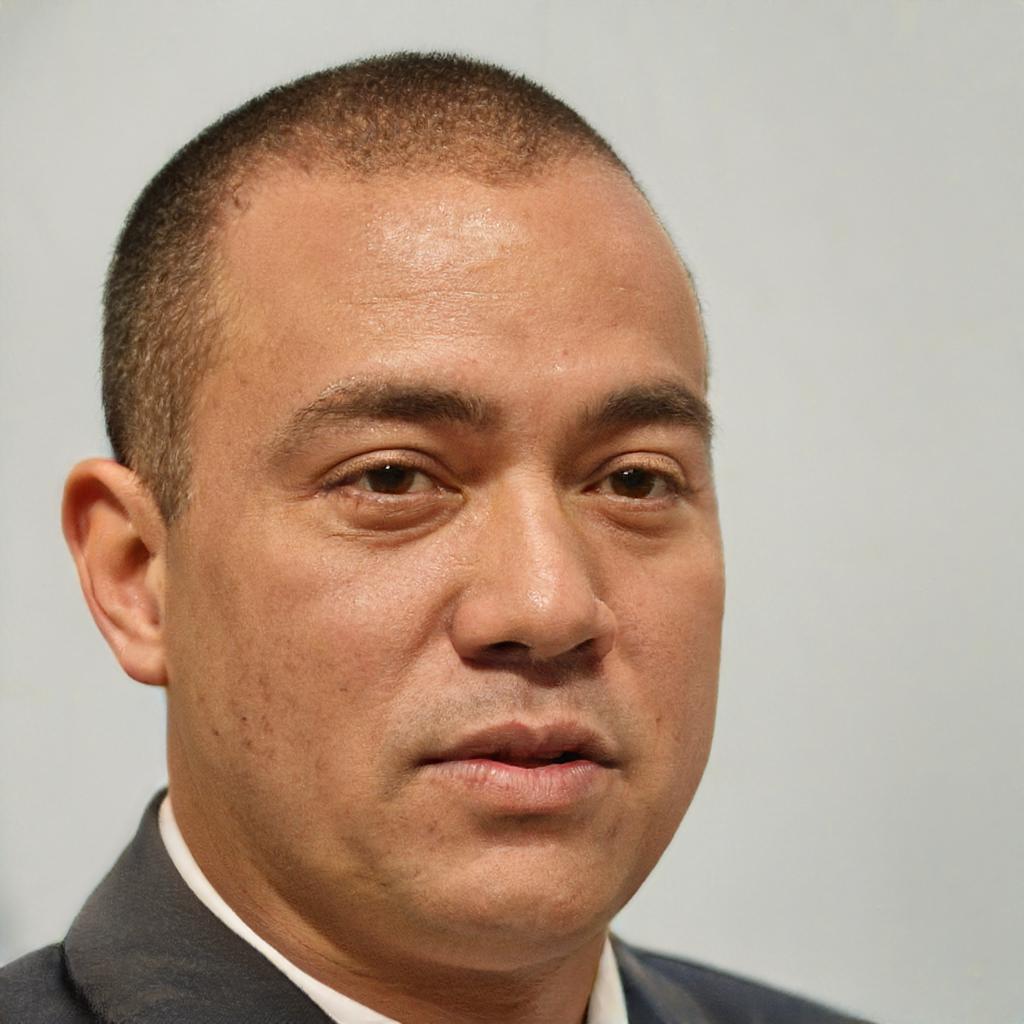}
\includegraphics[width=0.18\textwidth]{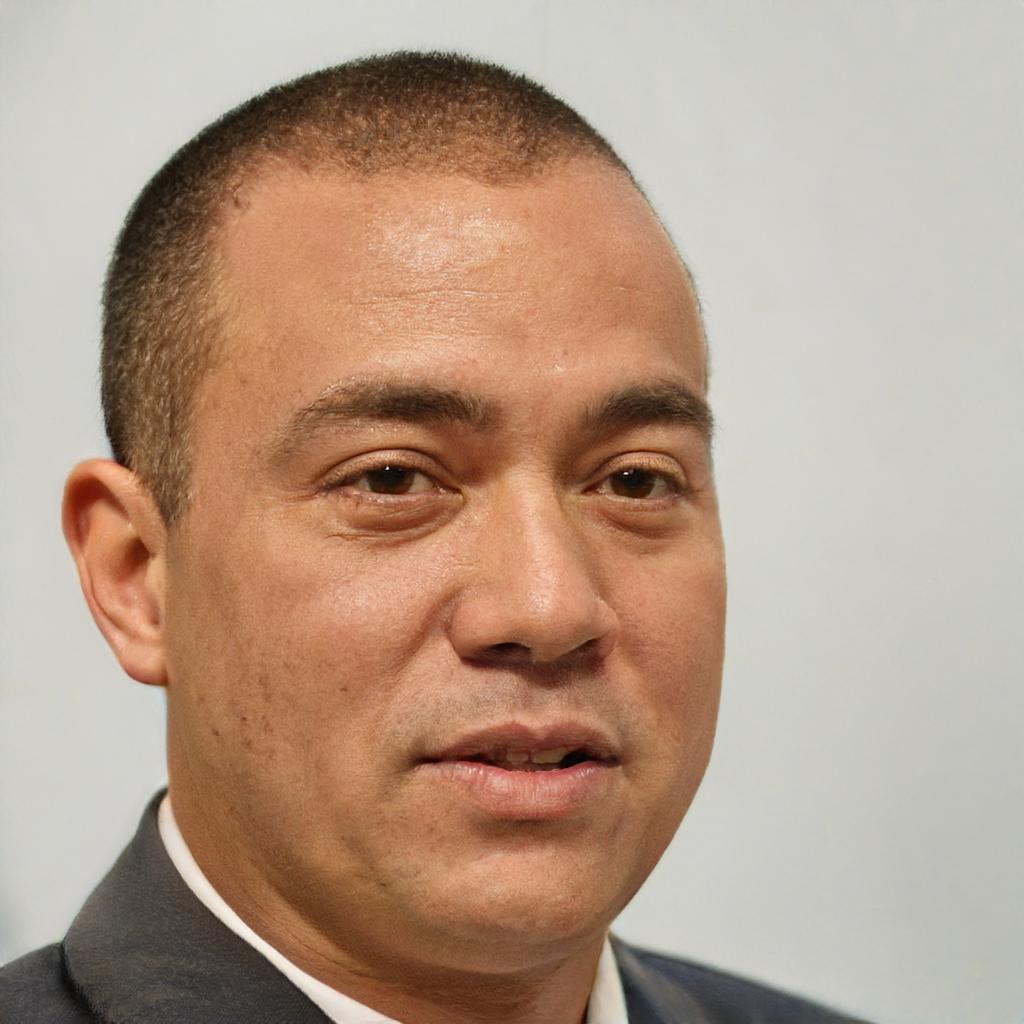}
\includegraphics[width=0.18\textwidth]{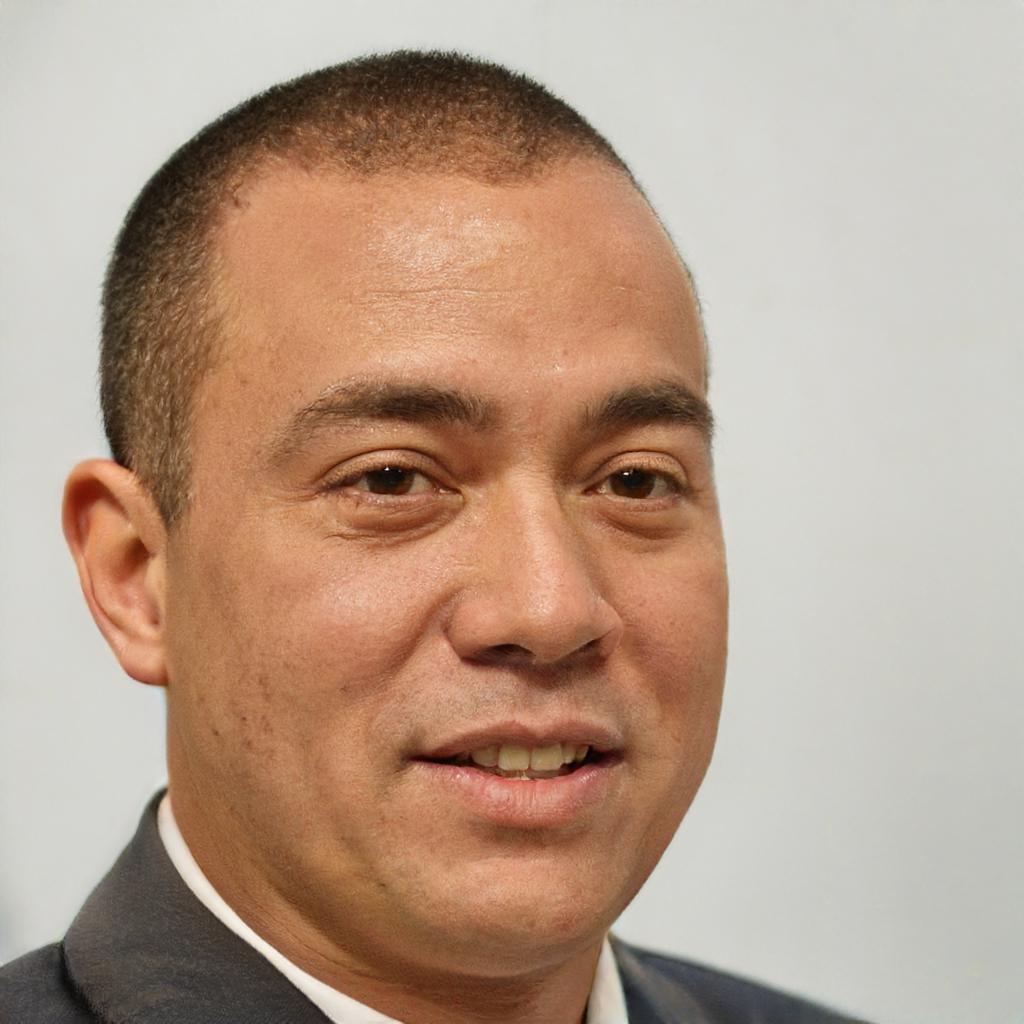}
\includegraphics[width=0.18\textwidth]{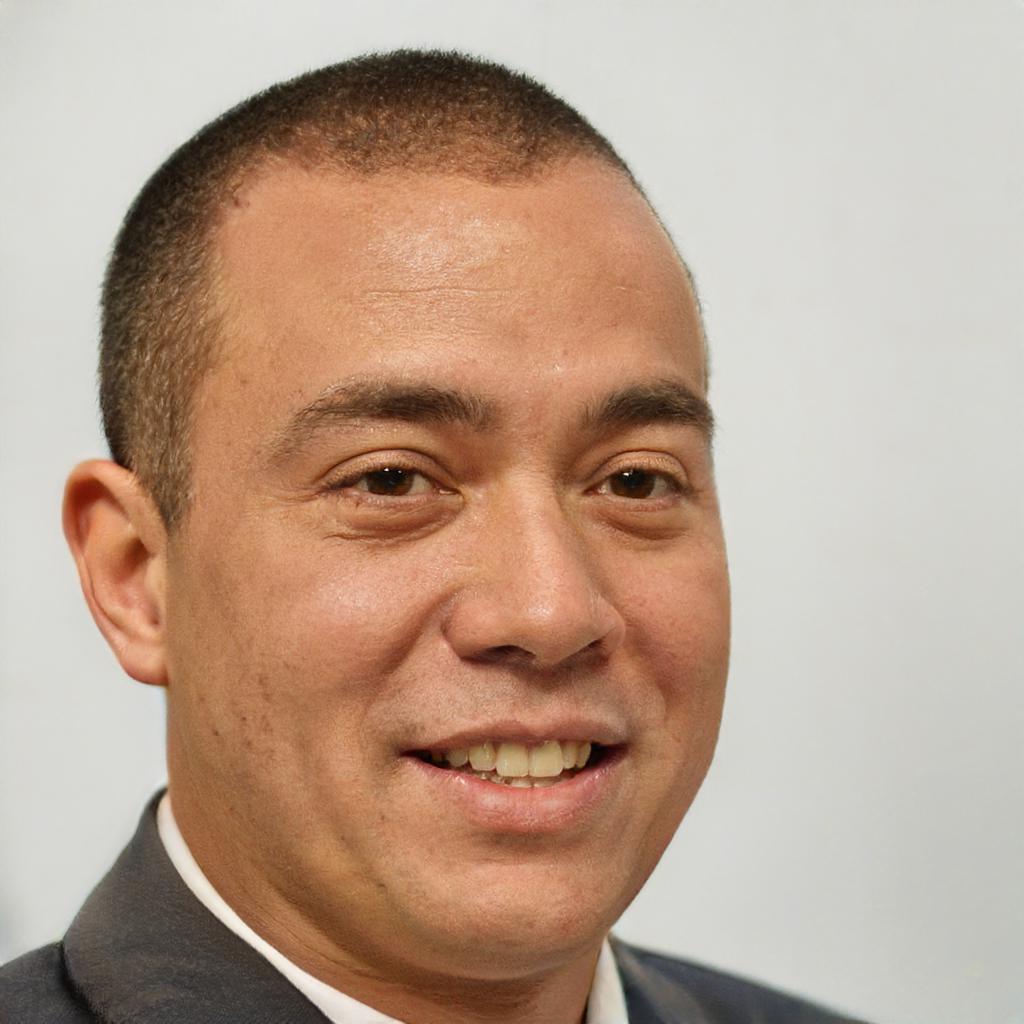}
\end{tabular}
\vspace{-1.0em}
\caption{Smiling}
\label{fig:micromotions:smiling}
\end{subfigure}
\begin{subfigure}{\textwidth}
\centering
\begin{tabular}{ccccc}
\includegraphics[width=0.18\textwidth]{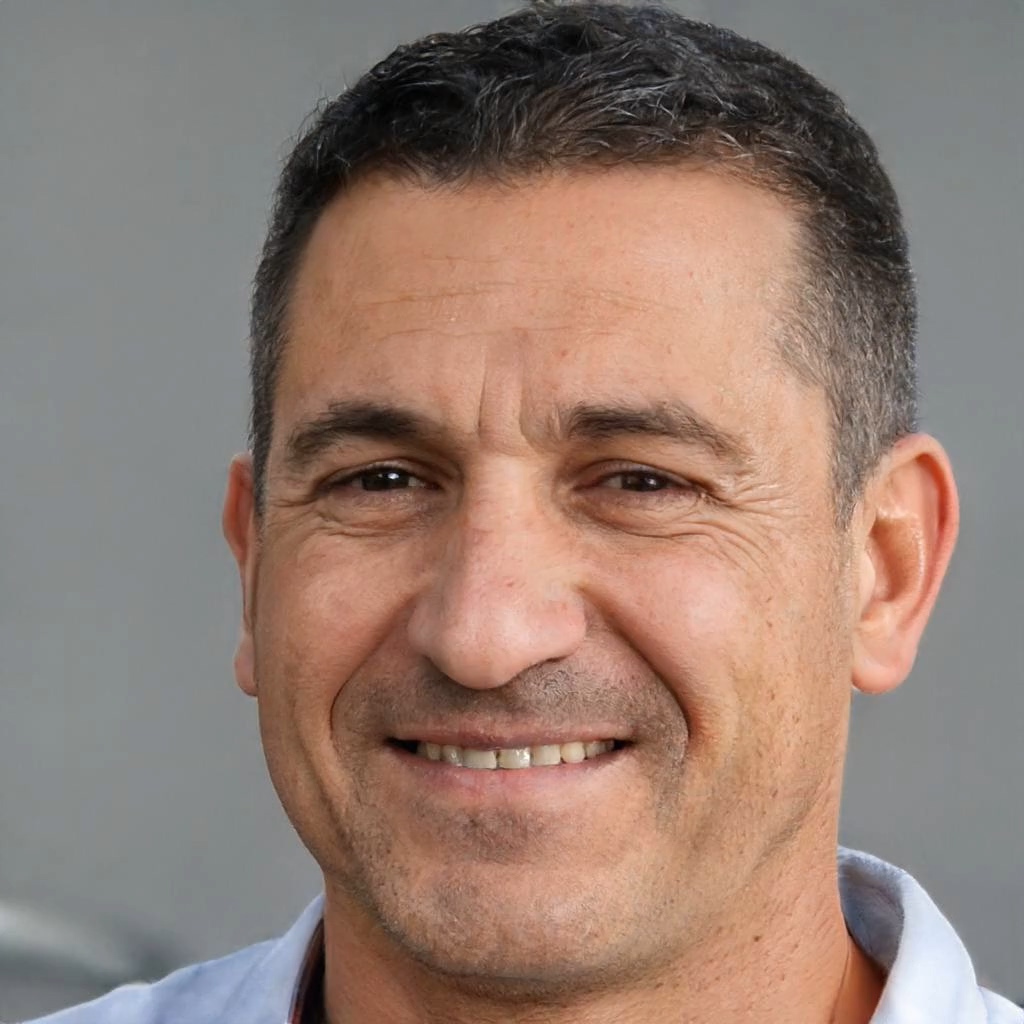}
\includegraphics[width=0.18\textwidth]{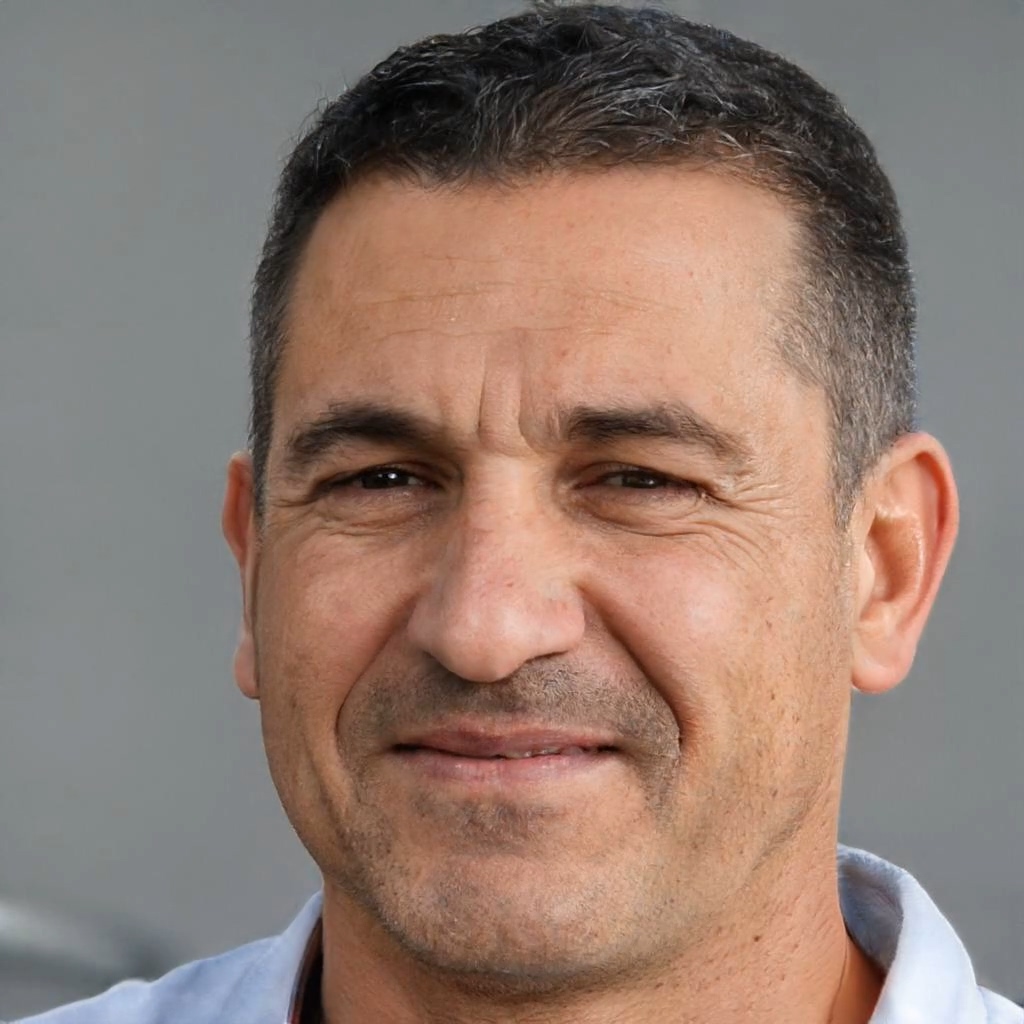}
\includegraphics[width=0.18\textwidth]{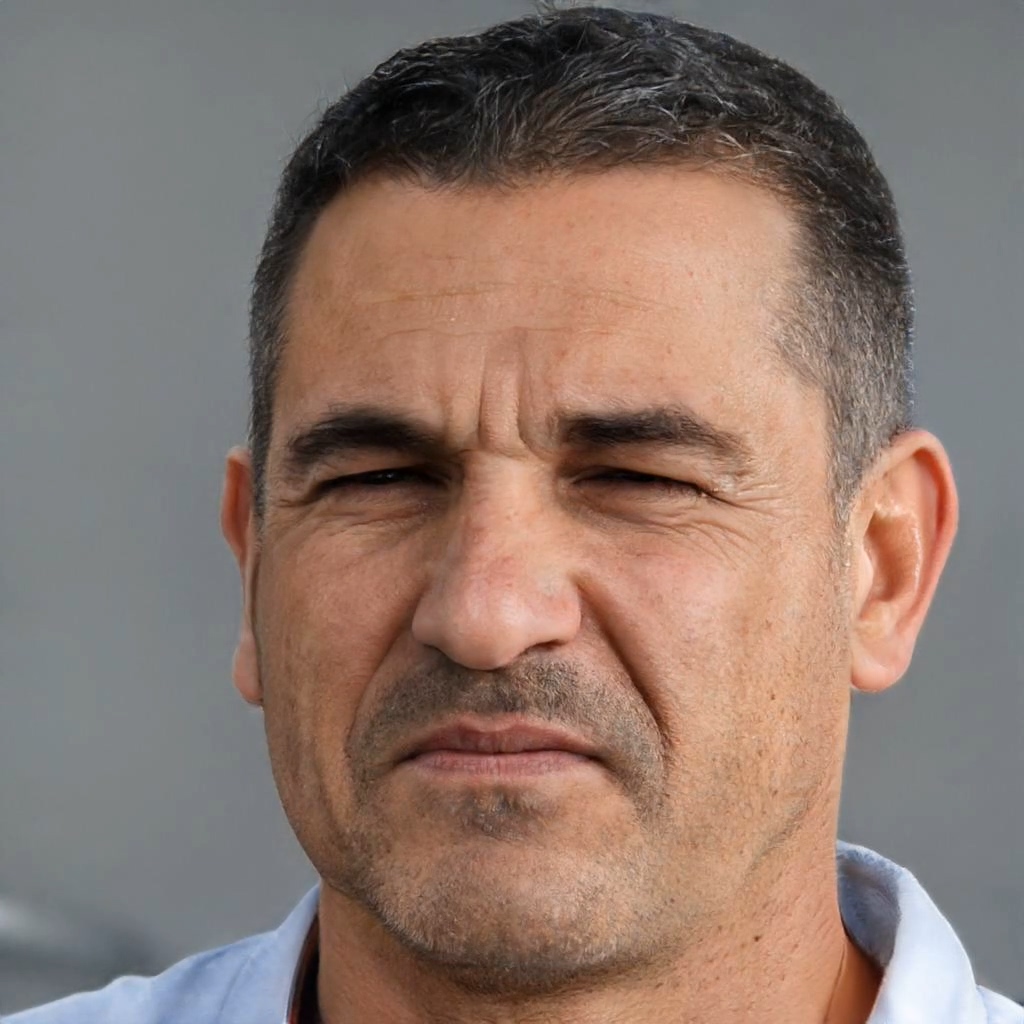}
\includegraphics[width=0.18\textwidth]{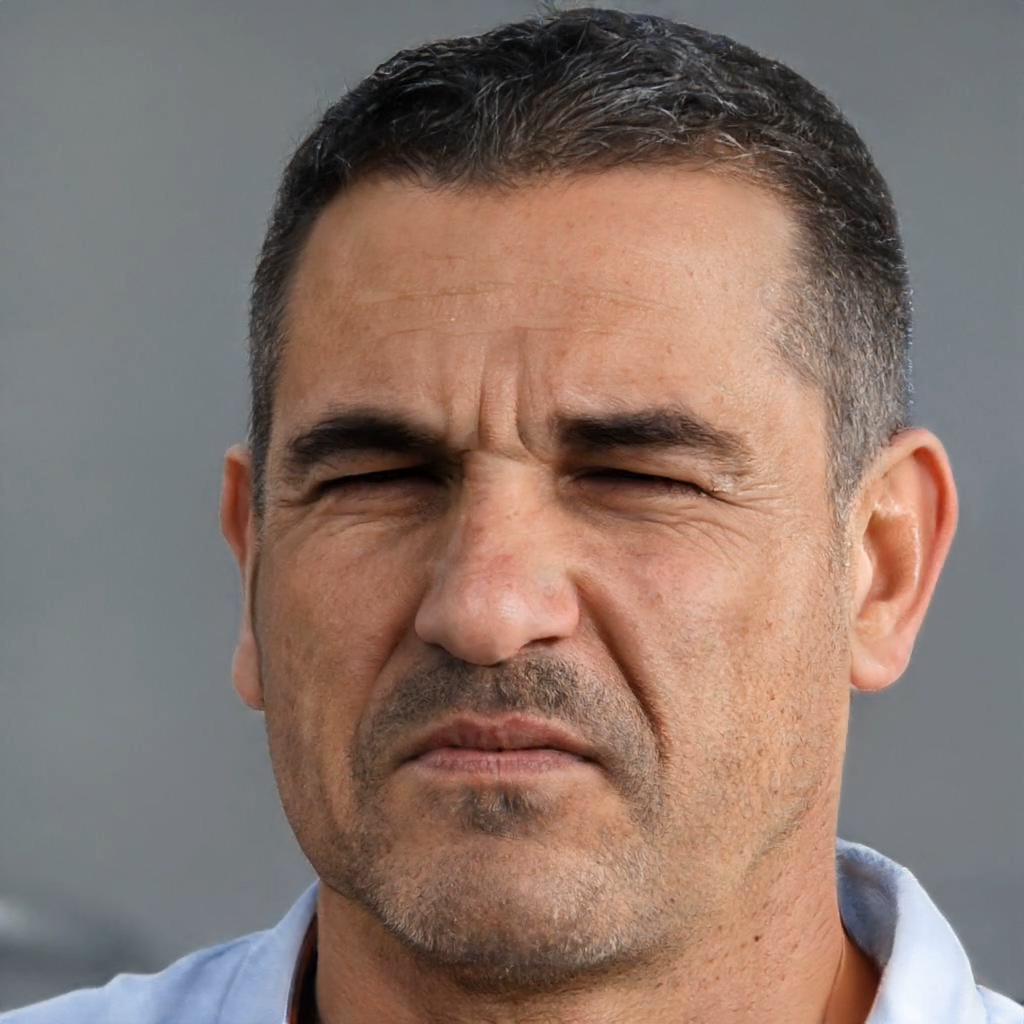}
\includegraphics[width=0.18\textwidth]{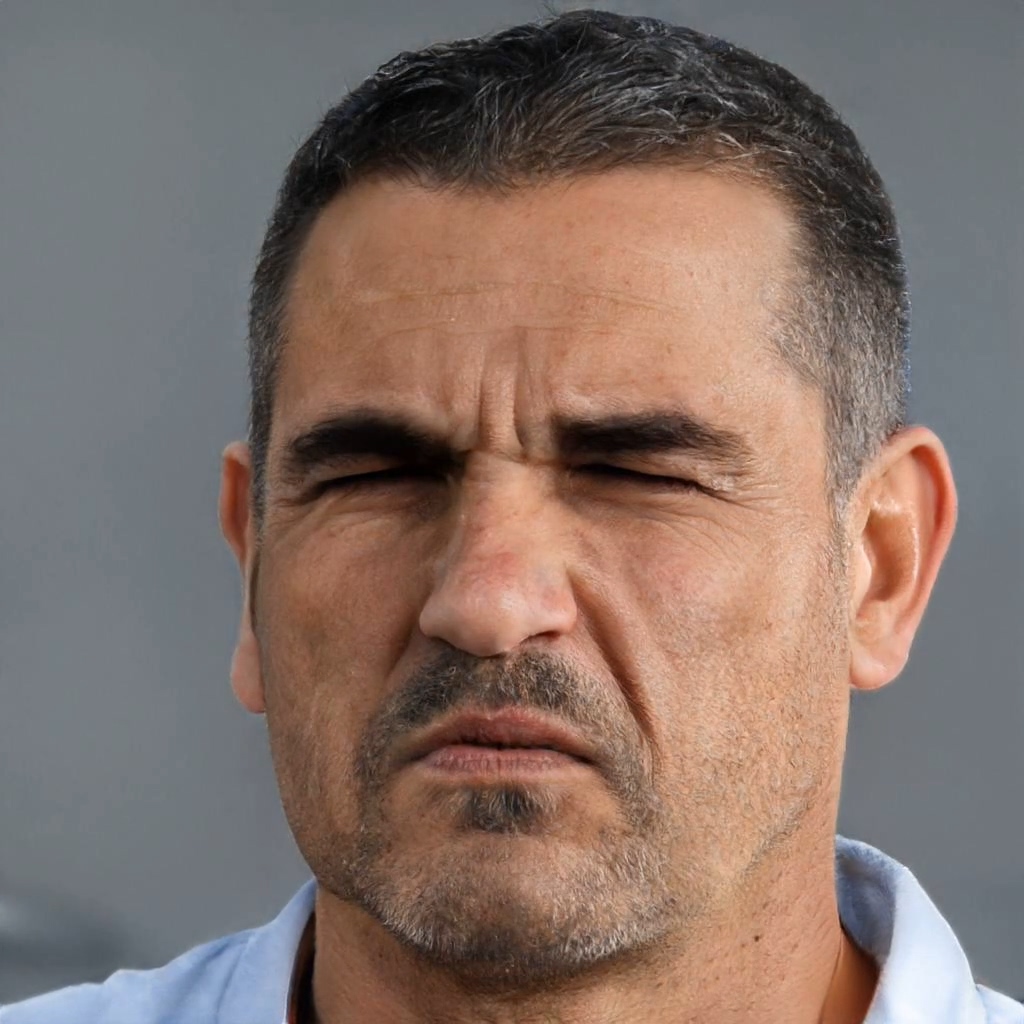}
\end{tabular}
\vspace{-1.0em}
\caption{Anger}
\label{fig:micromotions:angry}
\end{subfigure}
\begin{subfigure}{\textwidth}
\centering
\begin{tabular}{ccccc}
\includegraphics[width=0.18\textwidth]{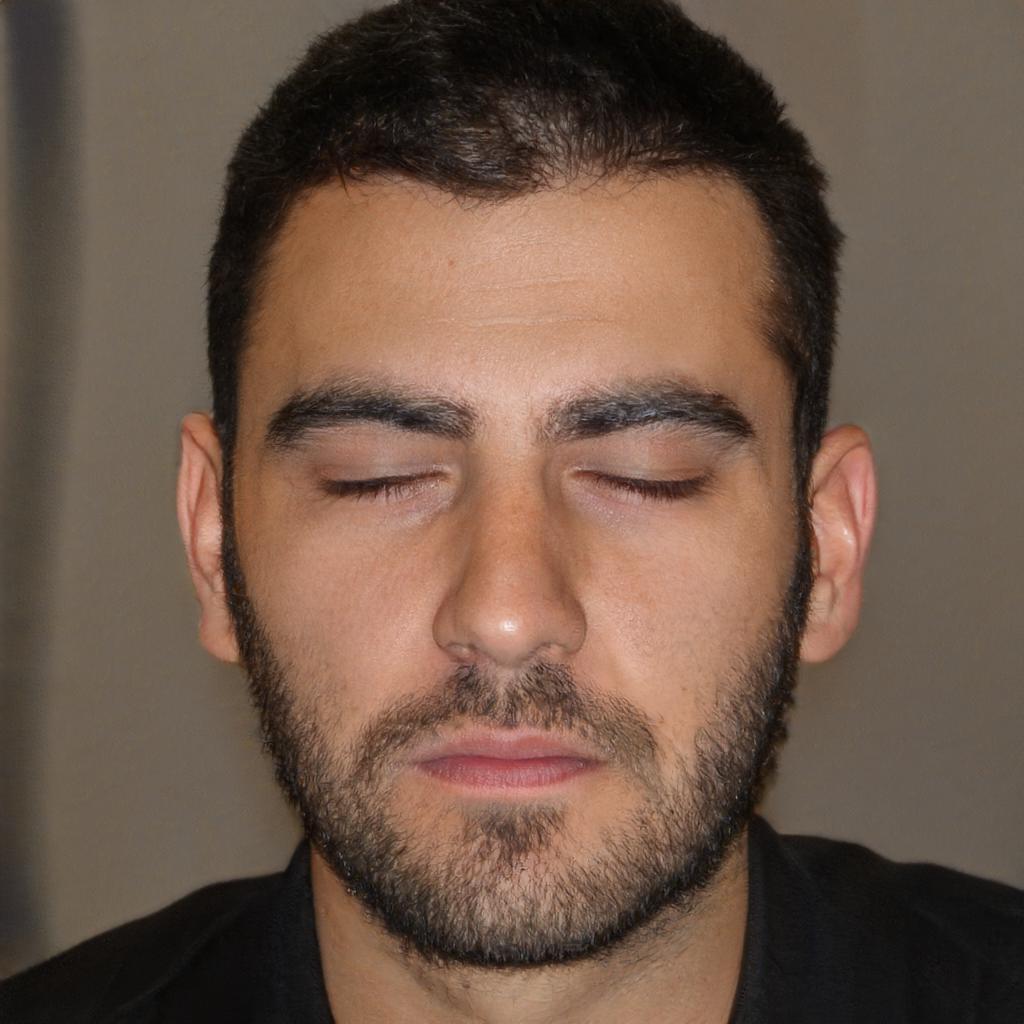}
\includegraphics[width=0.18\textwidth]{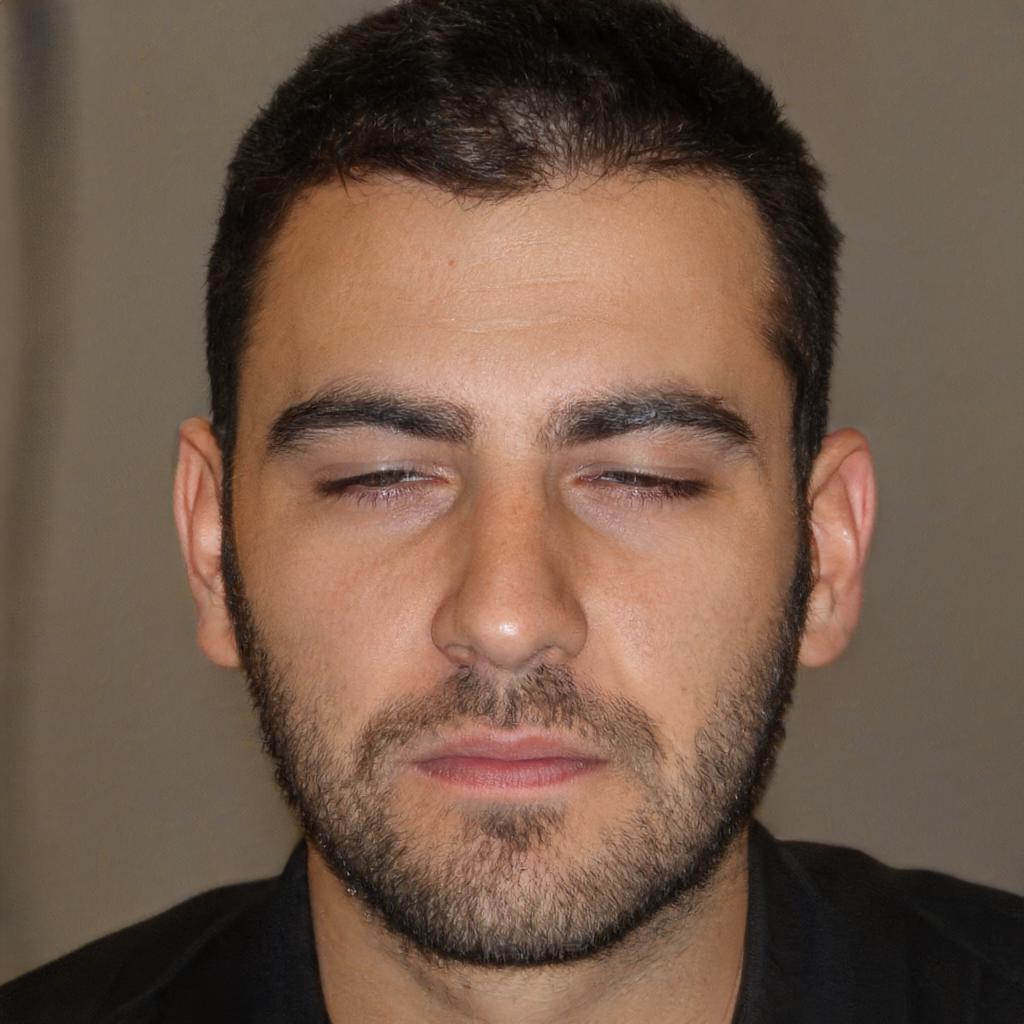}
\includegraphics[width=0.18\textwidth]{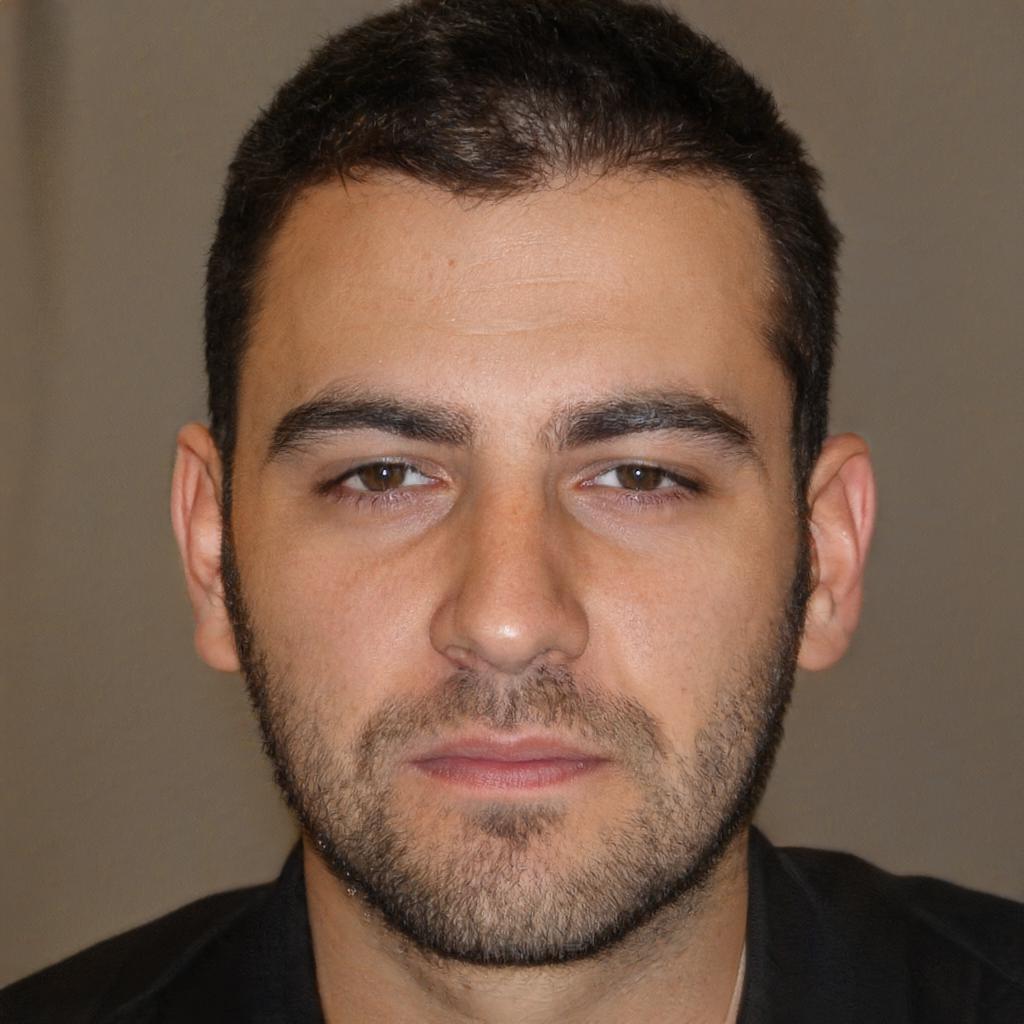}
\includegraphics[width=0.18\textwidth]{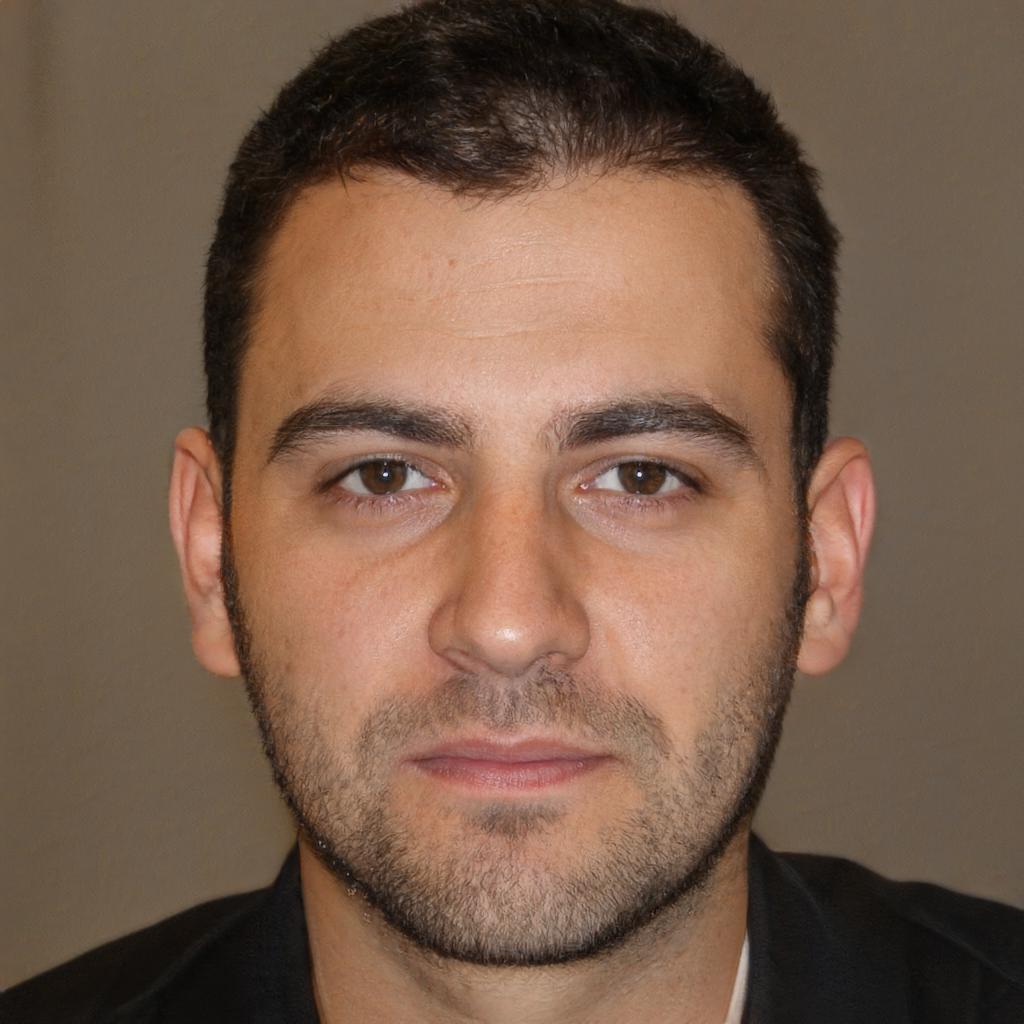}
\includegraphics[width=0.18\textwidth]{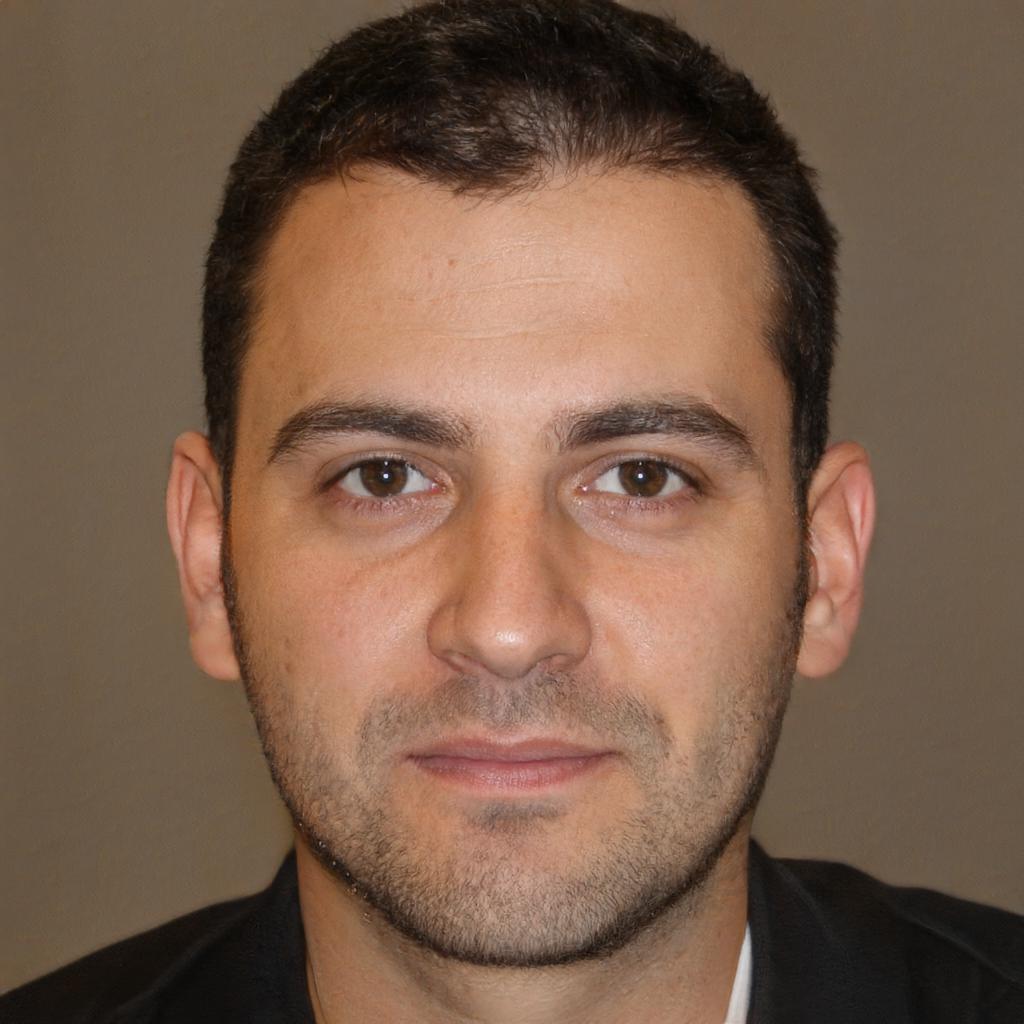}
\end{tabular}
\vspace{-1.0em}
\caption{Opening Eyes}
\label{fig:micromotions:opening_eyes}
\end{subfigure}
\begin{subfigure}{\textwidth}
\centering
\begin{tabular}{ccccc}
\includegraphics[width=0.18\textwidth]{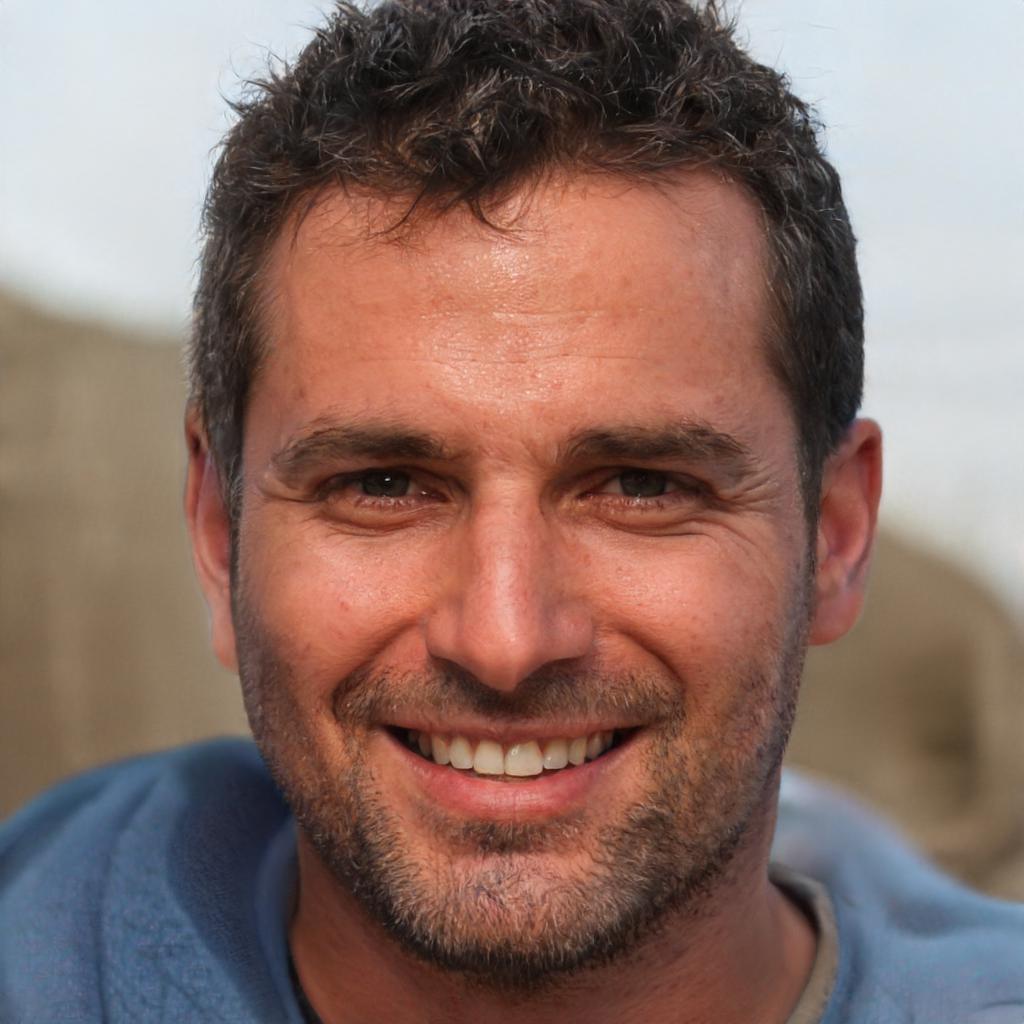}
\includegraphics[width=0.18\textwidth]{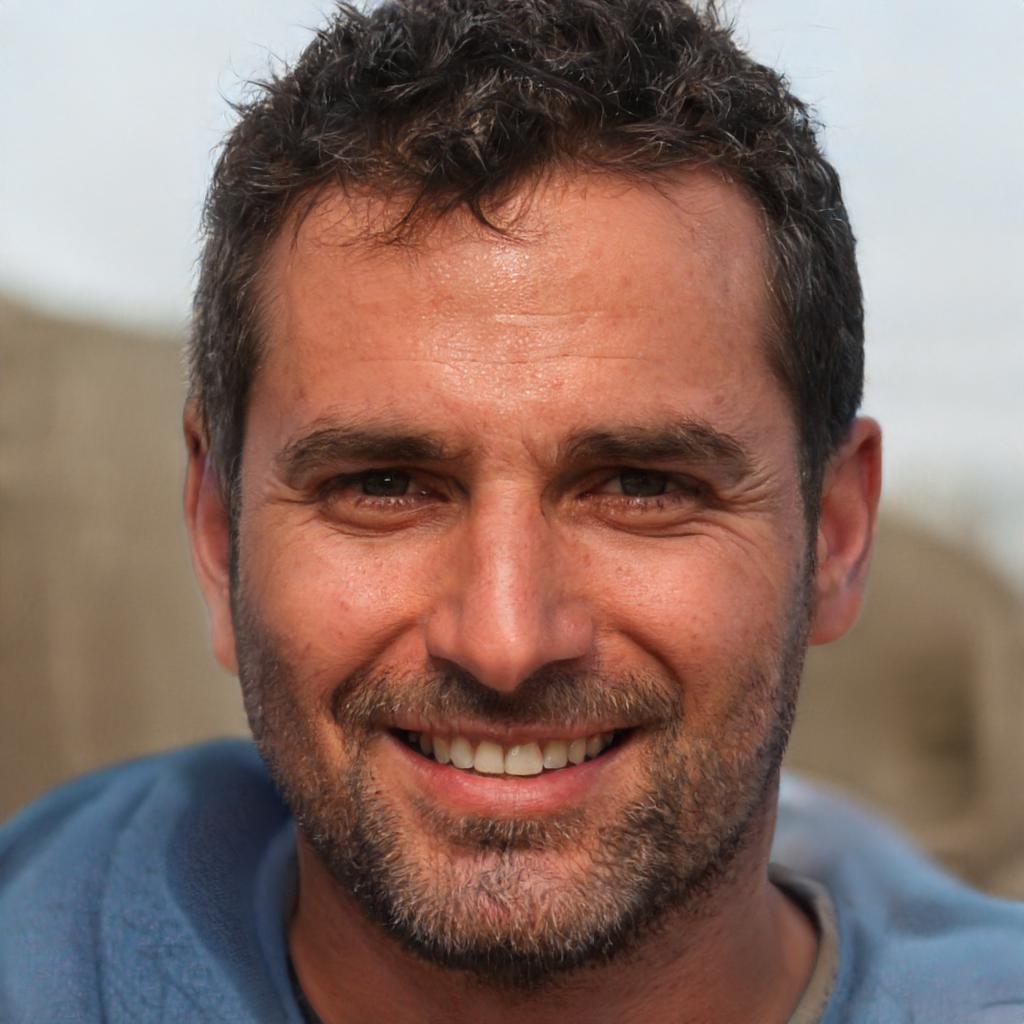}
\includegraphics[width=0.18\textwidth]{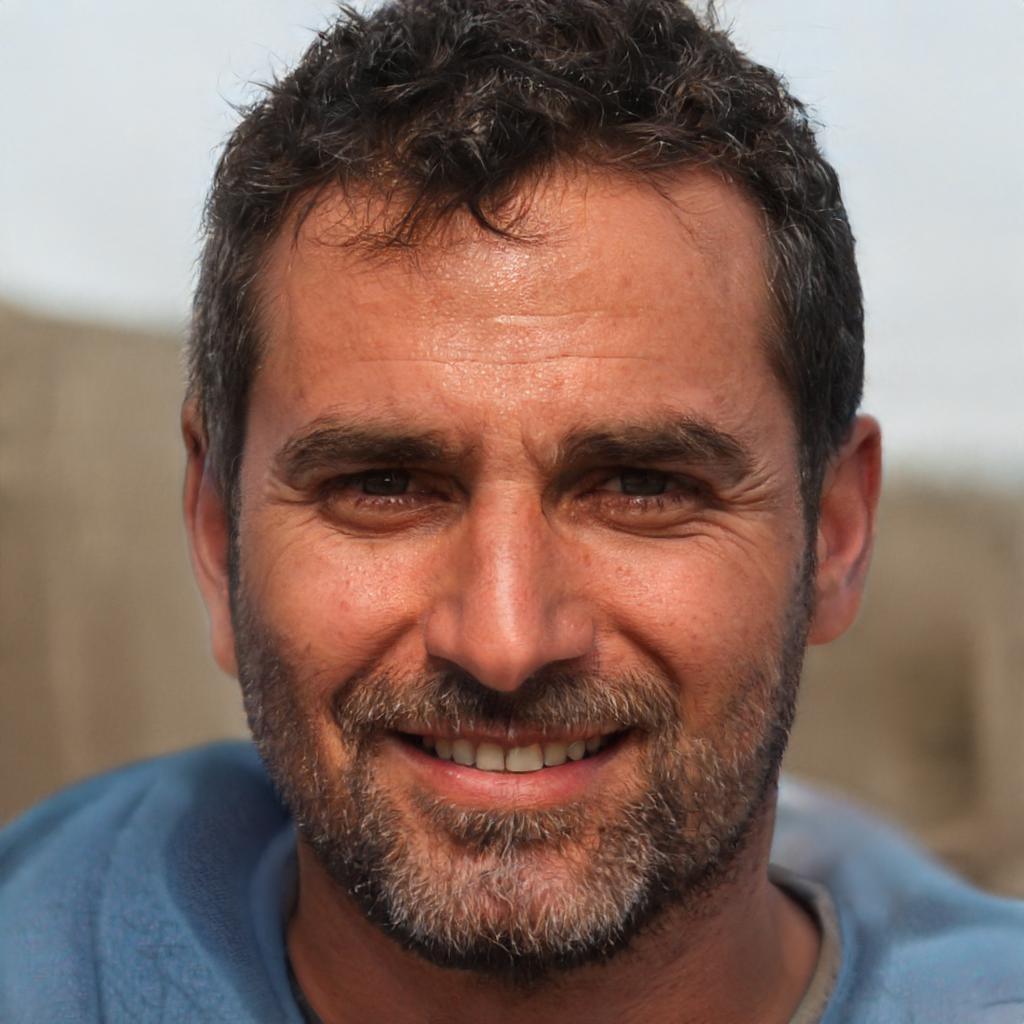}
\includegraphics[width=0.18\textwidth]{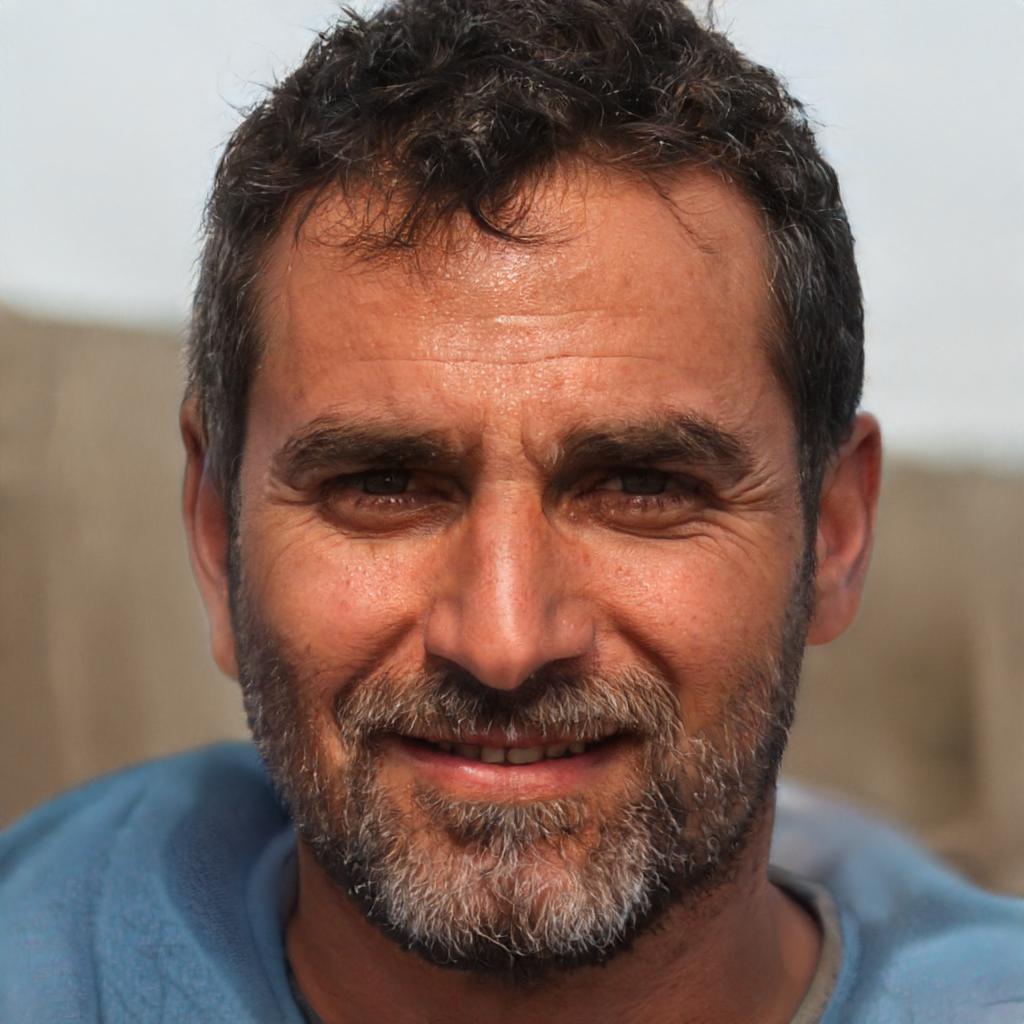}
\includegraphics[width=0.18\textwidth]{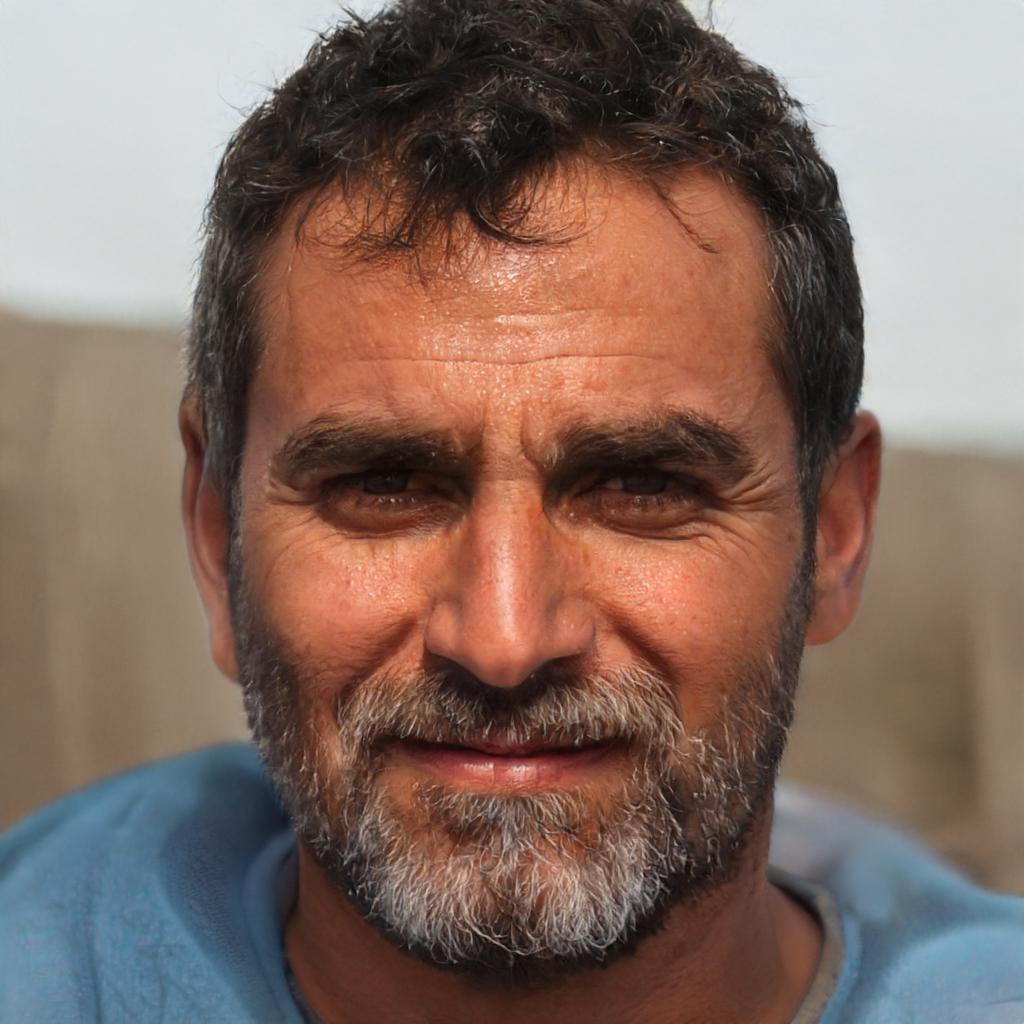}
\end{tabular}
\vspace{-1.0em}
\caption{Aging Face}
\label{fig:micromotions:aging_face}
\end{subfigure}
\vspace{-0.6em}
\caption{\textbf{Illustrations of versatile micromotions founded by text-anchored method.} We decode the micromotions across different identities, and apply them to in-domain identities.
From Top to Bottom: (a) Smiling (b) Anger (c) Opening Eyes (d) Aging Face. Best view when zoomed in. Please refer to our repository for complete video sequences. }
\label{fig:versatile_micromotions}
\end{figure*}
In the experiments, we focus on the following questions related to our hypothesis and workflow:
\begin{compactitem}
    \item Could our pipeline locate subspaces for various meaningful micromotions?
    \item Could the subspaces be effectively decoded from only a few identities, even only one?
    \item Could we transfer decoded micromotion subspace to other subjects in both the same domain and across the domain?
    \item Could we extend the micromotions to novel subjects with no computation overhead?
\end{compactitem}

In short, we 
want to prove two concepts in the following experiments: (a) \textbf{Universality}: The single pipeline can handle various micromotion, and the single decoded micromotion can be extended to different subjects in various domains; (b) \textbf{Lightweight}: Transferring the micromotion only requires a small computation overhead.

To explore these two concepts in our workflow, we now turn to analyze our proposed methods on the synthesized micromotions. We consider five micromotions as examples: (a) smiling, (b) angry, (c) opening eyes, (d) turning head, and (e) aging face. Following the workflow, we obtain the robustness aware edit directions for each micromotion from one face image, and then synthesize on other cross-domain images including characters in animations, sculptures, paintings with different styles.

\begin{figure*}[!htb]
\centering
\begin{subfigure}{\textwidth}
\centering
\begin{tabular}{ccccc}
\includegraphics[width=0.18\textwidth]{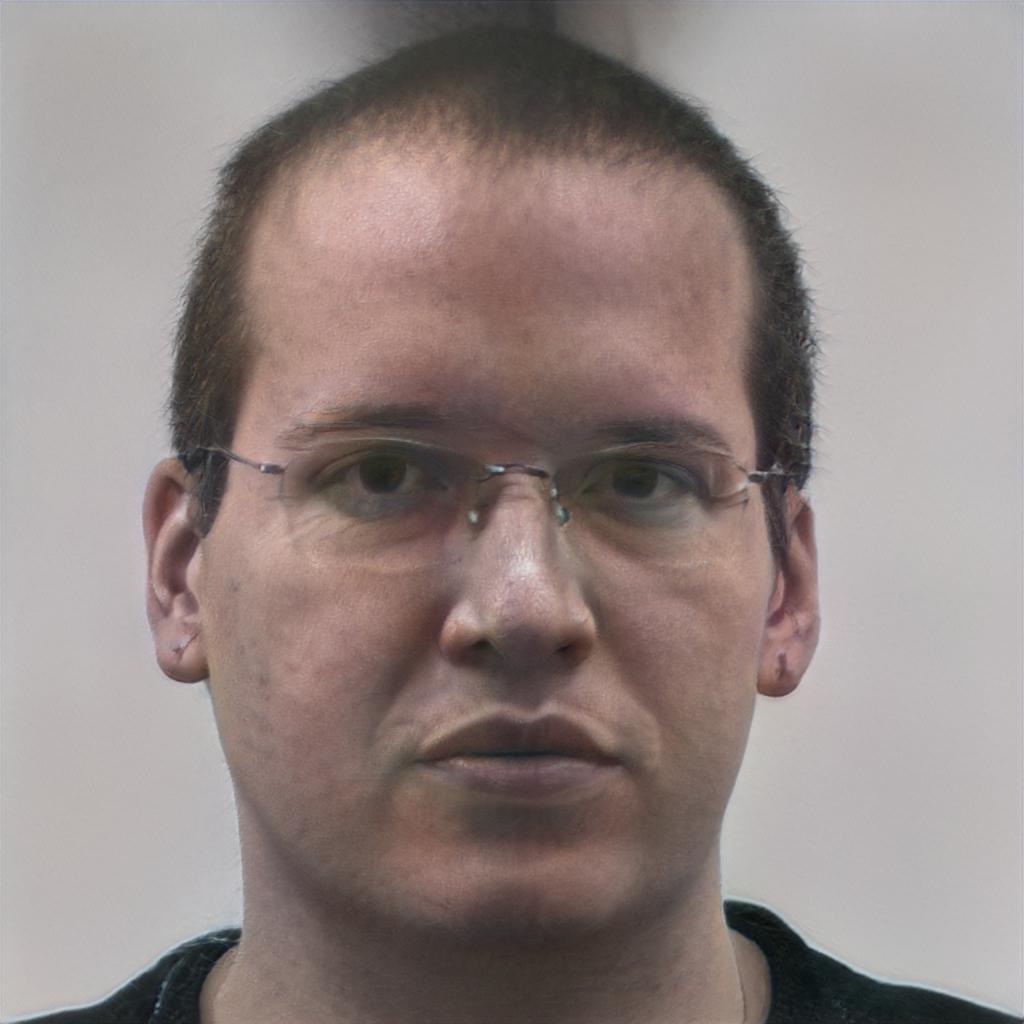}
\includegraphics[width=0.18\textwidth]{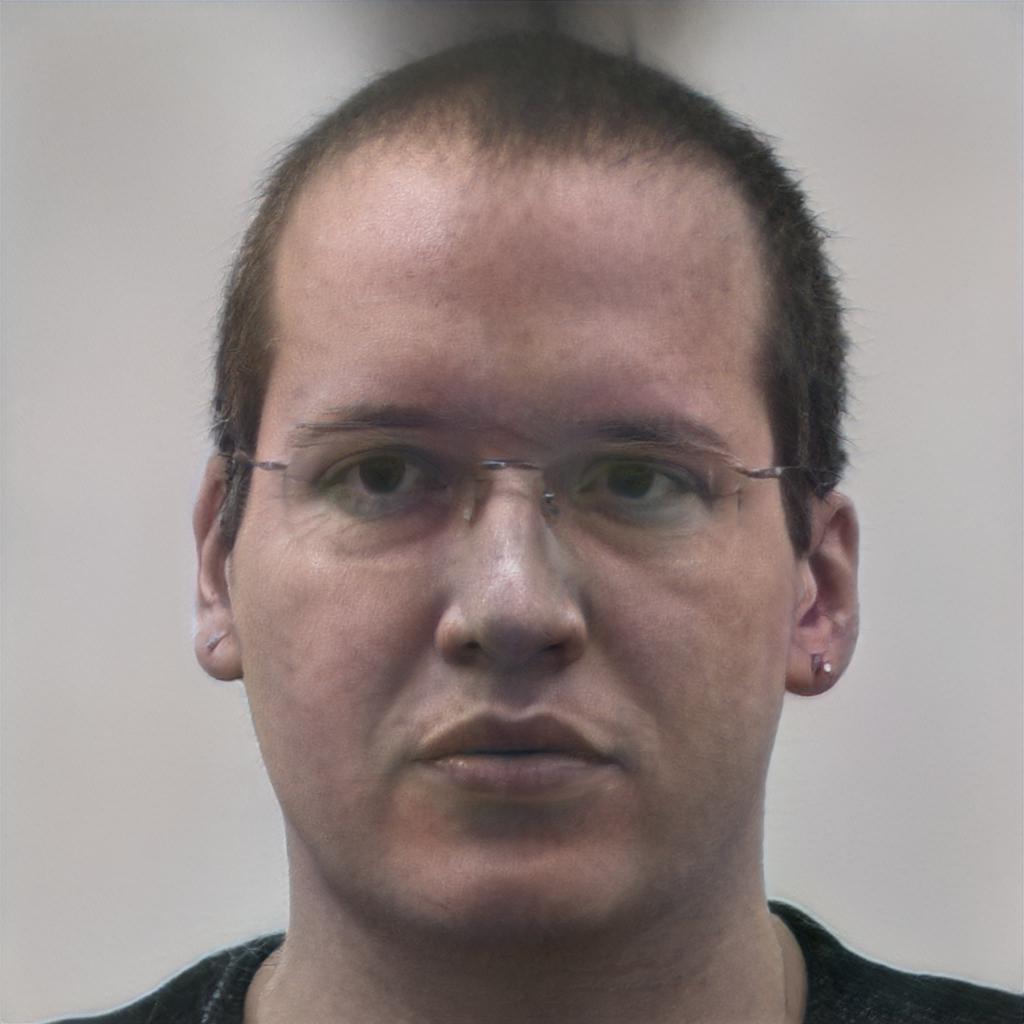}
\includegraphics[width=0.18\textwidth]{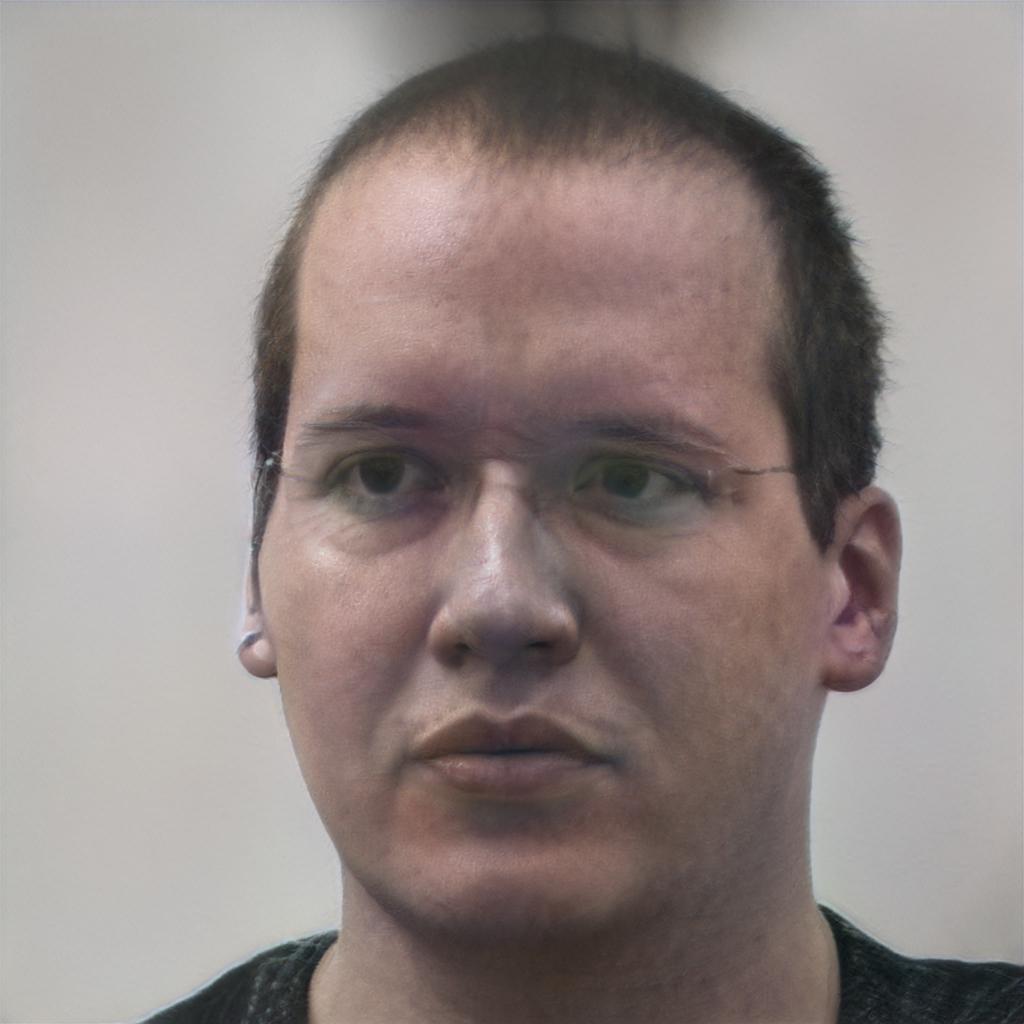}
\includegraphics[width=0.18\textwidth]{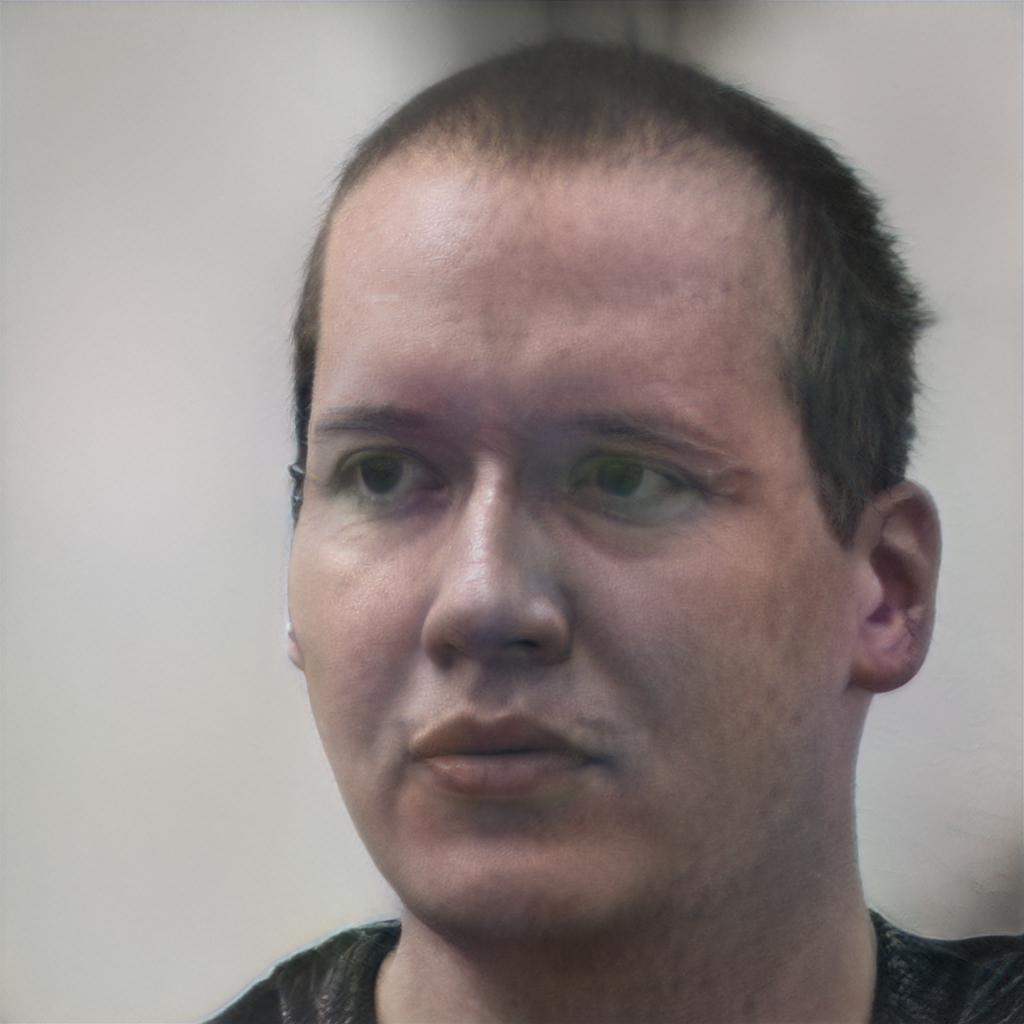}
\includegraphics[width=0.18\textwidth]{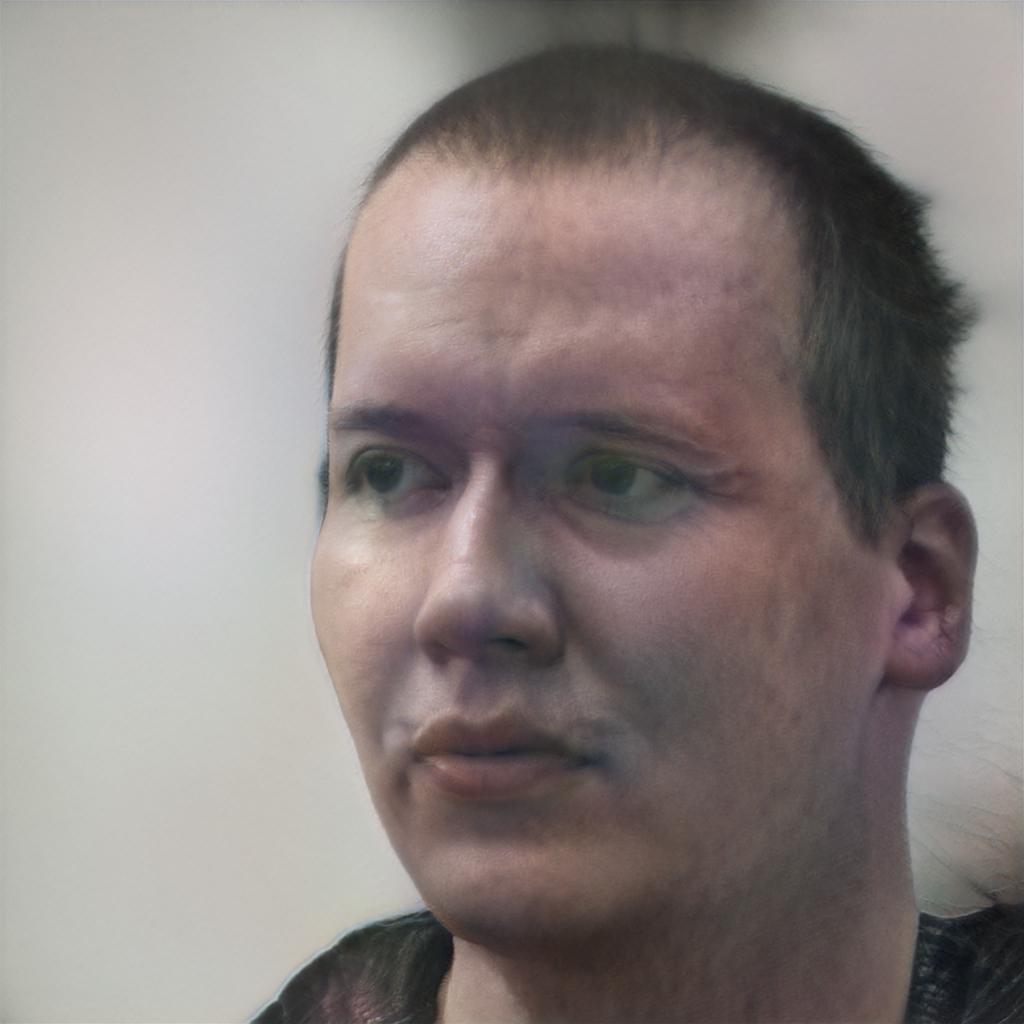}
\end{tabular}
\label{fig:micromotions:turning_head}
\end{subfigure}
\begin{subfigure}{\textwidth}
\centering
\begin{tabular}{ccccc}
\includegraphics[width=0.18\textwidth]{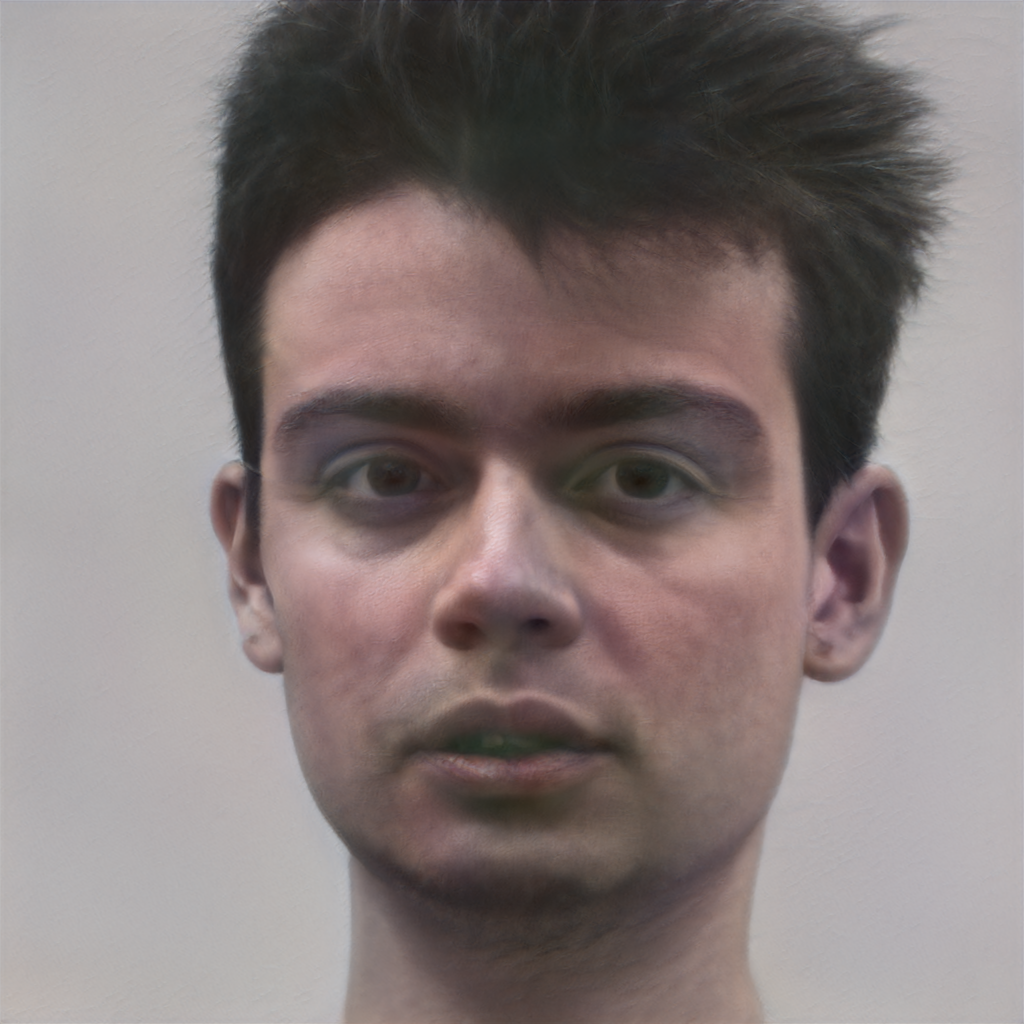}
\includegraphics[width=0.18\textwidth]{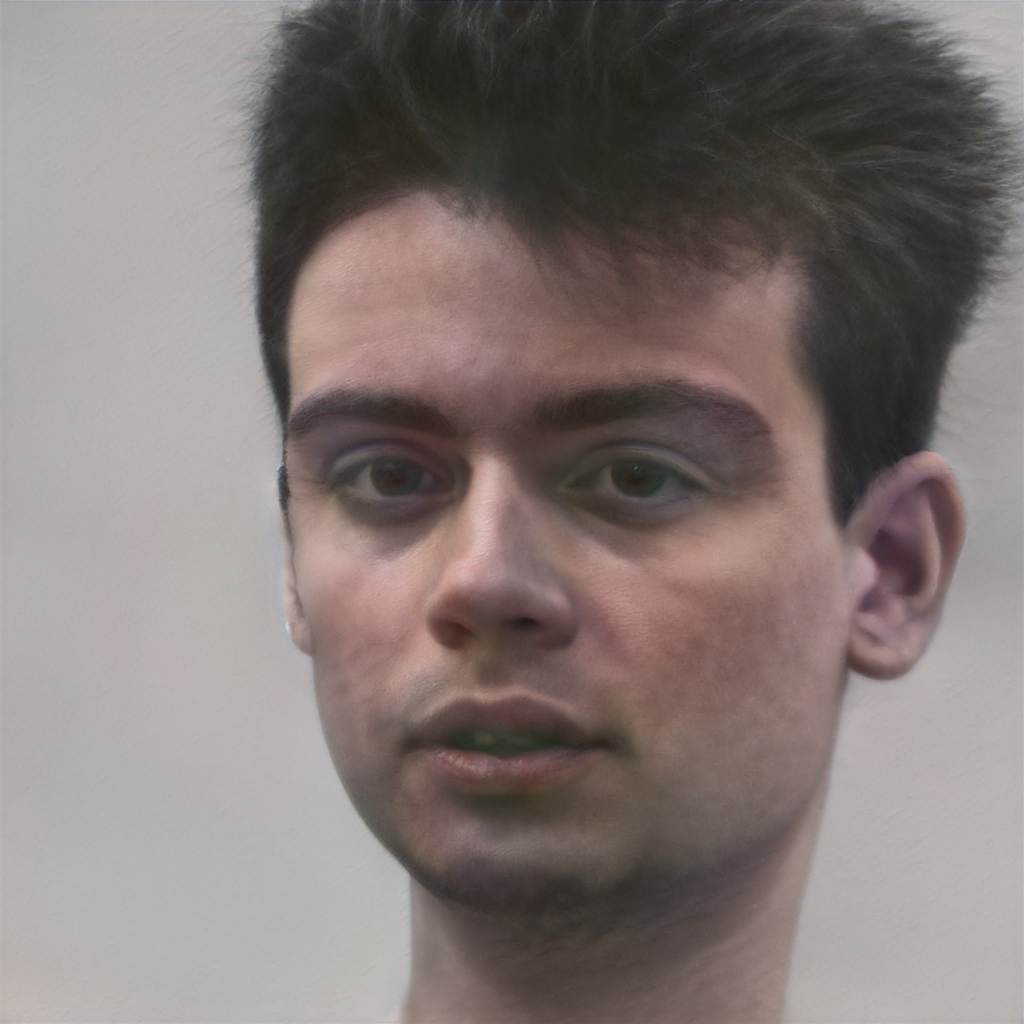}
\includegraphics[width=0.18\textwidth]{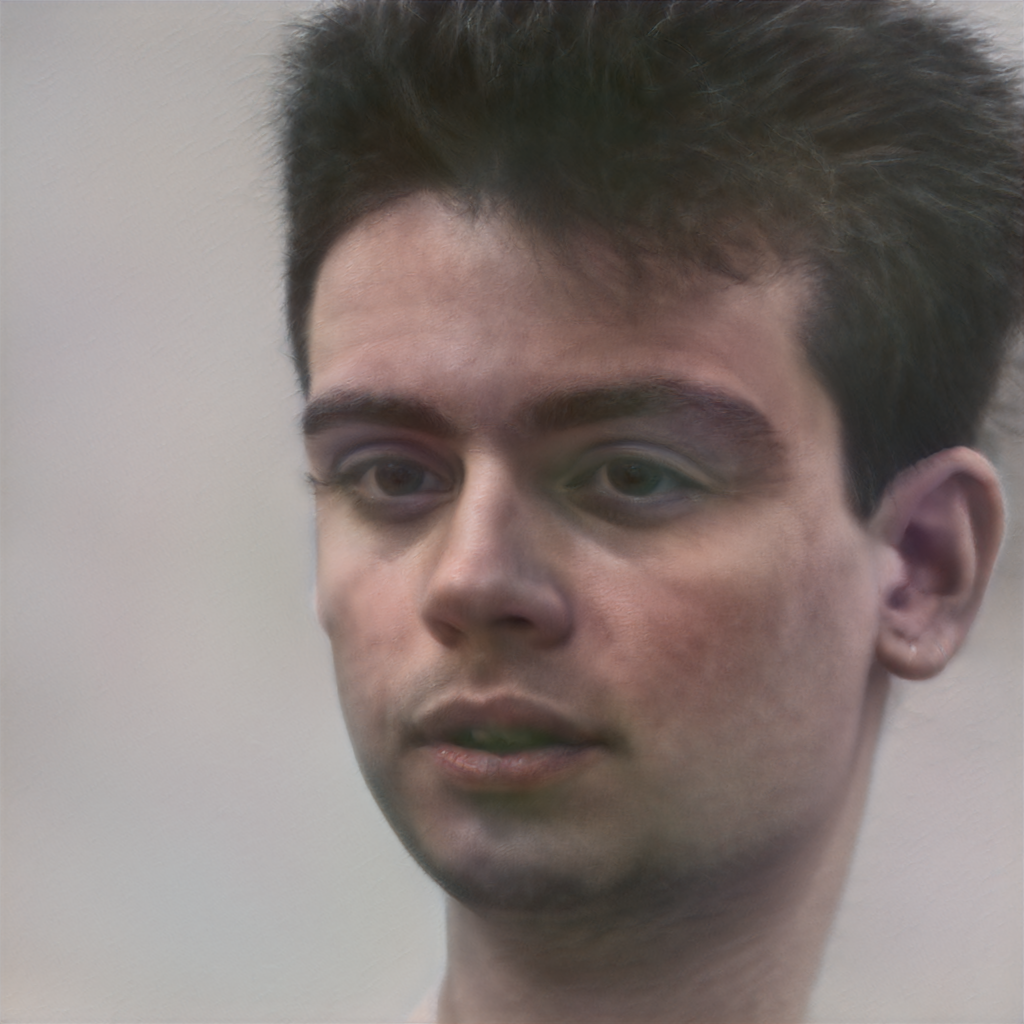}
\includegraphics[width=0.18\textwidth]{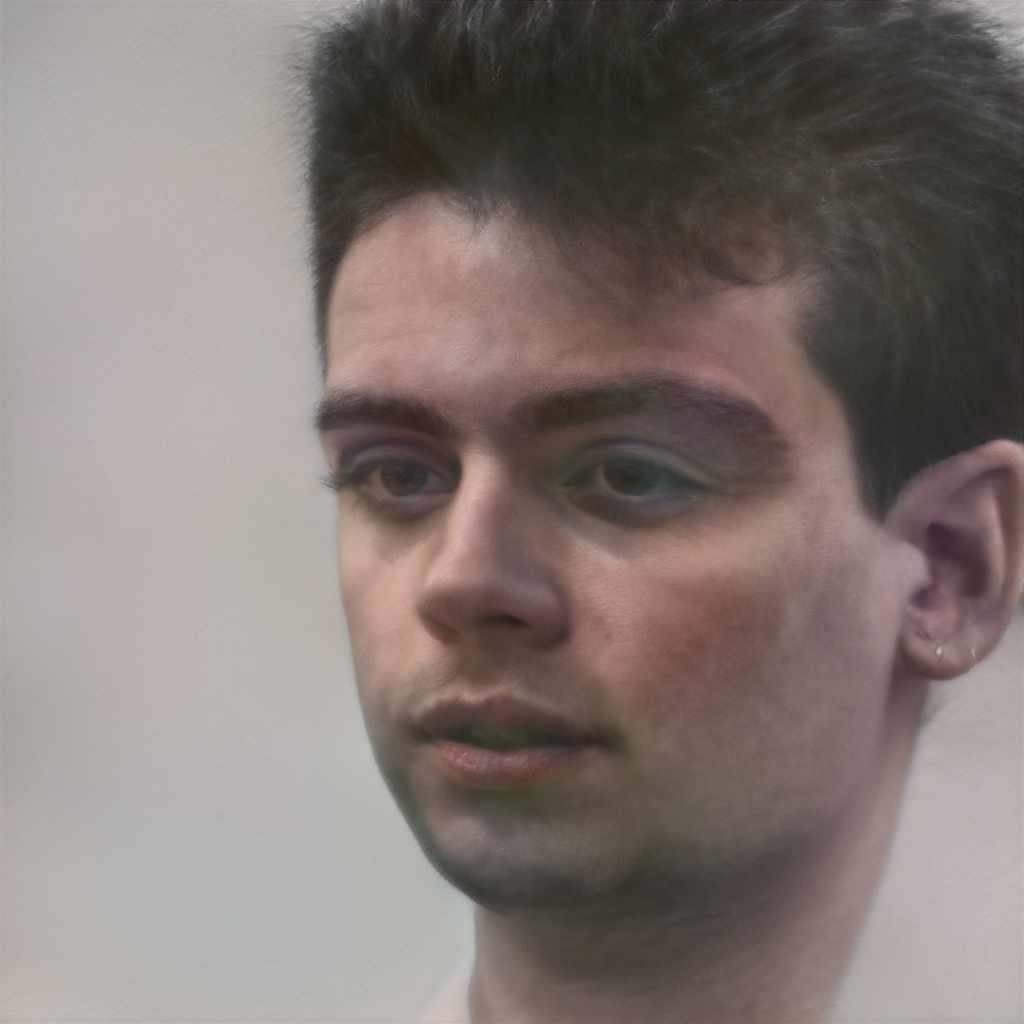}
\includegraphics[width=0.18\textwidth]{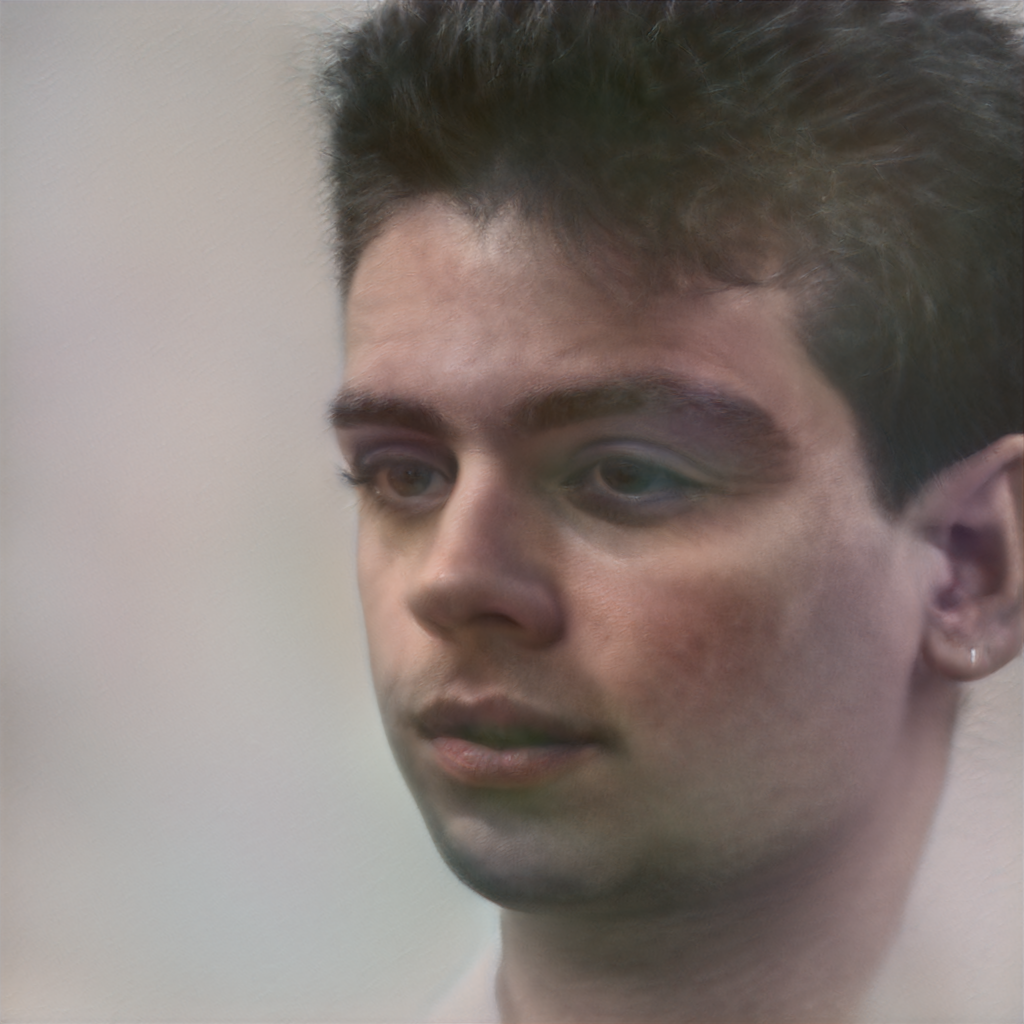}
\end{tabular}
\vspace{-0.7em}
\label{fig:micromotions:turning_head2}
\end{subfigure}
\caption{\textbf{Illustrations of the micromotion ``turning head'' founded by video-anchored method.} 
Best view when zoomed in. Please refer to our repository for complete video sequences. }
\label{fig:versatile_micromotions_head}
\end{figure*}

\subsection{Experiment Settings}\label{expSetting}
The pre-trained models, consist of StyleGAN-v2, StyleCLIP, and StyleGAN encoders, are all loaded from the publicly available repositories~\cite{abdal2020image2stylegan++,alaluf2021restyle,patashnik2021styleclip,radford2021learning}. When optimizing the latent codes, the learning rate was set to 0.1 and we adopted Adam optimizer. For the text-anchored and video-anchored methods, the numbers of latent codes we generate were 16 and 7 respectively. In robust PCA, 4 principal dimensions were chosen. We also searched the extrapolation scale hyperparameter $\alpha$ between 0.1 and 10. All the following results are generated at testing time, without any retraining. 

For the text-anchored experiments, the original images are generated using random latent codes in StyleGAN-v2 feature space. The text prompts we construct is in the general form of (a) ``A person with \{\} smile''; (b) ``A person with \{\} angry face''; (c) ``A person with eyes \{\} closed''; (d) ``\{\} old person with gray hair'', which correspond to the micromotions of smiling, angry, eyes opening and face aging. Here, the wildcard ``\{\}'' are replaced by a combination of both qualitative adjectives set including \{``no'', ``a big'', ``big'', ``a slight'', ``slight'', ``a large'', ``large'', `` ''\} and quantitative percentages set including \{10\%, ..., 90\%, 100\%\}. We will discuss the choice of various text templates and their outcomes in the ablation study. For the video-anchored experiments, we consider the micromotion of turning heads. The referential frames are collected from the Pointing04 DB dataset~\cite{gourier2004estimating}, and the frames we used for anchoring include a single identity with different postures, which has the angle of \{$-45\degree, -30\degree, -15\degree, 0\degree, 15\degree, 30\degree, 45\degree$\}.

\subsection{Micromotion Subspace Decoding}\label{exp1}
In this section, we consider both anchoring methods to decode the micromotion subspace from one single identity, and apply it to the in-domain identities to generate desired micromotions.

\paragraph{Text-anchored Reference Generation:}
Figure~\ref{fig:versatile_micromotions} shows the generated four micromotions via text prompts. Within each row, the five demonstrated frames are sampled from our synthesized continuous video with the desired micromotions. As we can see, all the results illustrate a continuous transition of one identity performing micromotions, which indicates the edit direction decoded from the micromotion subspace is meaningful and semantically correct. It is worth noting that the micromotion space is extremely low-rank since only 4 principal dimensions are used. The smooth edit direction from the low-rank space verifies our first hypothesis, that the micromotions can indeed be represented in low-dimensional space.

\paragraph{Video-anchored Reference Generation:}
Figure~\ref{fig:versatile_micromotions_head} shows the generated turning head micromotion via reference videos. Similar to the text-anchored method, the five frames are also sampled from the video synthesized by our workflow. From the results, we can observe that although with small deformation and artifacts, the synthesized frames also formulate a continuous transition of the person turning around his head, and such micromotion can also be decoded from low-dimensional micromotion space. Therefore, we conclude that the video-anchored method also effectively anchors the low-rank space and helps to decode the micromotion successfully.

\subsection{Micromotion Applications on Cross-domain Identities} 
\begin{figure*}[!htb]
\vspace{0.2em}
\centering
\begin{subfigure}{\textwidth}
\centering
\begin{tabular}{ccccc}
\includegraphics[width=0.18\textwidth]{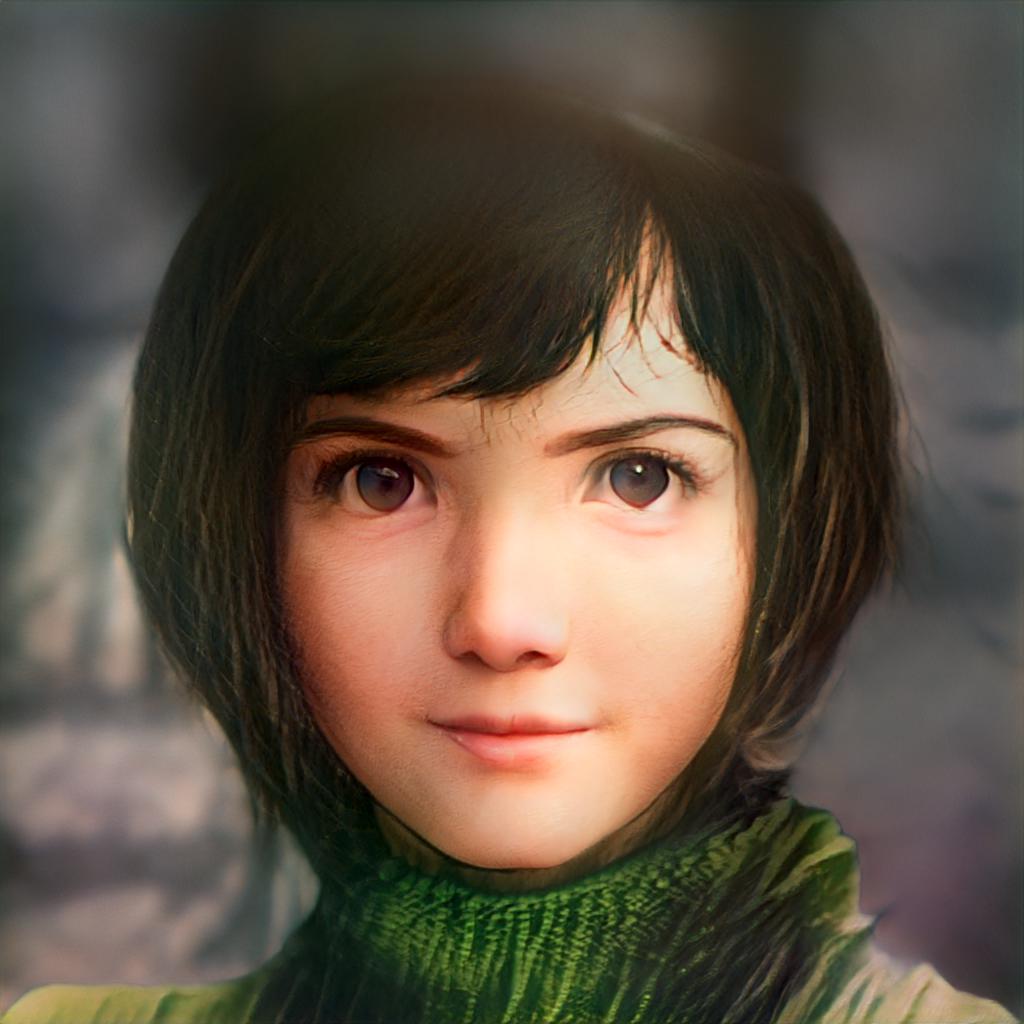}
\includegraphics[width=0.18\textwidth]{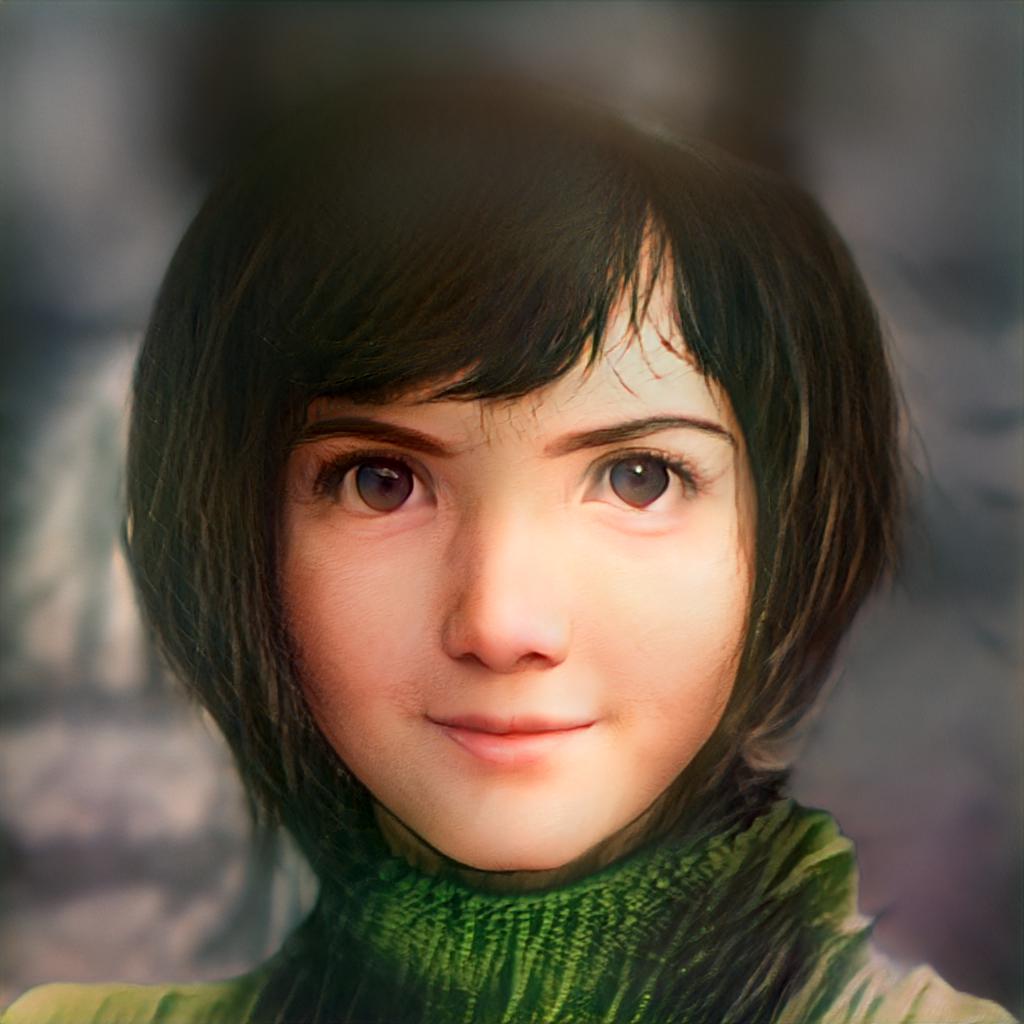}
\includegraphics[width=0.18\textwidth]{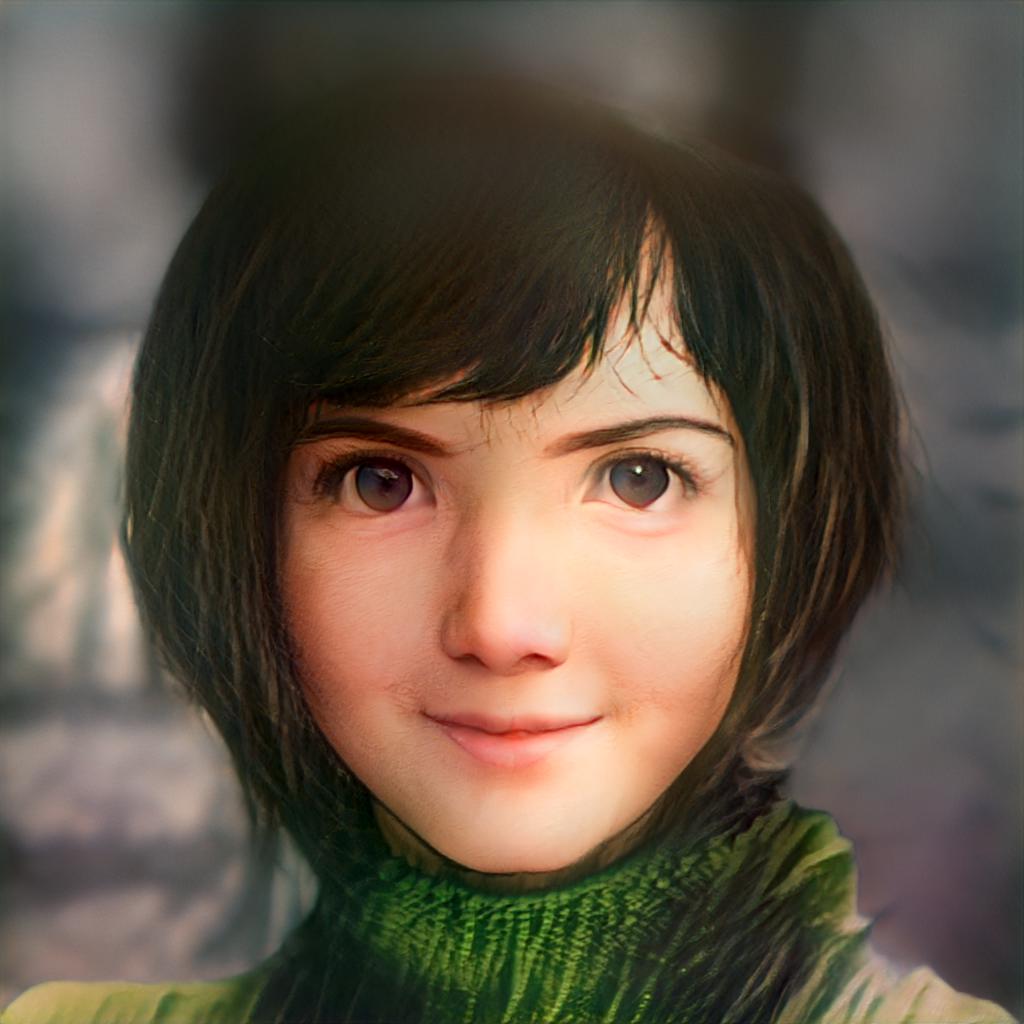}
\includegraphics[width=0.18\textwidth]{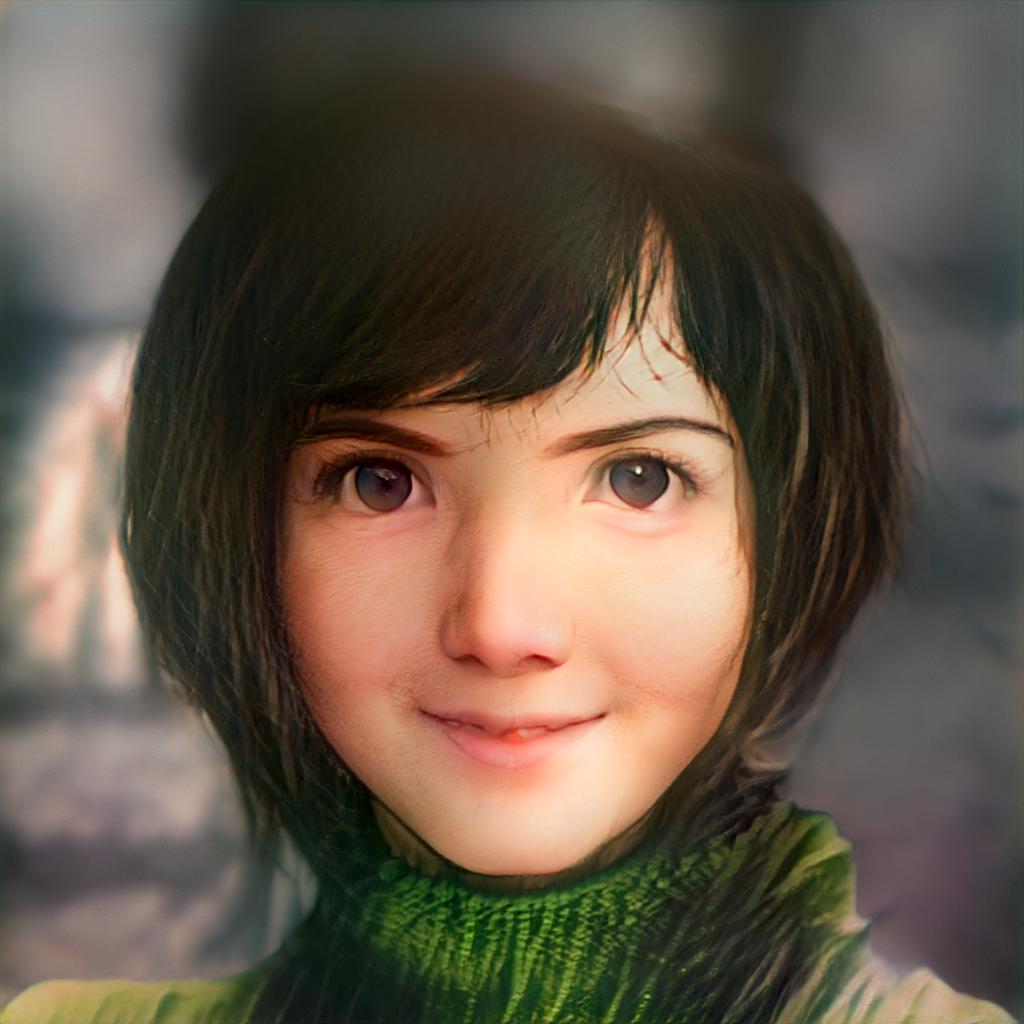}
\includegraphics[width=0.18\textwidth]{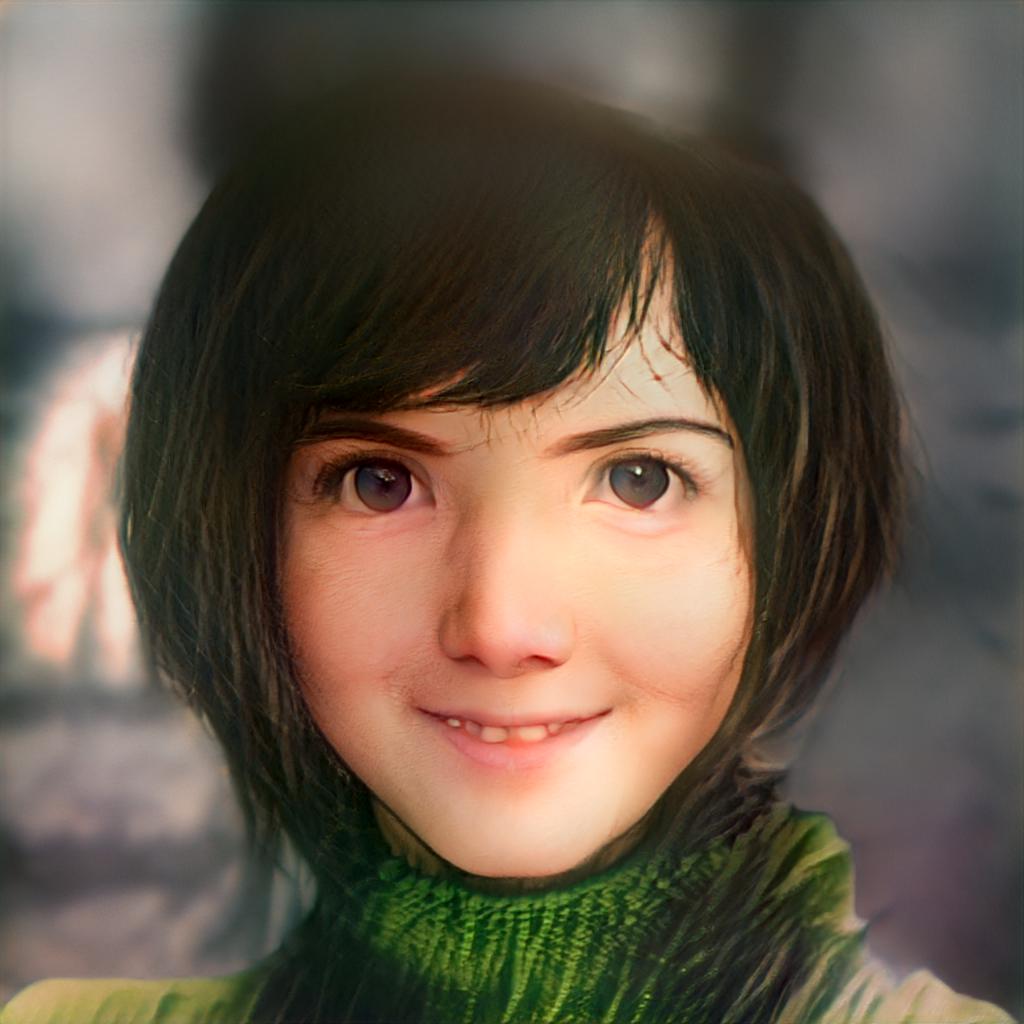}
\end{tabular}
\vspace{-0.8em}
\caption{Anime Character, Smiling}
\end{subfigure}
\begin{subfigure}{\textwidth}
\centering
\begin{tabular}{ccccc}
\includegraphics[width=0.18\textwidth]{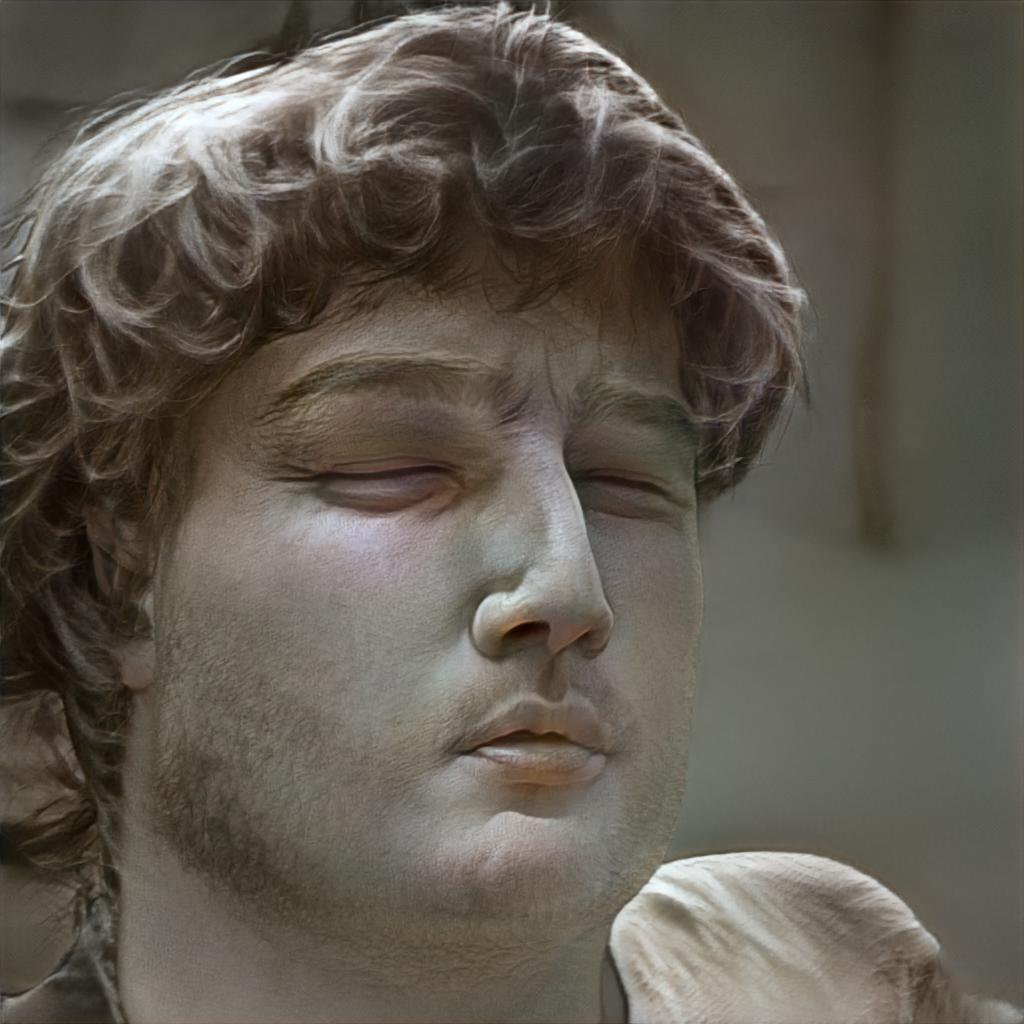}
\includegraphics[width=0.18\textwidth]{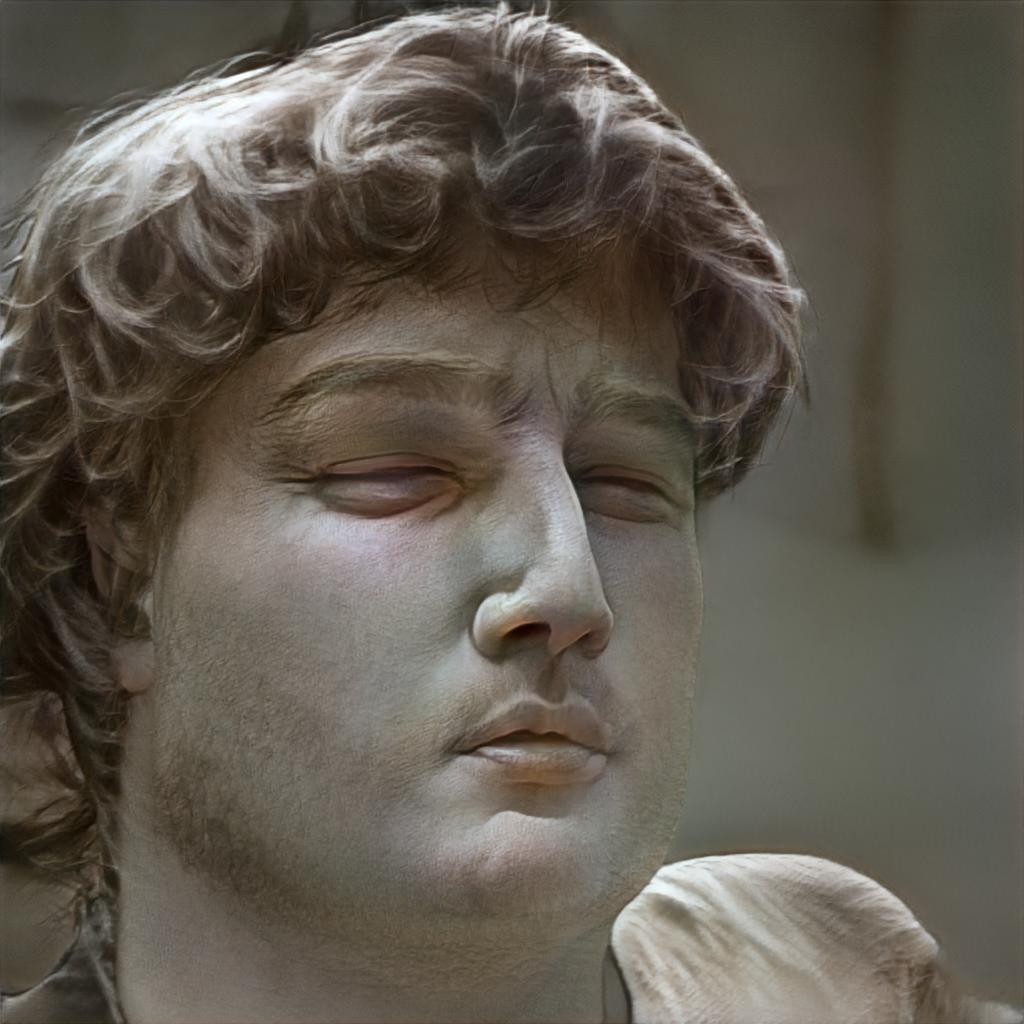}
\includegraphics[width=0.18\textwidth]{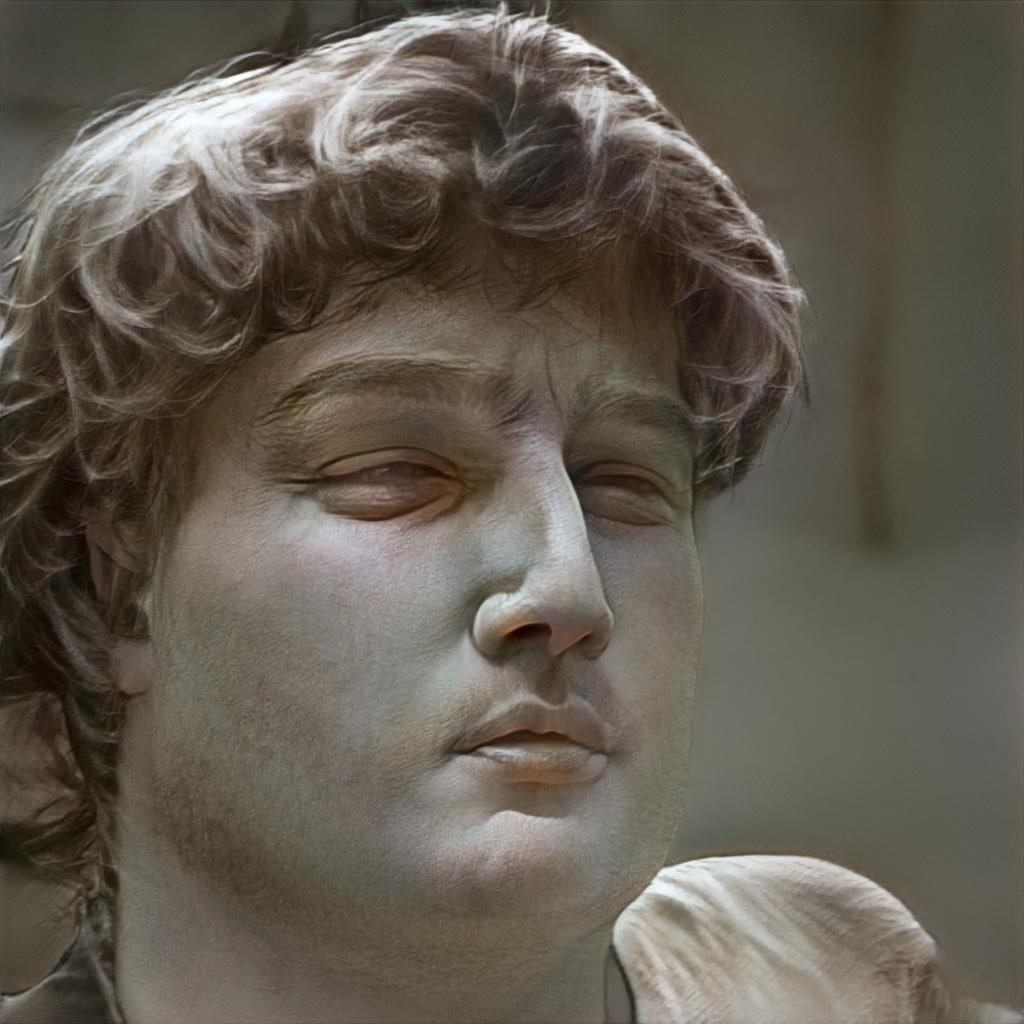}
\includegraphics[width=0.18\textwidth]{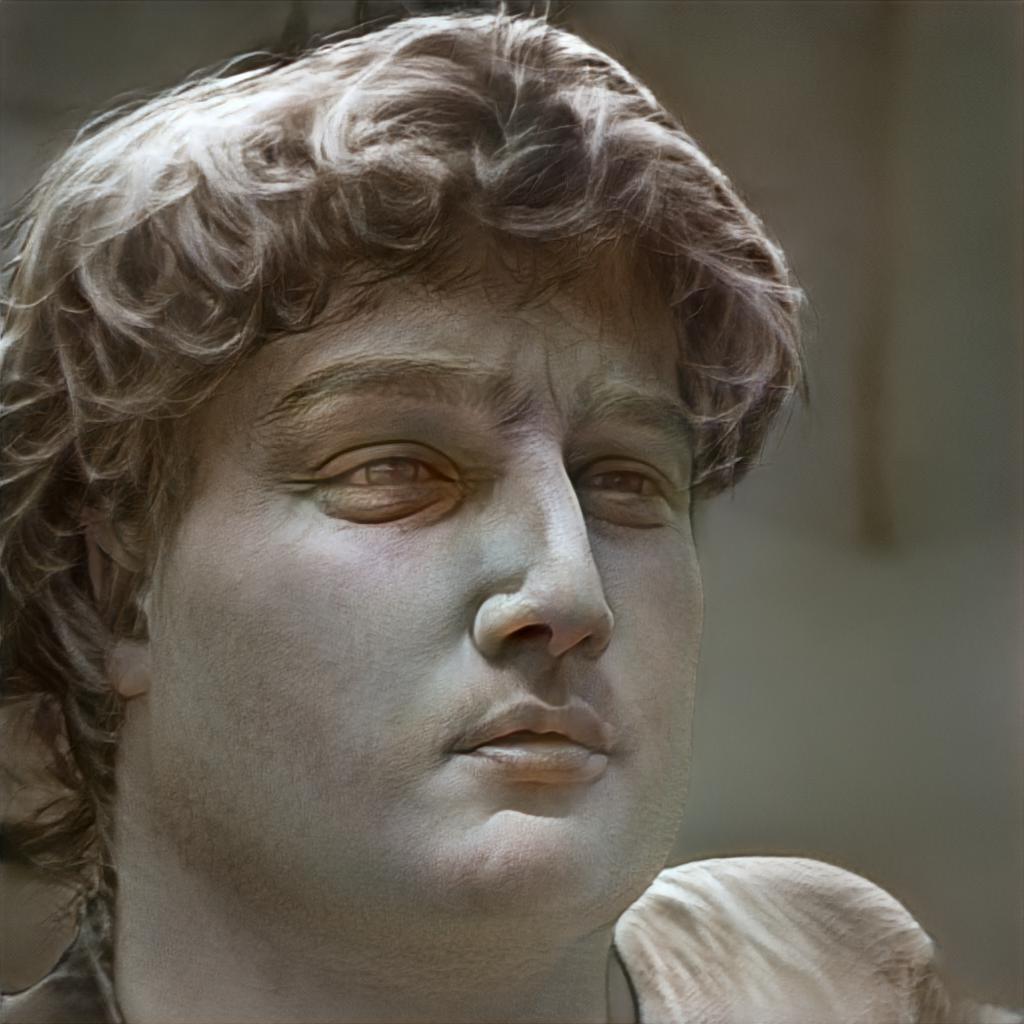}
\includegraphics[width=0.18\textwidth]{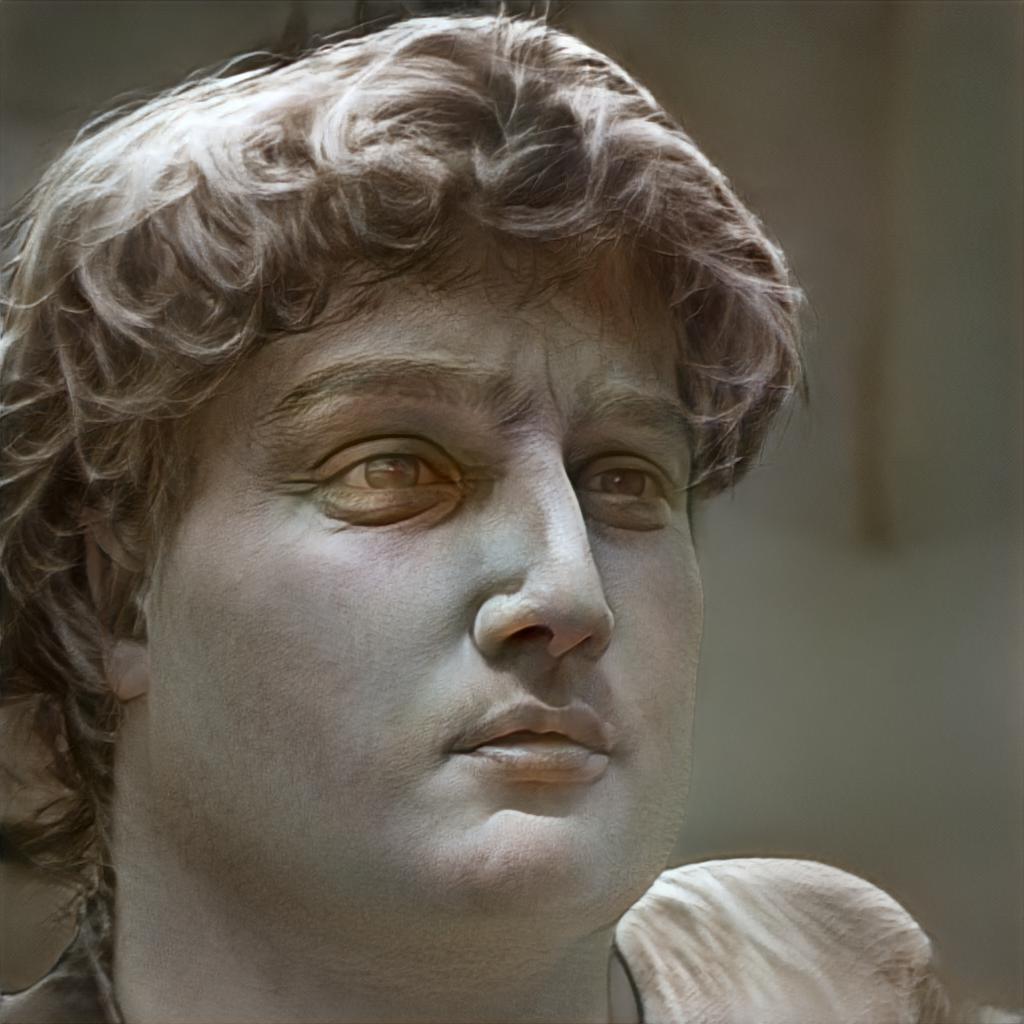}
\end{tabular}
\vspace{-0.8em}
\caption{Marble Sculpture, Opening Eyes}
\end{subfigure}

\begin{subfigure}{\textwidth}
\centering
\begin{tabular}{ccccc}
\includegraphics[width=0.18\textwidth]{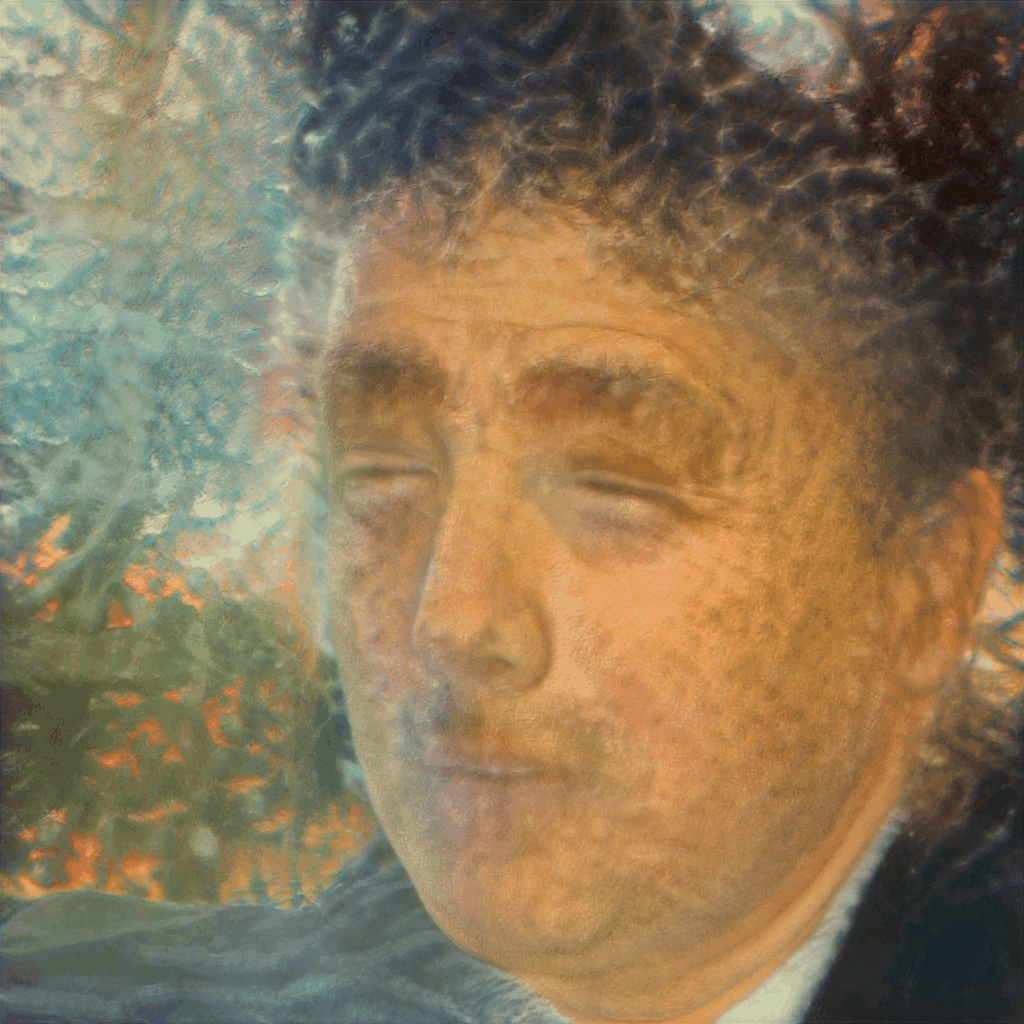}
\includegraphics[width=0.18\textwidth]{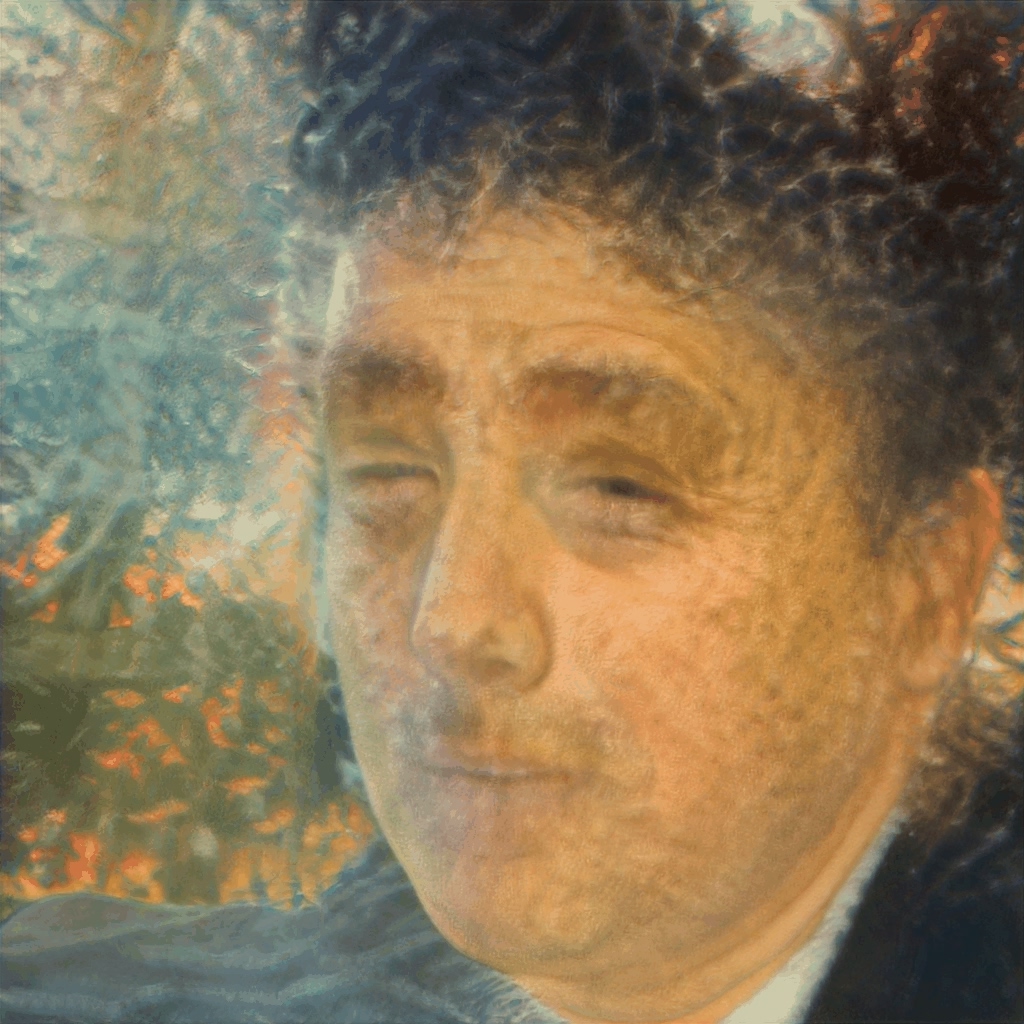}
\includegraphics[width=0.18\textwidth]{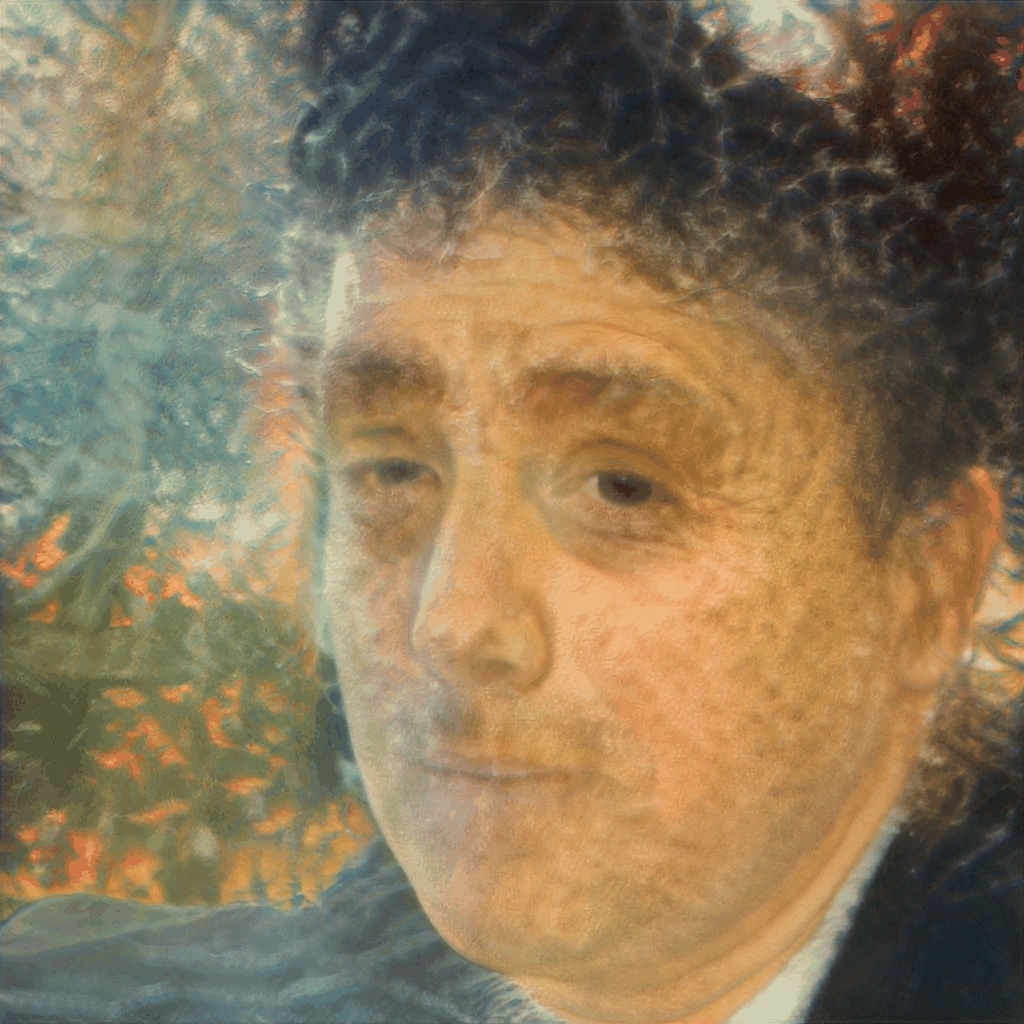}
\includegraphics[width=0.18\textwidth]{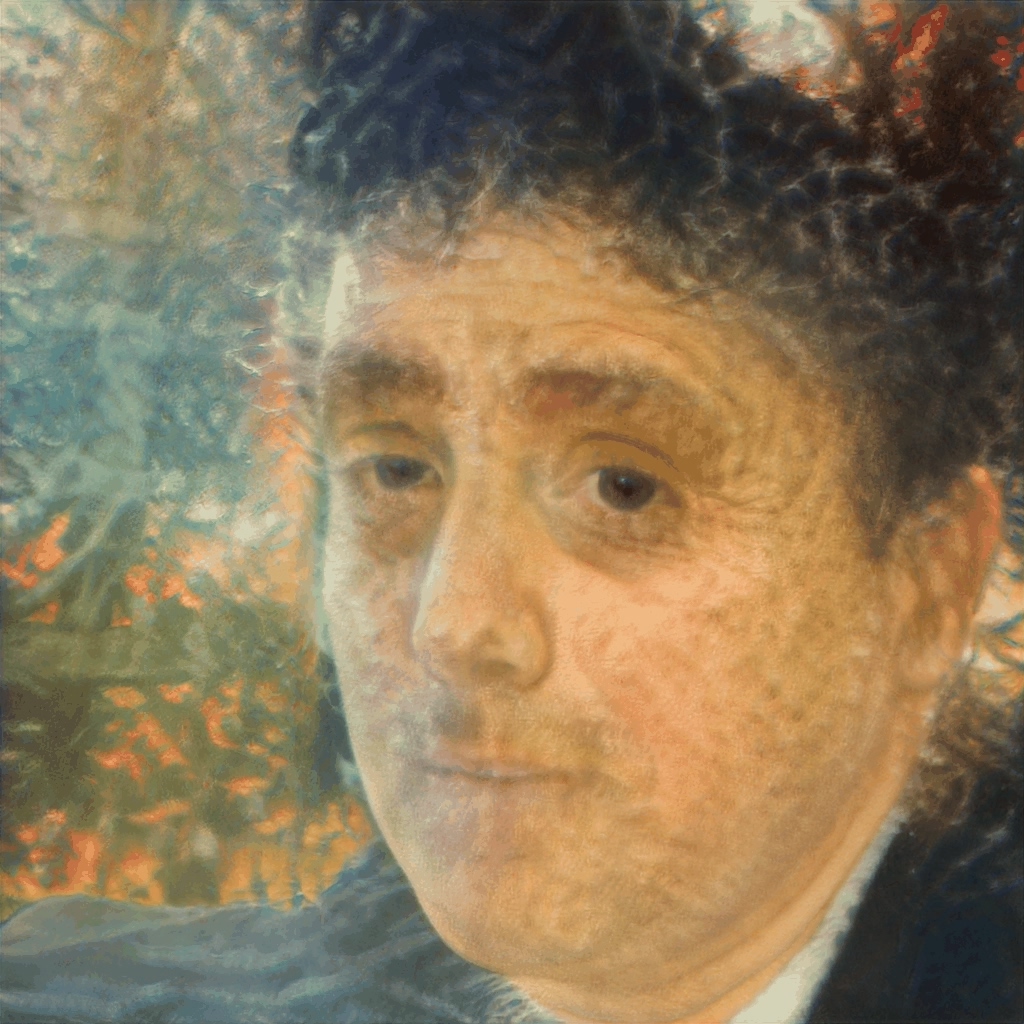}
\includegraphics[width=0.18\textwidth]{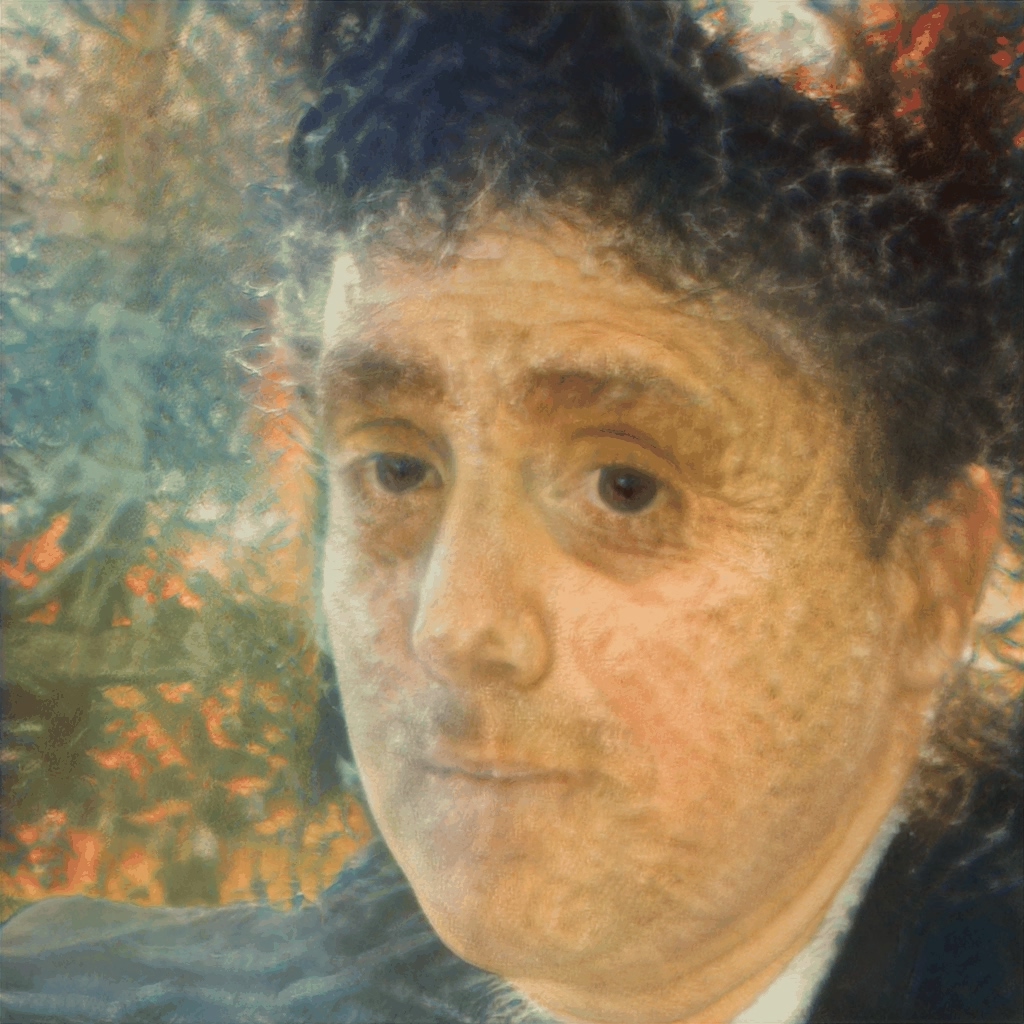}
\end{tabular}
\vspace{-0.8em}
\caption{Painting: Monet, Opening Eyes}
\end{subfigure}
\begin{subfigure}{\textwidth}
\centering
\begin{tabular}{ccccc}
\includegraphics[width=0.18\textwidth]{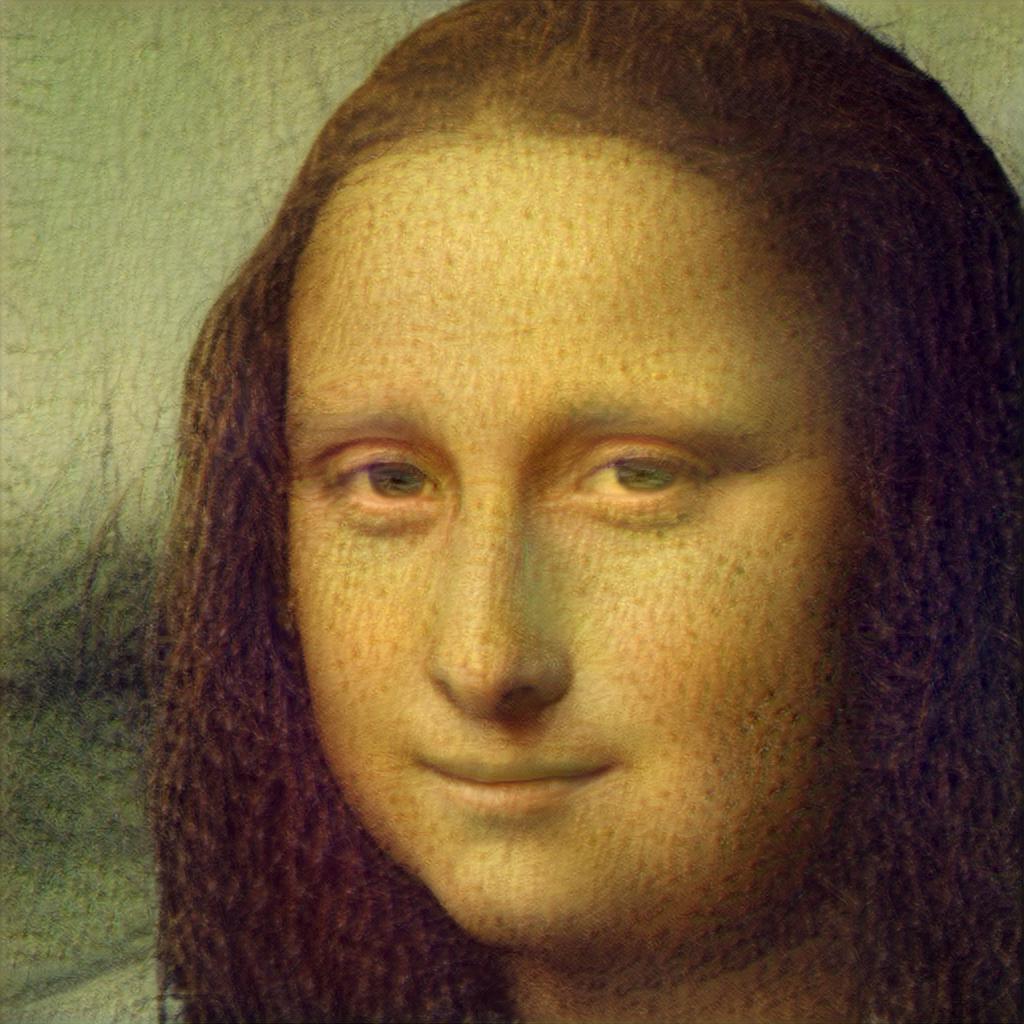}
\includegraphics[width=0.18\textwidth]{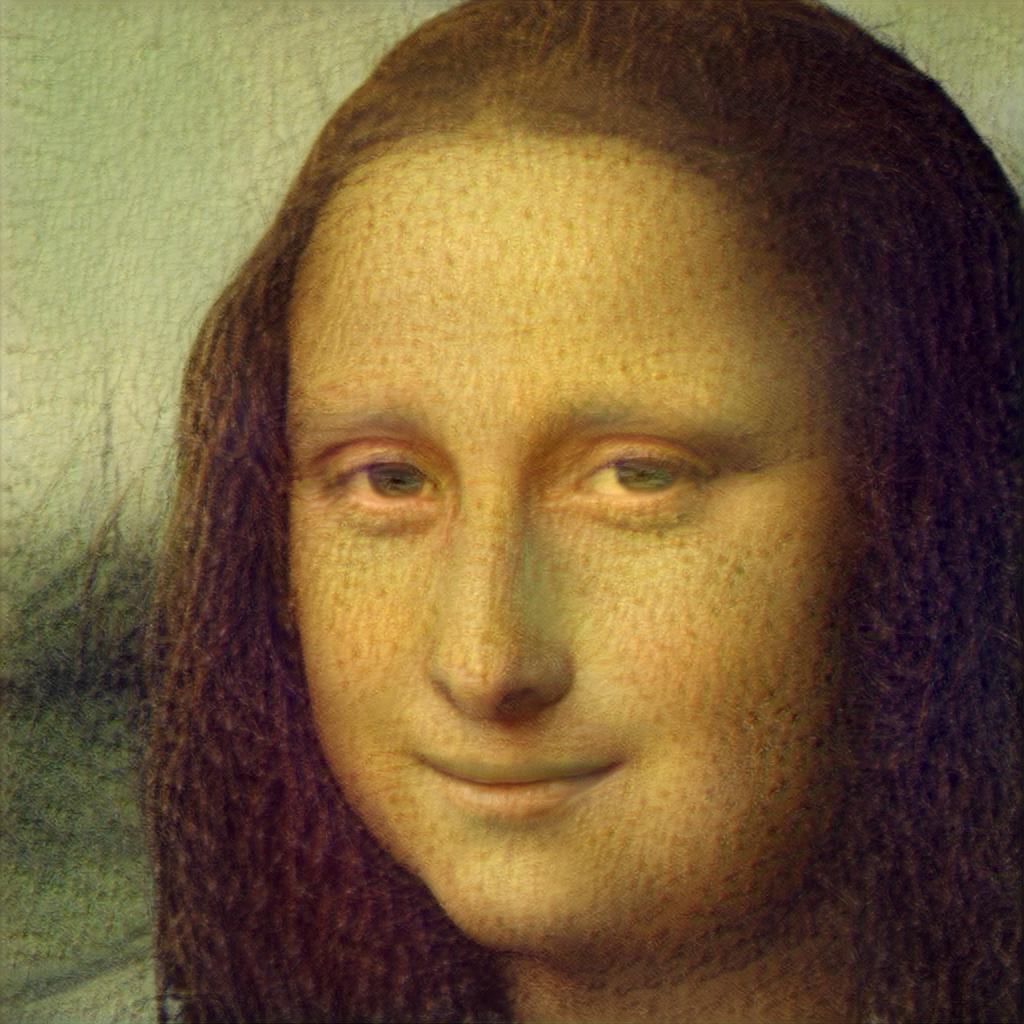}
\includegraphics[width=0.18\textwidth]{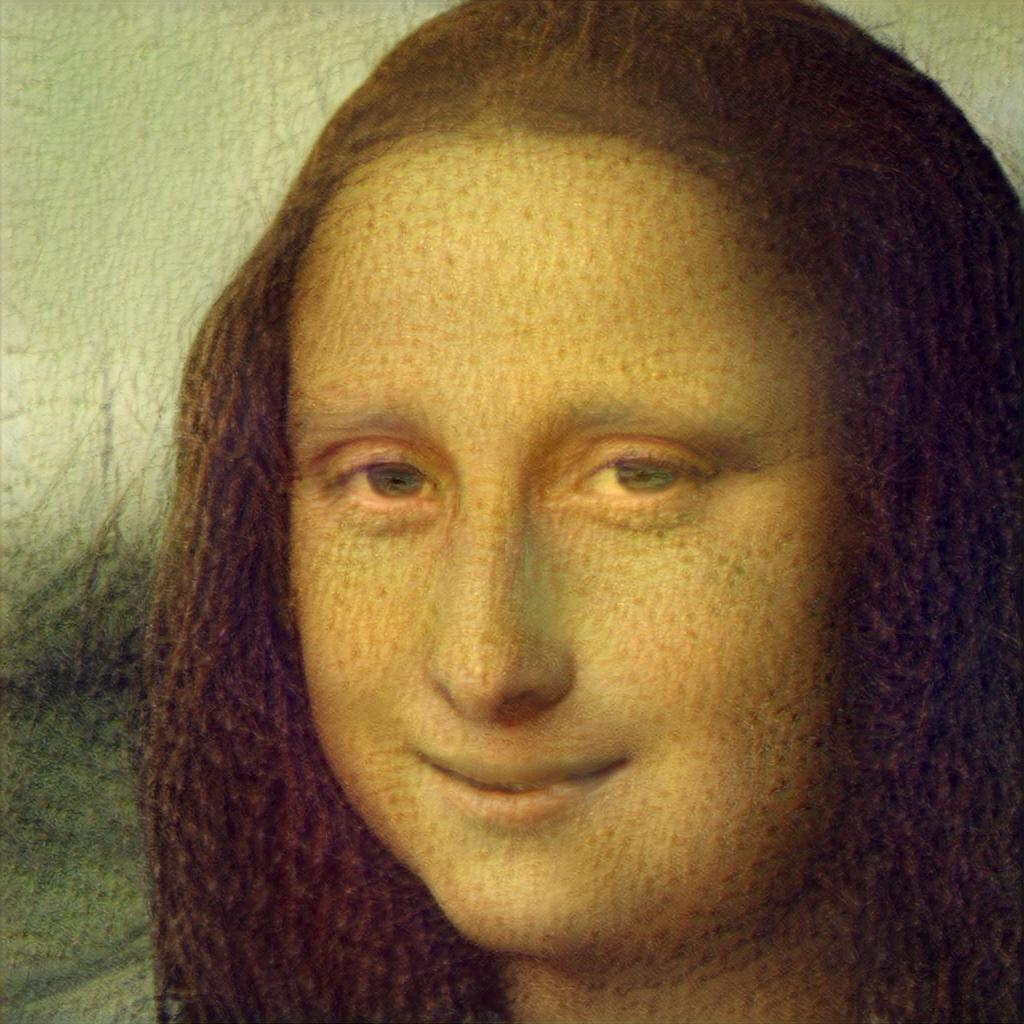}
\includegraphics[width=0.18\textwidth]{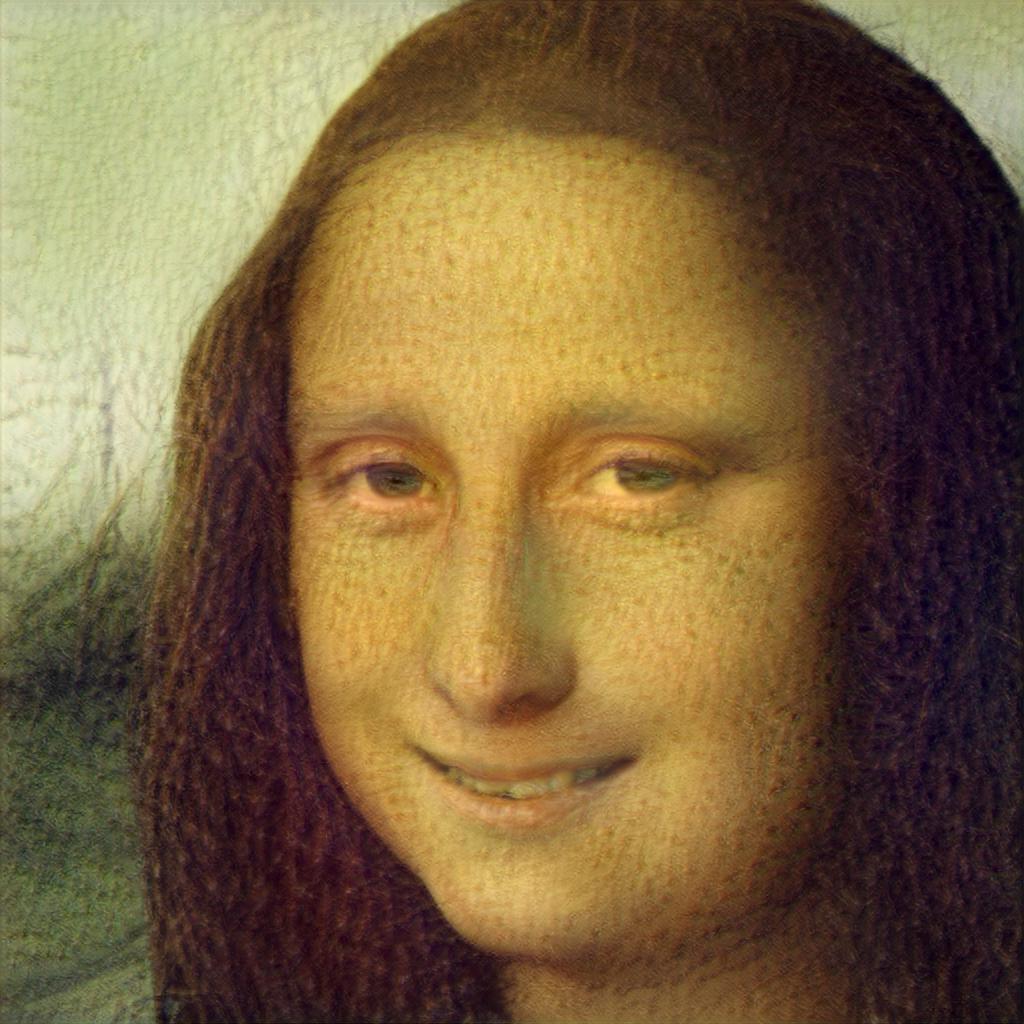}
\includegraphics[width=0.18\textwidth]{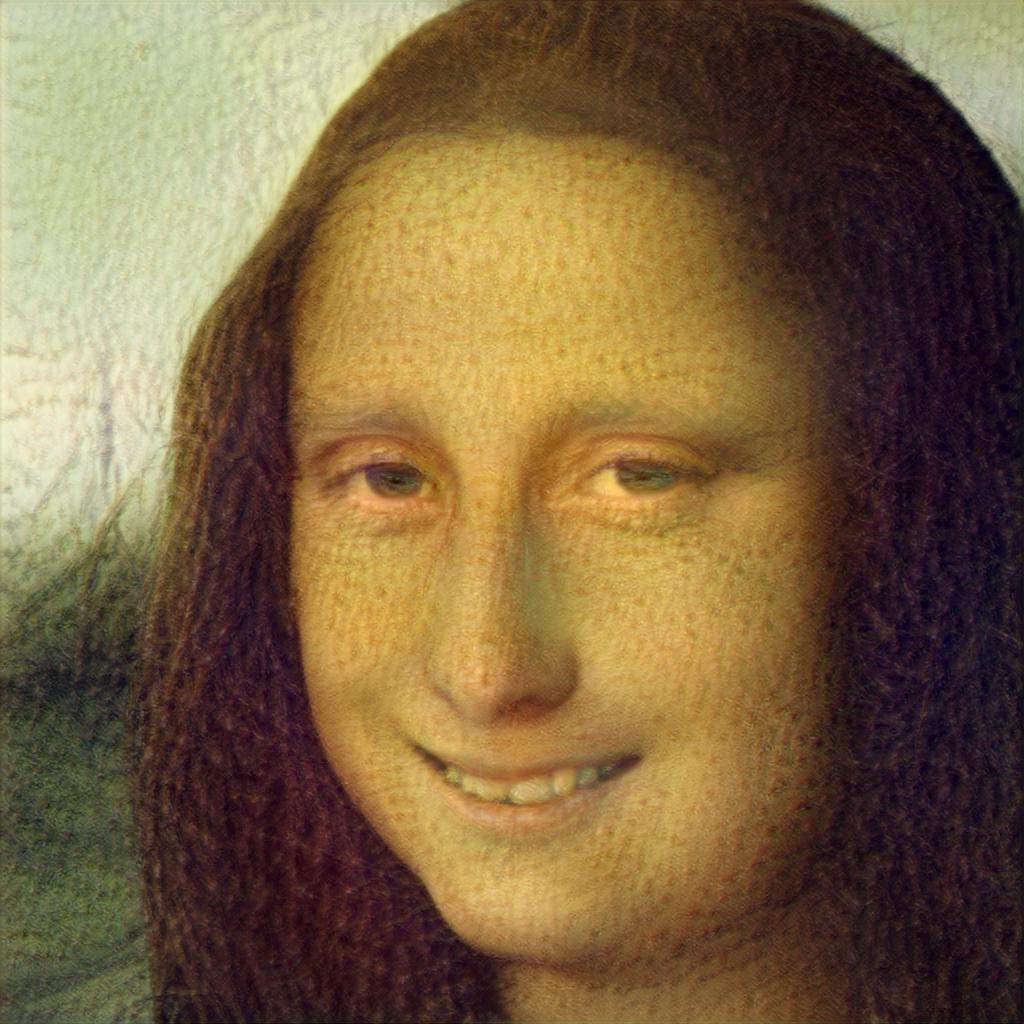}
\end{tabular}
\vspace{-0.8em}
\caption{Painting: Da Vinci, Smiling}
\end{subfigure}
\begin{subfigure}{\textwidth}
\centering
\begin{tabular}{ccccc}
\includegraphics[width=0.18\textwidth]{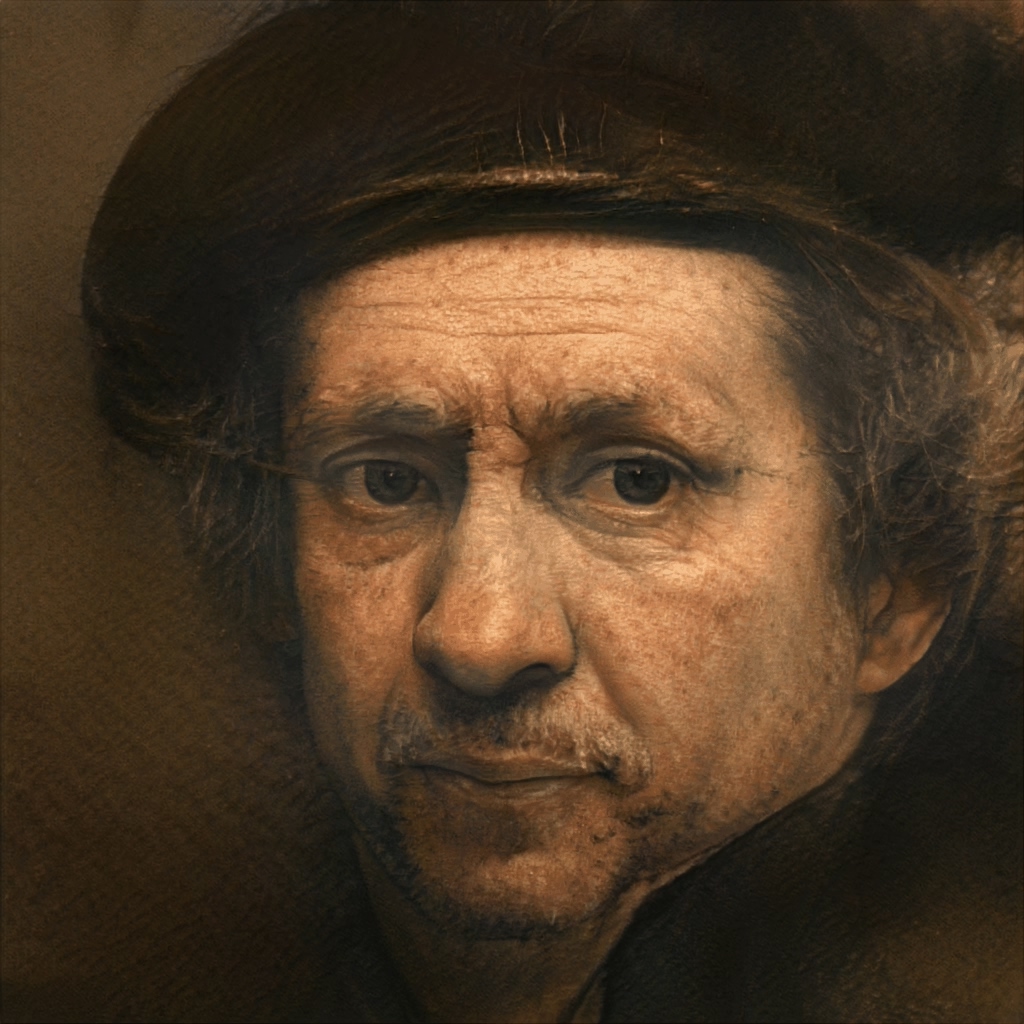}
\includegraphics[width=0.18\textwidth]{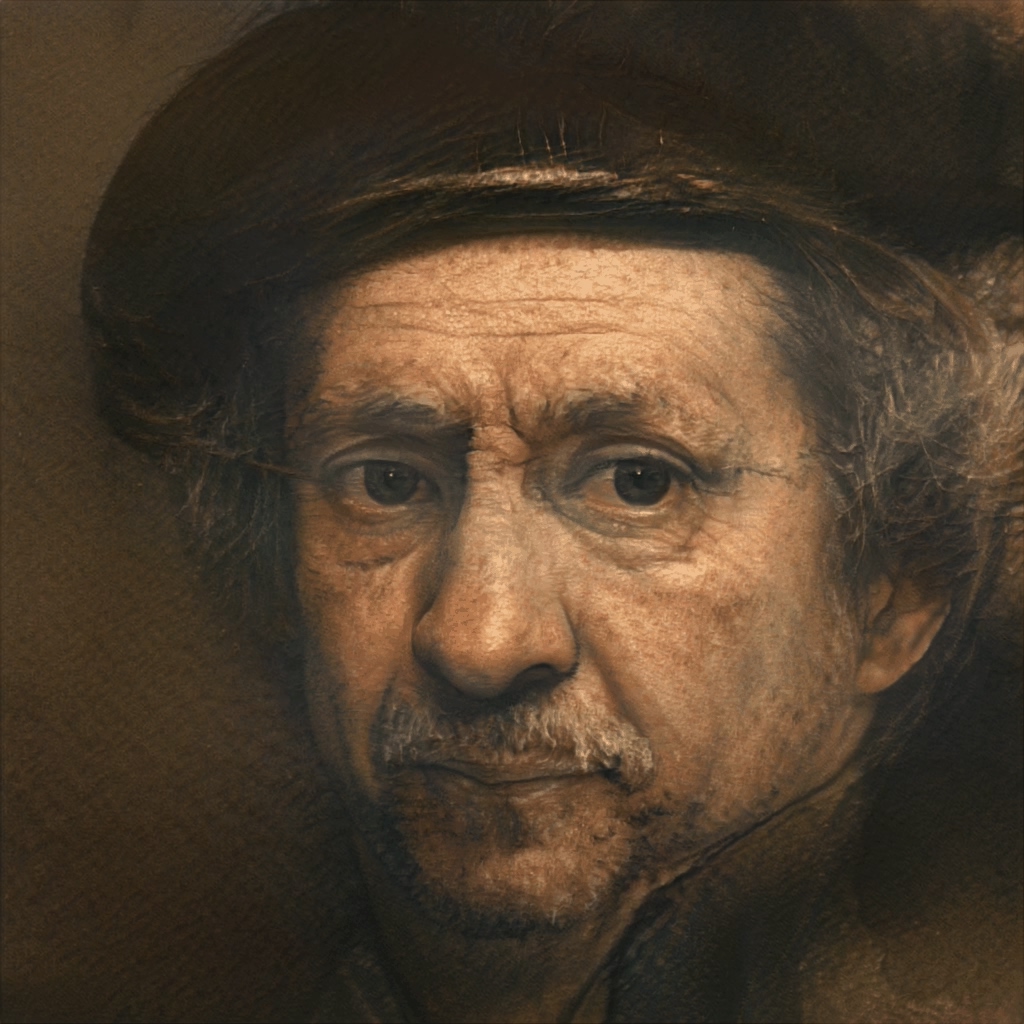}
\includegraphics[width=0.18\textwidth]{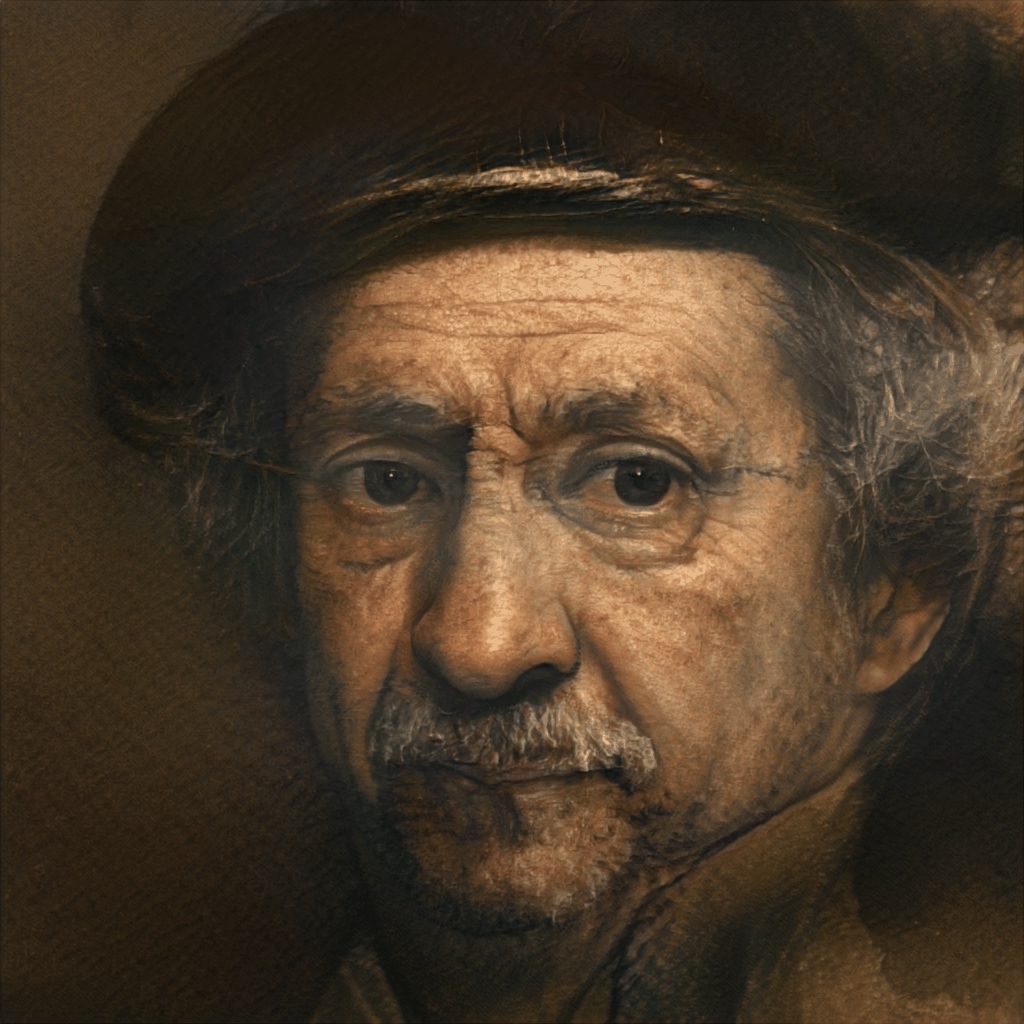}
\includegraphics[width=0.18\textwidth]{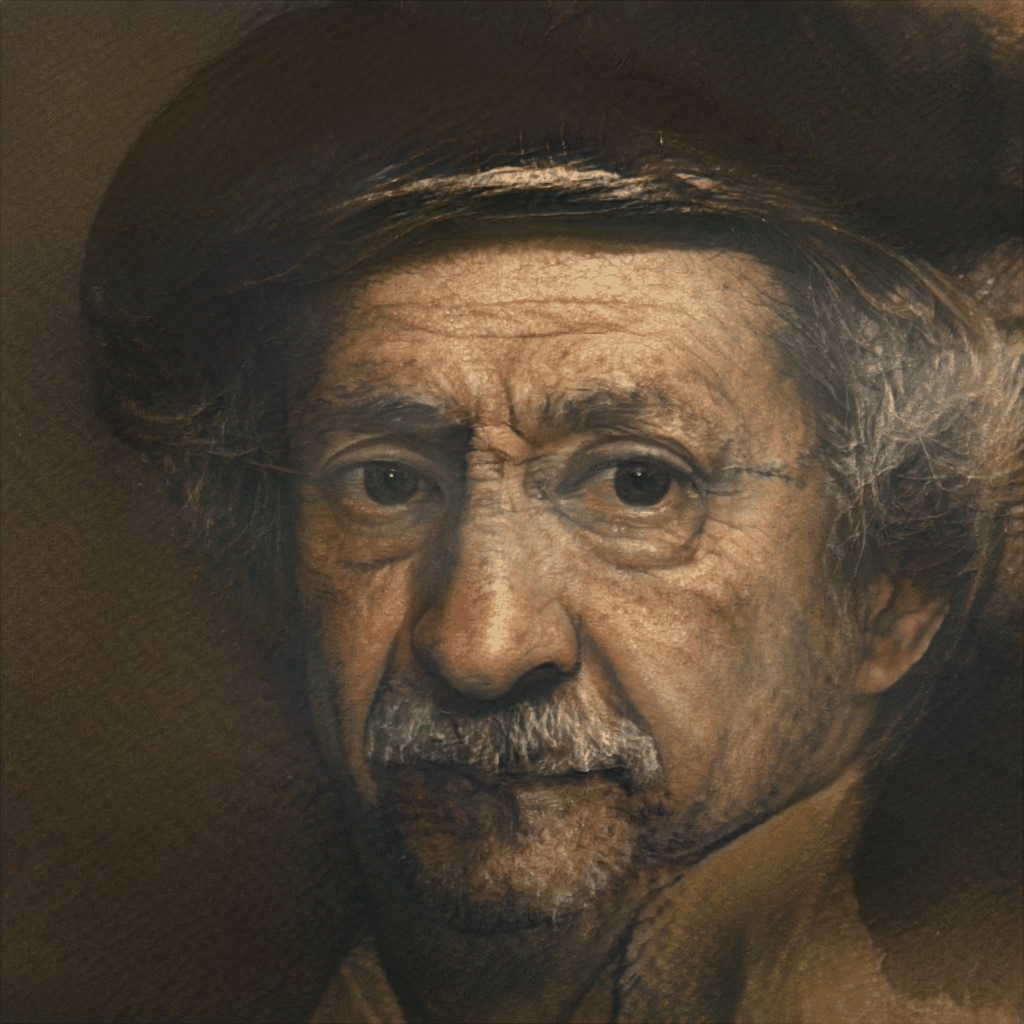}
\includegraphics[width=0.18\textwidth]{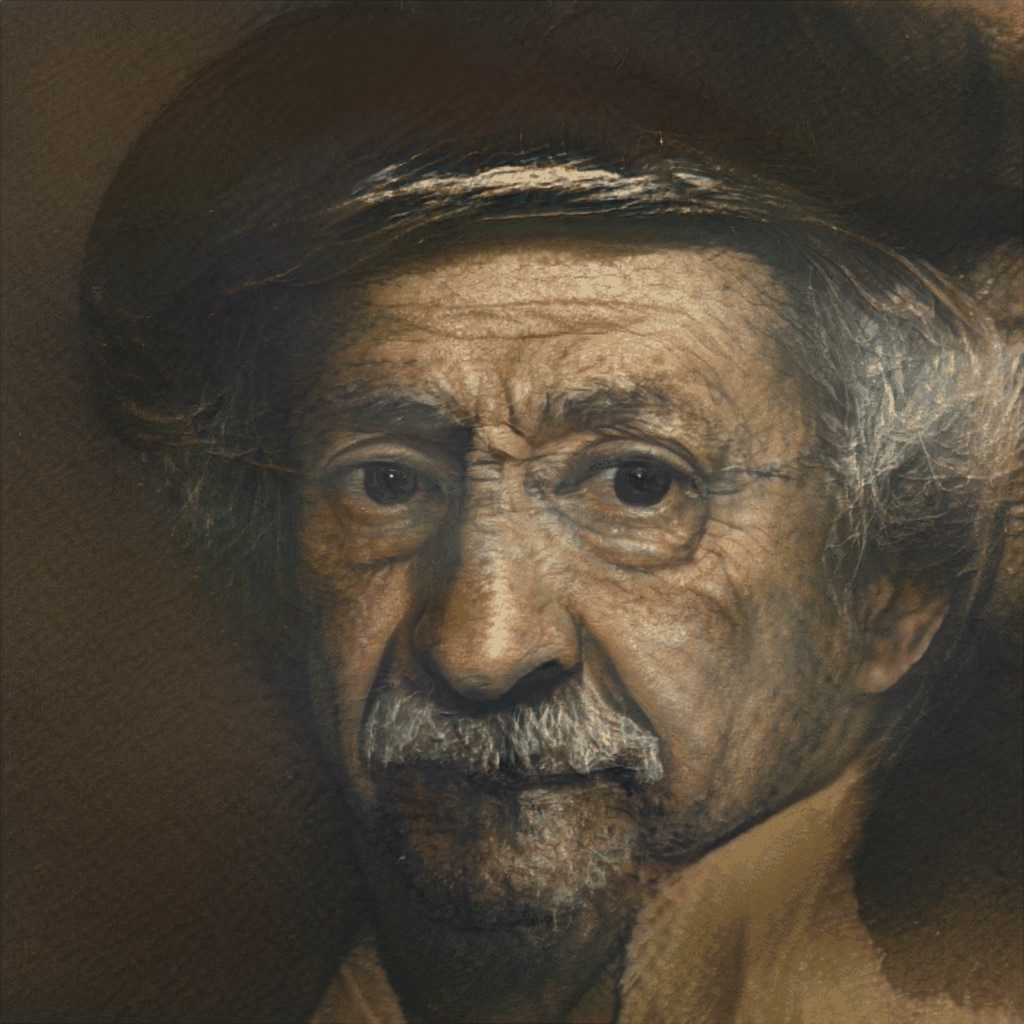}
\end{tabular}
\vspace{-0.8em}
\caption{Painting: Rembrandt, Aging face}
\end{subfigure}
\begin{subfigure}{\textwidth}
\centering
\begin{tabular}{ccccc}
\includegraphics[width=0.18\textwidth]{figures/teaser-rotate/00.jpg}
\includegraphics[width=0.18\textwidth]{figures/teaser-rotate/20.jpg}
\includegraphics[width=0.18\textwidth]{figures/teaser-rotate/42.jpg}
\includegraphics[width=0.18\textwidth]{figures/teaser-rotate/64.jpg}
\includegraphics[width=0.18\textwidth]{figures/teaser-rotate/88.jpg}
\end{tabular}
\vspace{-0.8em}
\caption{Painting: Van Gogh, Turning Head}
\end{subfigure}

\vspace{-0.6em}
\caption{\textbf{Micromotions on cross-domain identities.} Our micromotions generalize well when transferred to novel domains, including anime characters, marble sculptures, and various genres of paintings (Van Gogh, Monet, Da Vinci, Rembrandt). Best view when zoomed in. Please refer to our repository for complete video sequences. }
\label{fig:micromotions:transfer}
\end{figure*}
Sec.\ref{exp1} decodes the micromotion from low-dimensional micromotion subspace, which verifies the first part of the hypothesis. In this section, we further verify the second part of the hypothesis, exploring if the decoded micromotion can be applied to arbitrary and cross-domain identities.

Figure~\ref{fig:micromotions:transfer} shows the result of transferring the decoded micromotions on novel identities. Within each row, we exert the decoded micromotions on the novel identities, synthesize the desired movements, and demonstrate sampled frames from the generated continuous videos. From these results, we can observe that the sampled frames on each new identity also depict the continuous transitions of desired micromotions. This verifies that the decoded micromotions extracted from our workflow can be successfully transited to the out-domain identities, generating smooth and natural transformations. Furthermore, this phenomenon verifies the second part of the hypothesis: The low-dimensional micromotion subspace in StyleGAN are not isolated nor tied with certain identities. On the contrary, in StyleGAN latent space, the \textit{identity-agnostic} micromotions can indeed be represented as a low-rank space found in our workflow disentangled from various identities. As such, the decoded micromotion can be ubiquitously applied to those even out-of-domain identities.

Moreover, we emphasize that to generate dynamic micromotion on a novel identity, the entire computational cost boils down to inverting the identity into latent space and then extrapolating along the obtained edit direction, without the requirement of retraining the model or conducting identity-specific computations. Therefore, that enables effortless editing of new identity images using the found direction, with little extra cost.

\begin{figure*}[!t]
\centering
\begin{subfigure}{\textwidth}
\centering
\begin{tabular}{ccccc}
\includegraphics[width=0.18\textwidth]{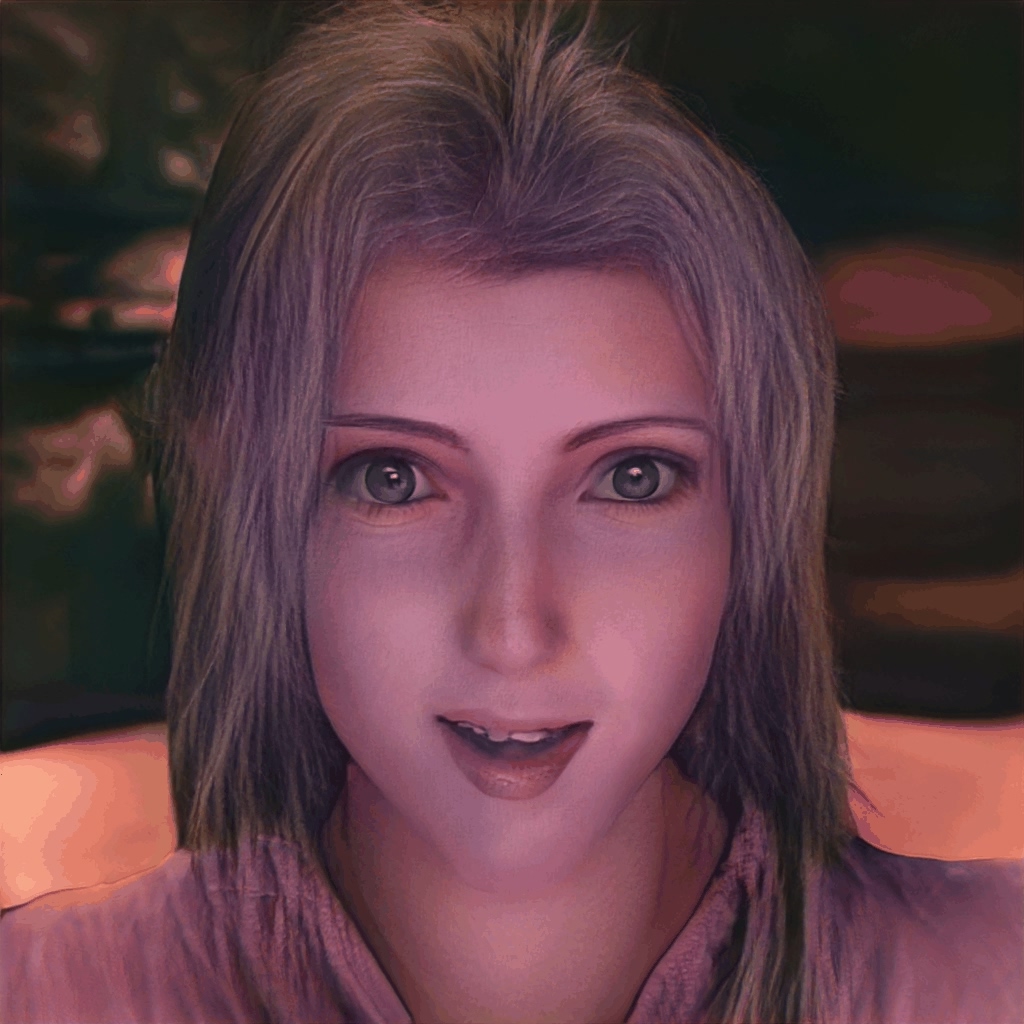}
\includegraphics[width=0.18\textwidth]{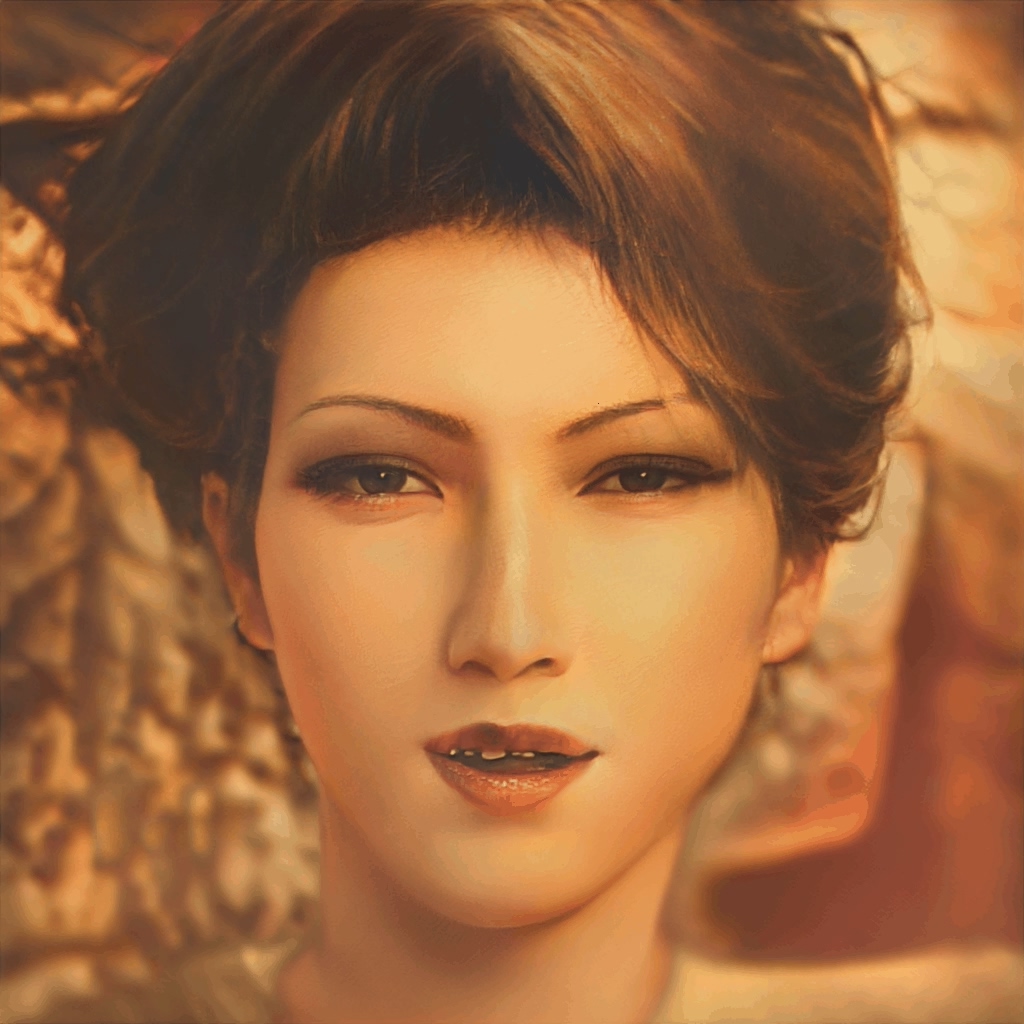}
\includegraphics[width=0.18\textwidth]{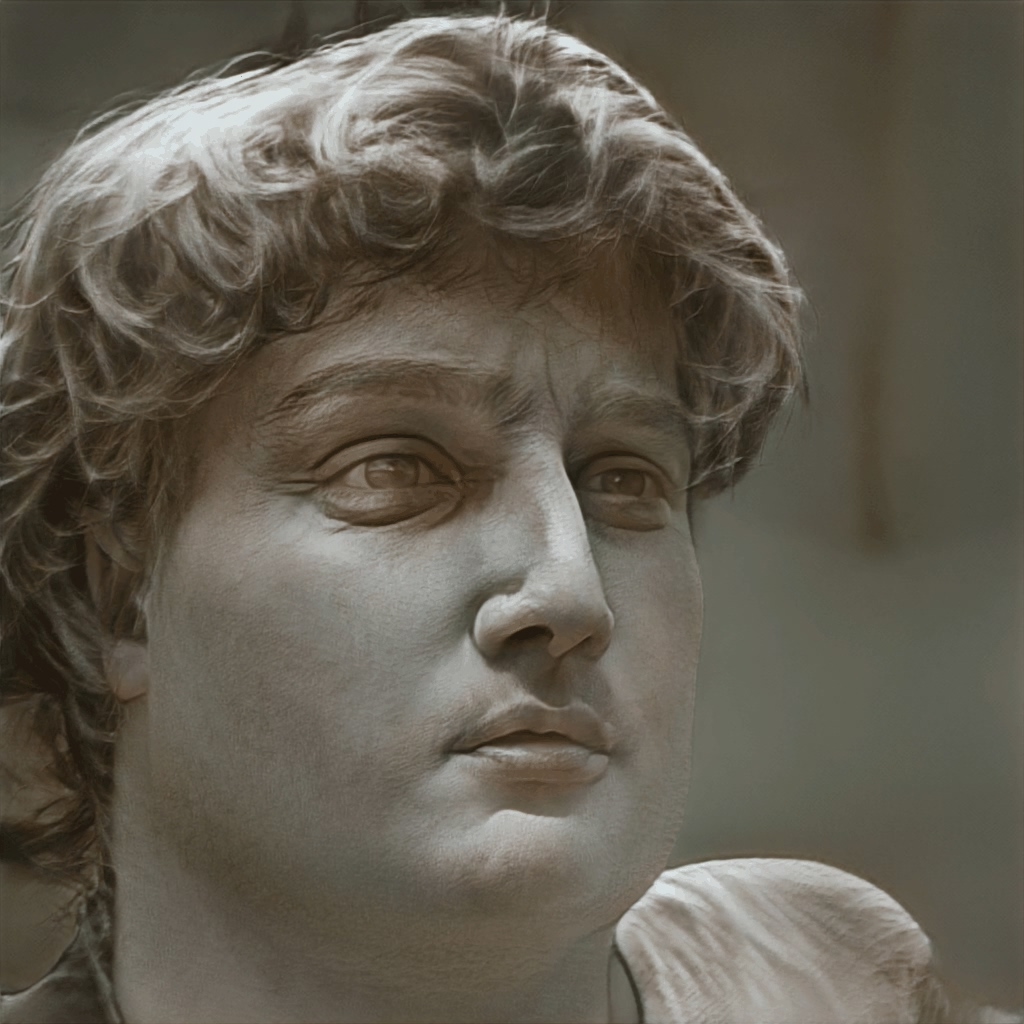}
\includegraphics[width=0.18\textwidth]{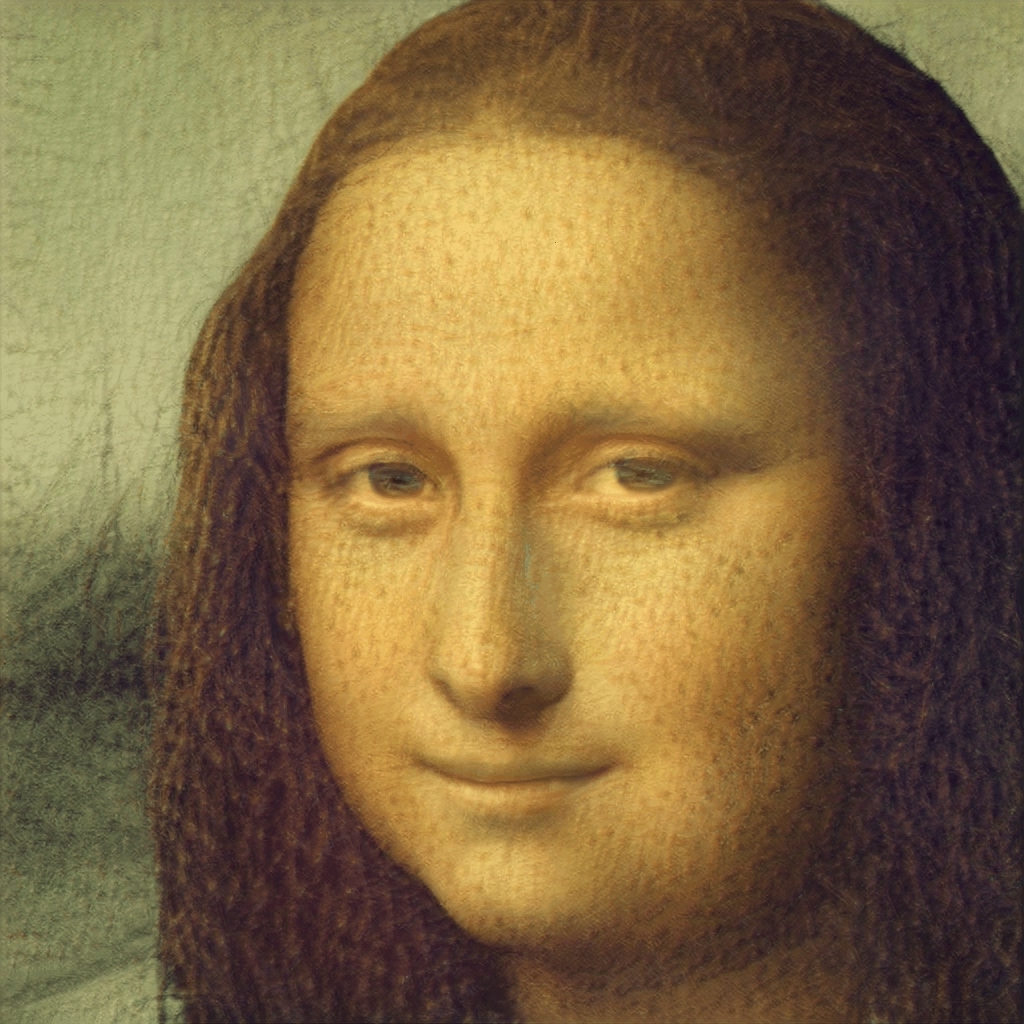}
\includegraphics[width=0.18\textwidth]{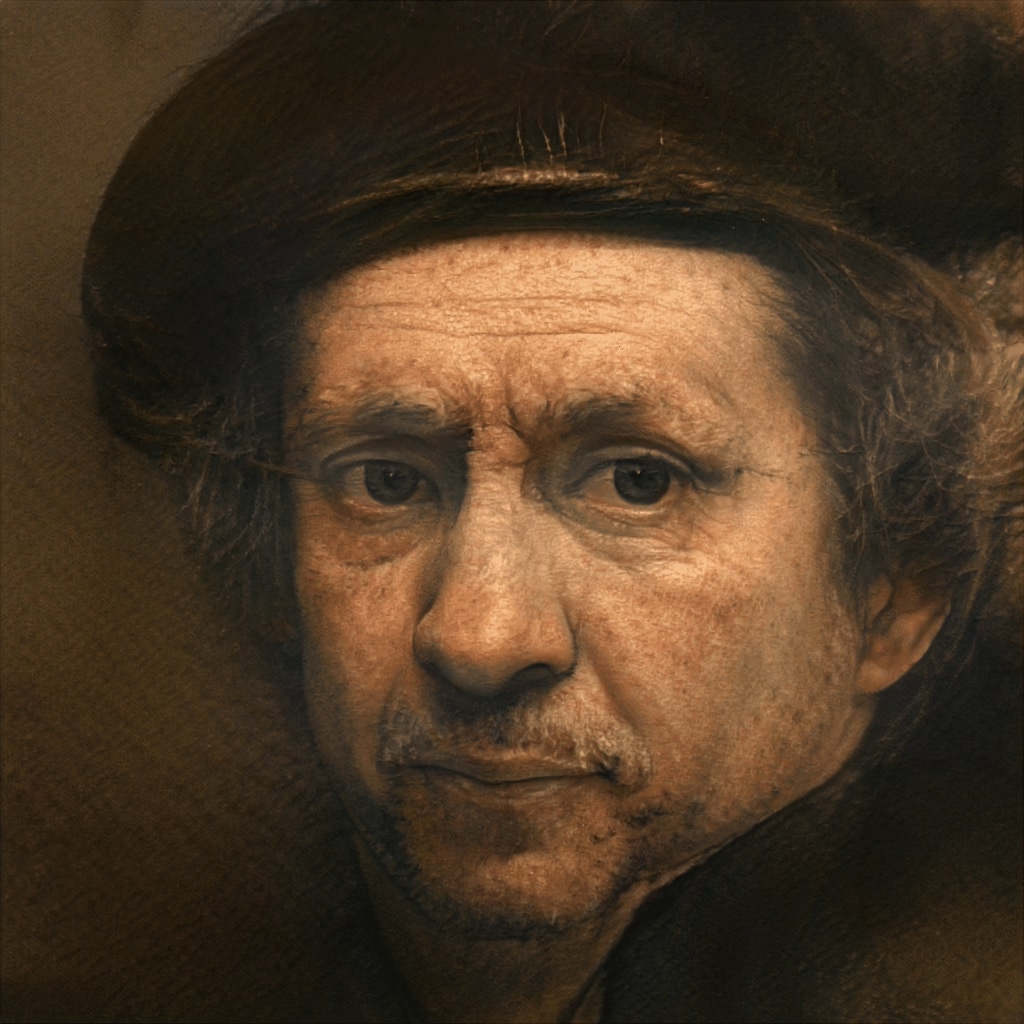}
\end{tabular}
\vspace{-0.8em}
\caption{Original}
\end{subfigure}
\begin{subfigure}{\textwidth}
\centering
\begin{tabular}{ccccc}
\includegraphics[width=0.18\textwidth]{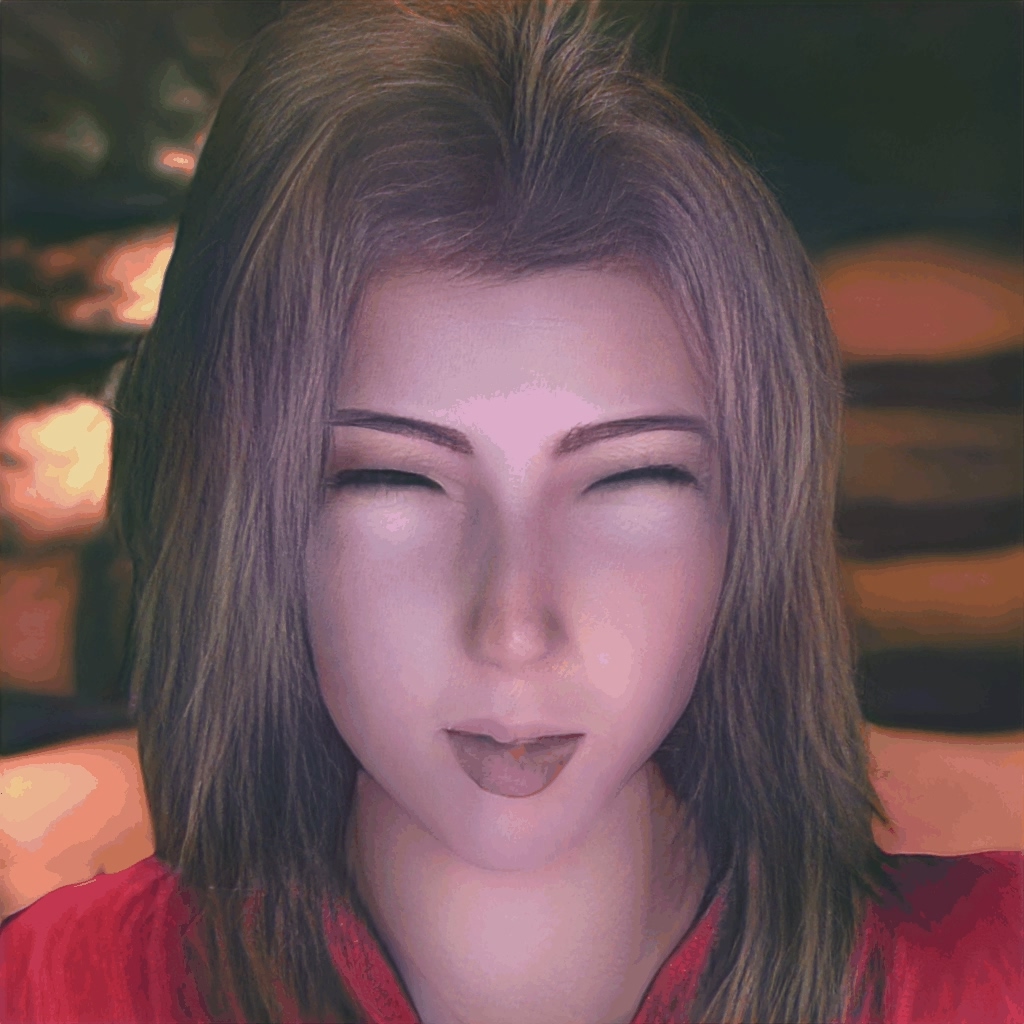}
\includegraphics[width=0.18\textwidth]{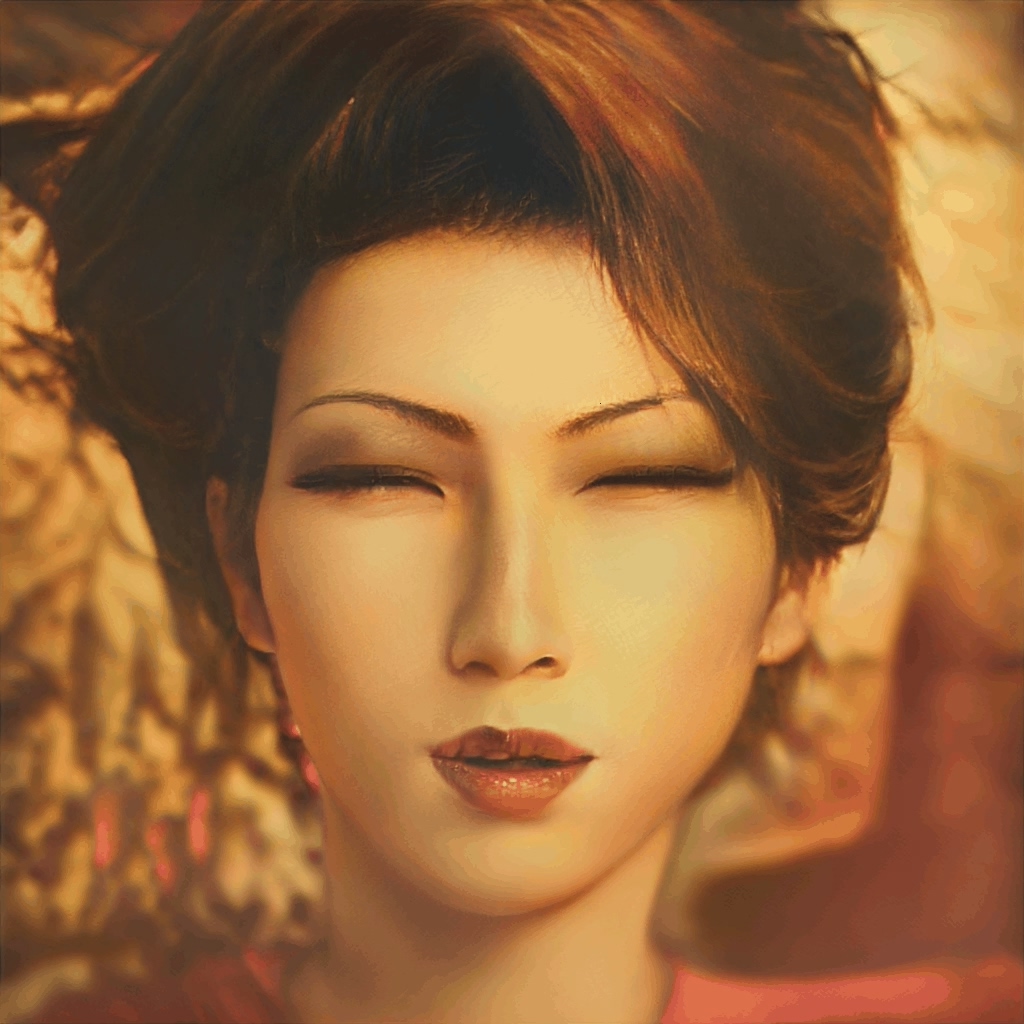}
\includegraphics[width=0.18\textwidth]{figures/david-openEyes-ours/00.jpg}
\includegraphics[width=0.18\textwidth]{figures/smile-transfer-comparison/ours/199.jpg}
\includegraphics[width=0.18\textwidth]{figures/transferred/painting/Rembrandt_6.jpg}
\end{tabular}
\vspace{-0.8em}
\caption{w/ Robust PCA}
\end{subfigure}
\begin{subfigure}{\textwidth}
\centering
\begin{tabular}{ccccc}
\includegraphics[width=0.18\textwidth]{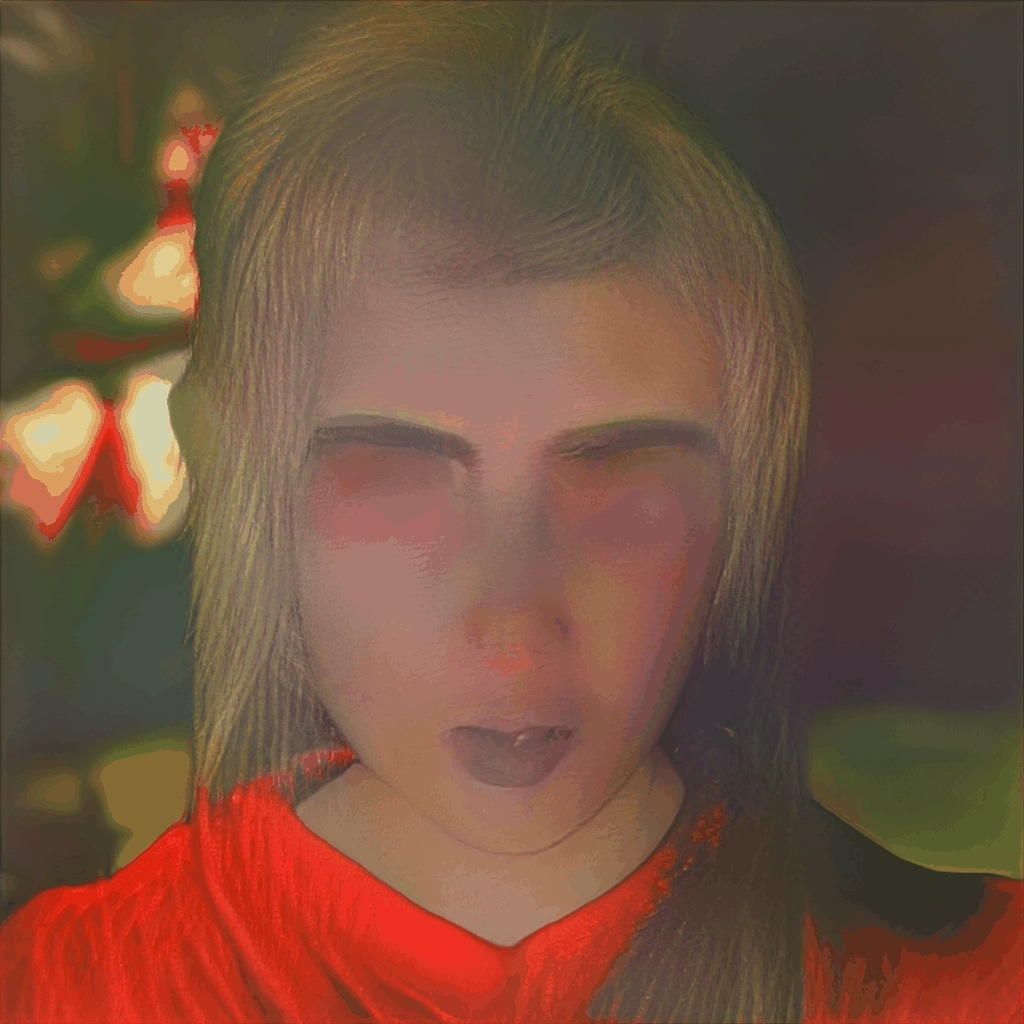}
\includegraphics[width=0.18\textwidth]{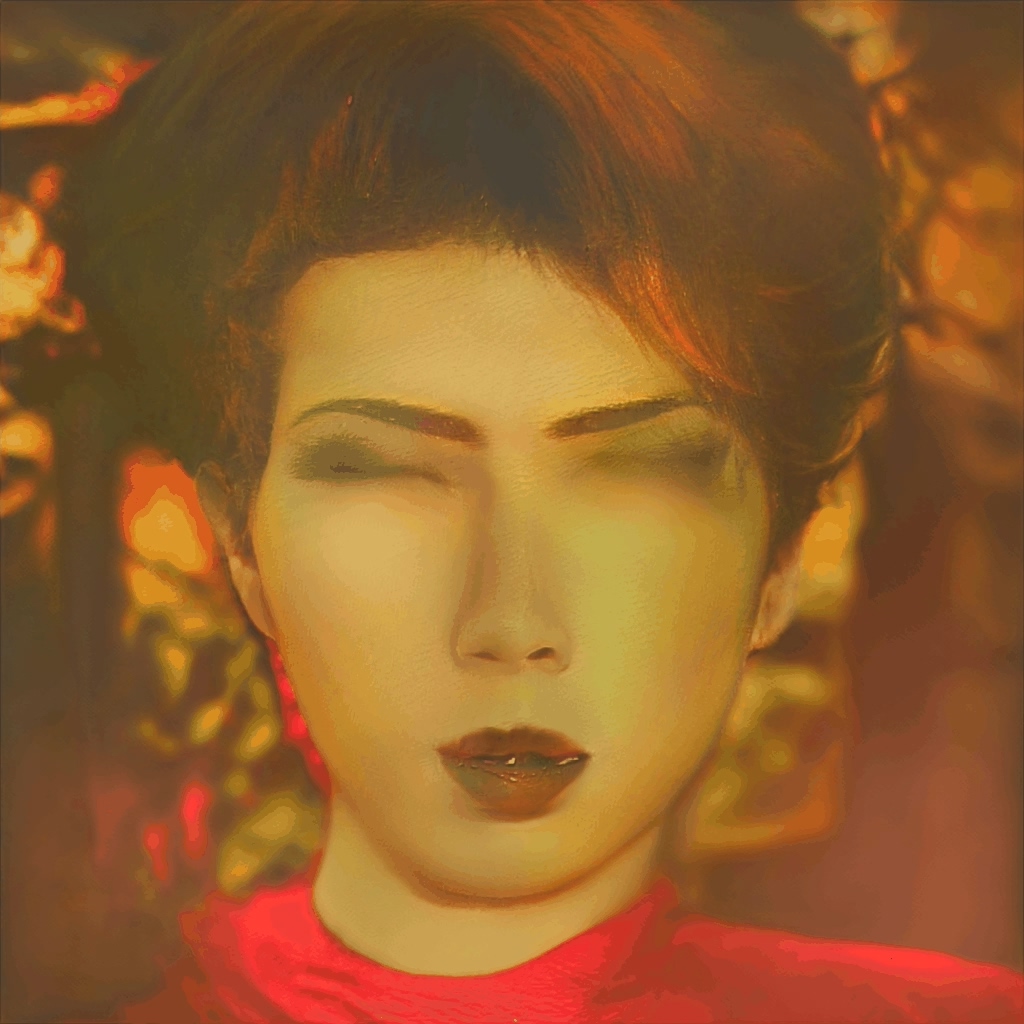}
\includegraphics[width=0.18\textwidth]{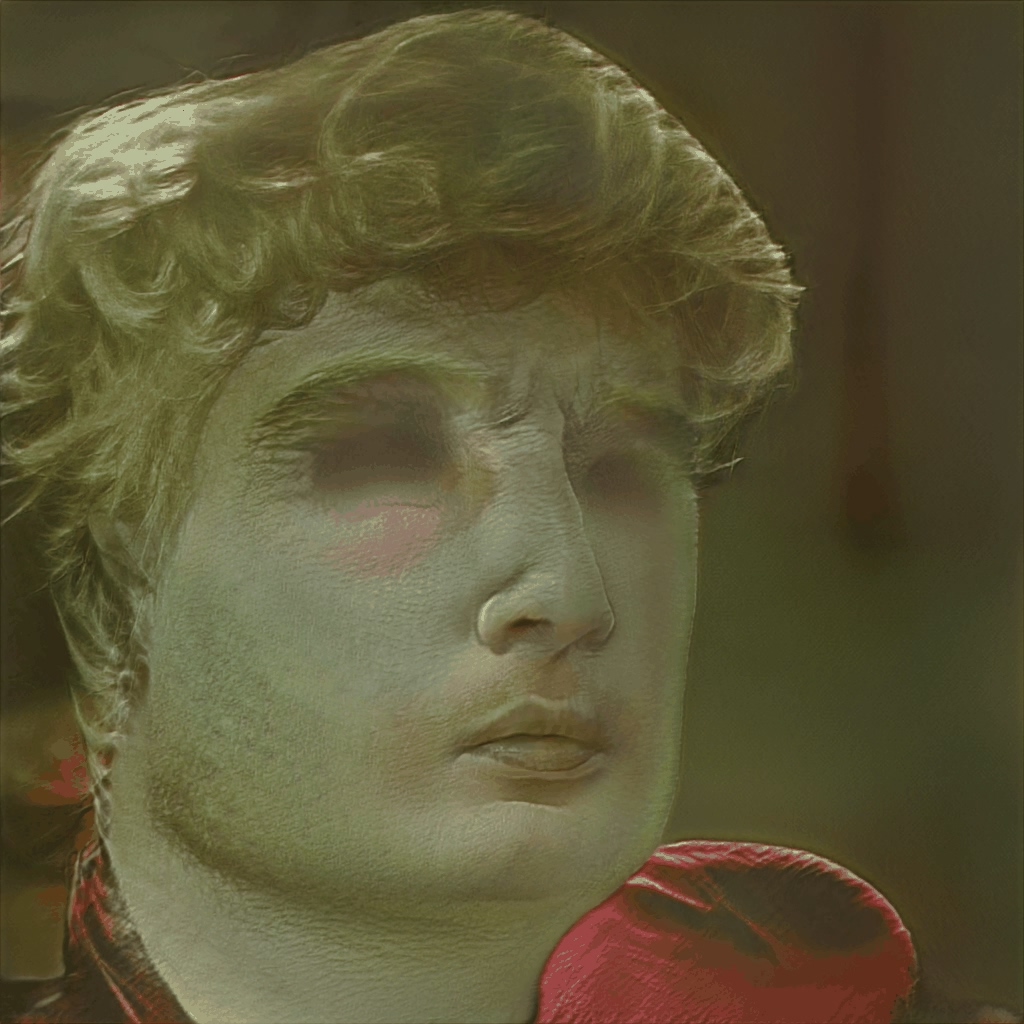}
\includegraphics[width=0.18\textwidth]{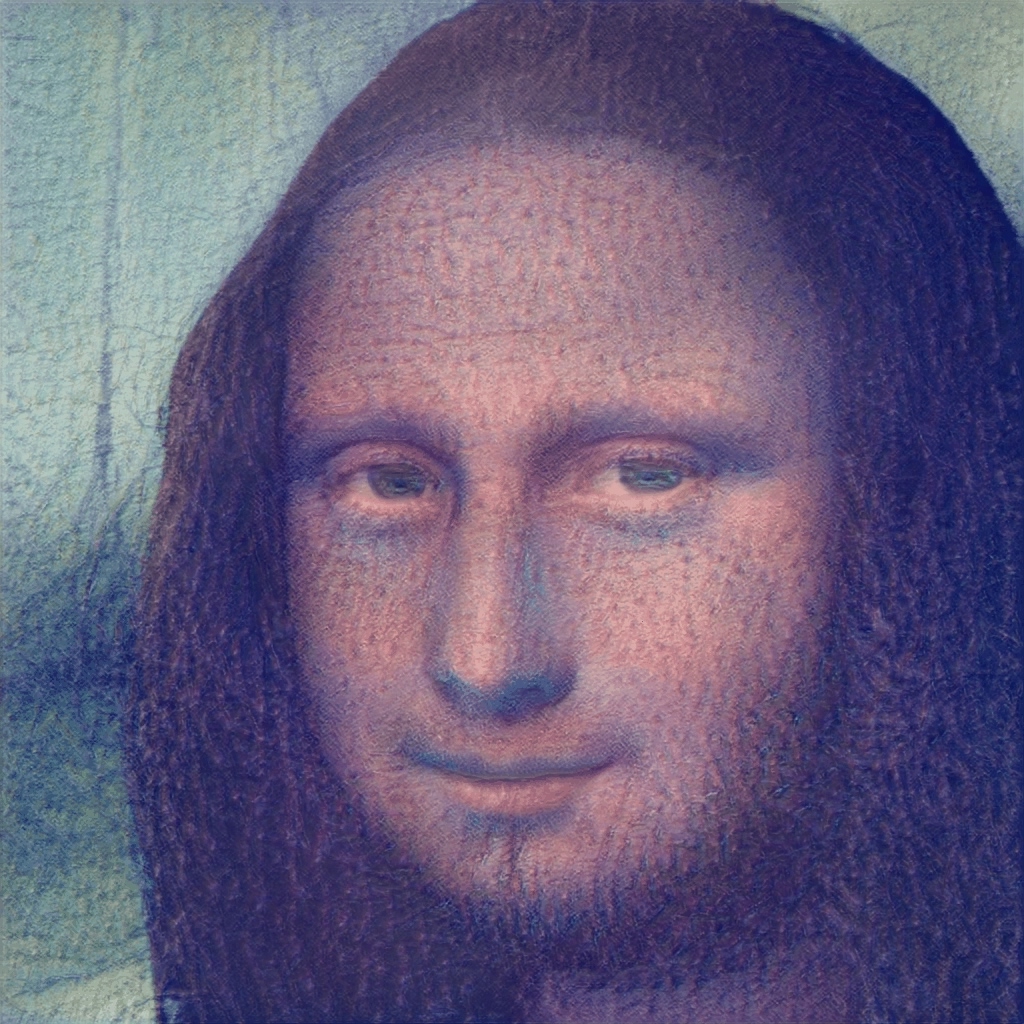}
\includegraphics[width=0.18\textwidth]{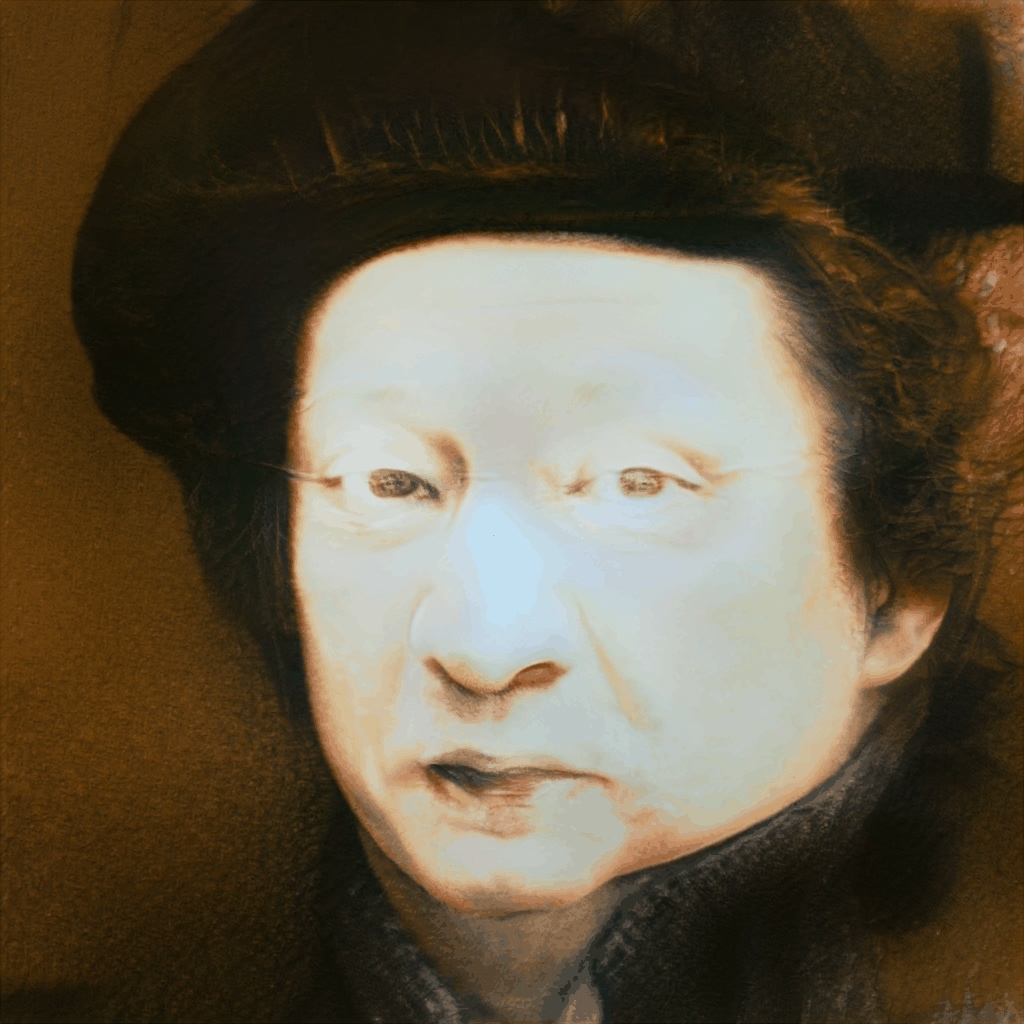}
\end{tabular}
\vspace{-0.8em}
\caption{w/o Robust PCA}
\label{fig:micromotions:rpca-c}
\end{subfigure}
\vspace{-0.5em}
\caption{\textbf{Comparison between with and without Robust PCA.} For each column, from left to right, the micromotions are ``closing eyes'' (for the first three columns), ``smiling'', ``aging face''. For conciseness, we only show the original and last frame. Best view when zoomed in.}
\label{fig:micromotions:rpca}
\end{figure*}

\begin{figure}[!htb]
\centering
\captionsetup[subfigure]{justification=centering}
\begin{subfigure}{.18\textwidth}
  \centering
  \hspace{-1.0em}\includegraphics[width=\linewidth]{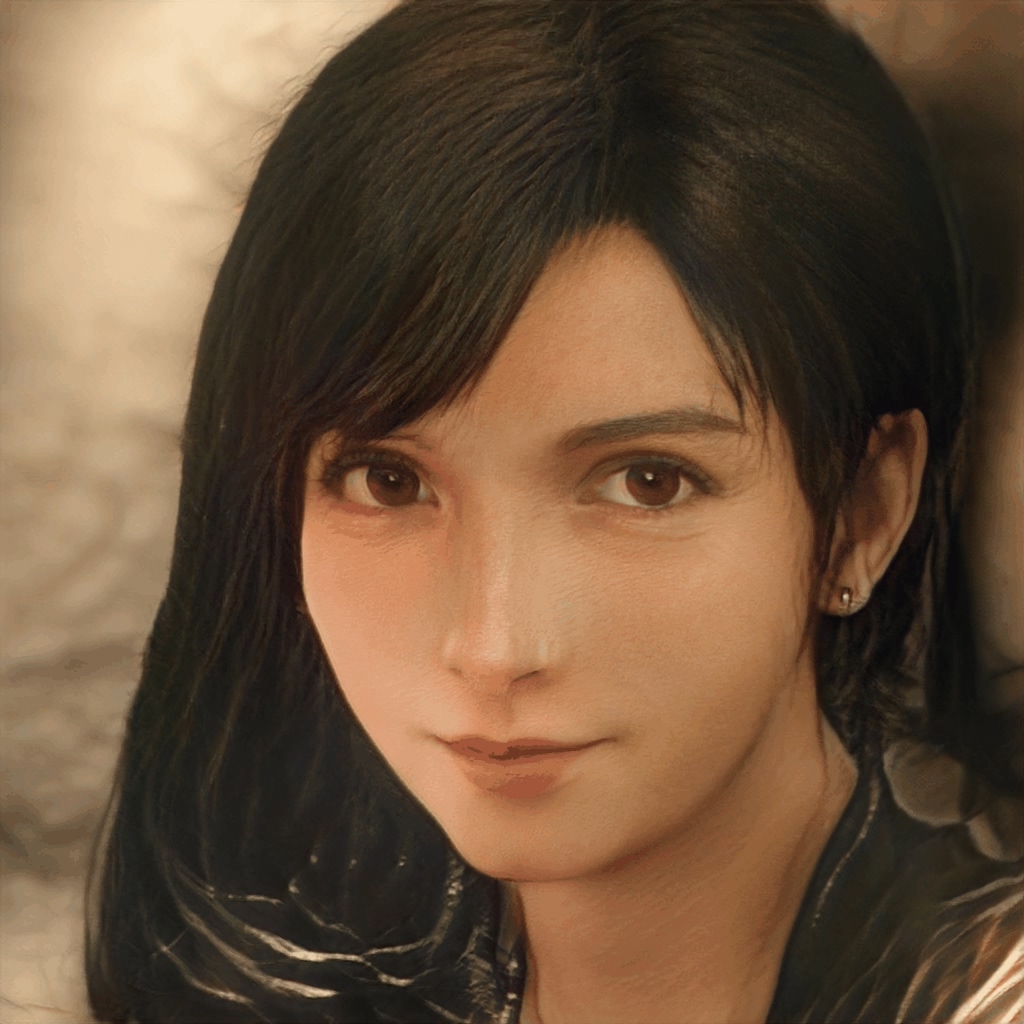}\vspace{-0.6em}
  \caption*{Original \\\hspace{0em}}
\end{subfigure}%
\begin{subfigure}{.18\textwidth}
  \centering
  \includegraphics[width=\linewidth]{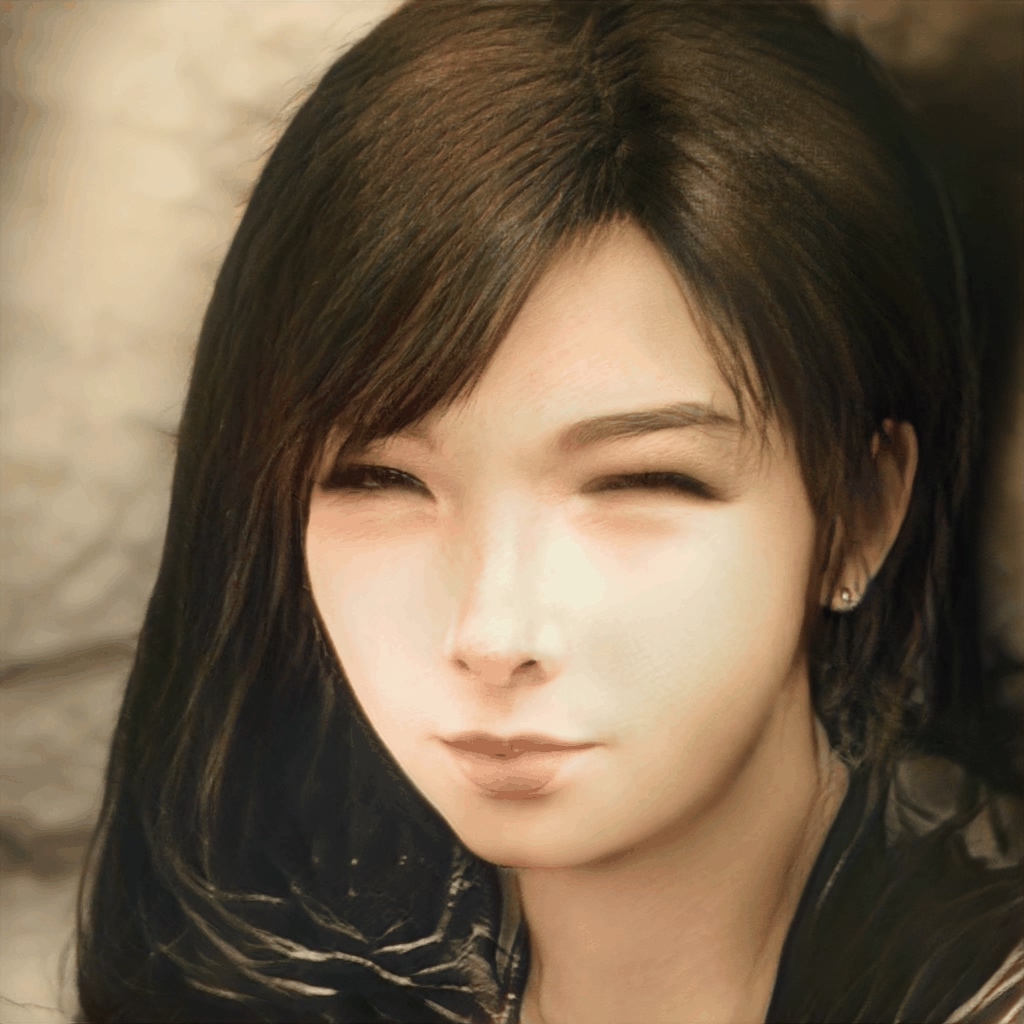}\vspace{-0.6em}
  \caption*{``eyes \{\} \\ open''}
\end{subfigure}
\begin{subfigure}{.18\textwidth}
  \centering
  \includegraphics[width=\linewidth]{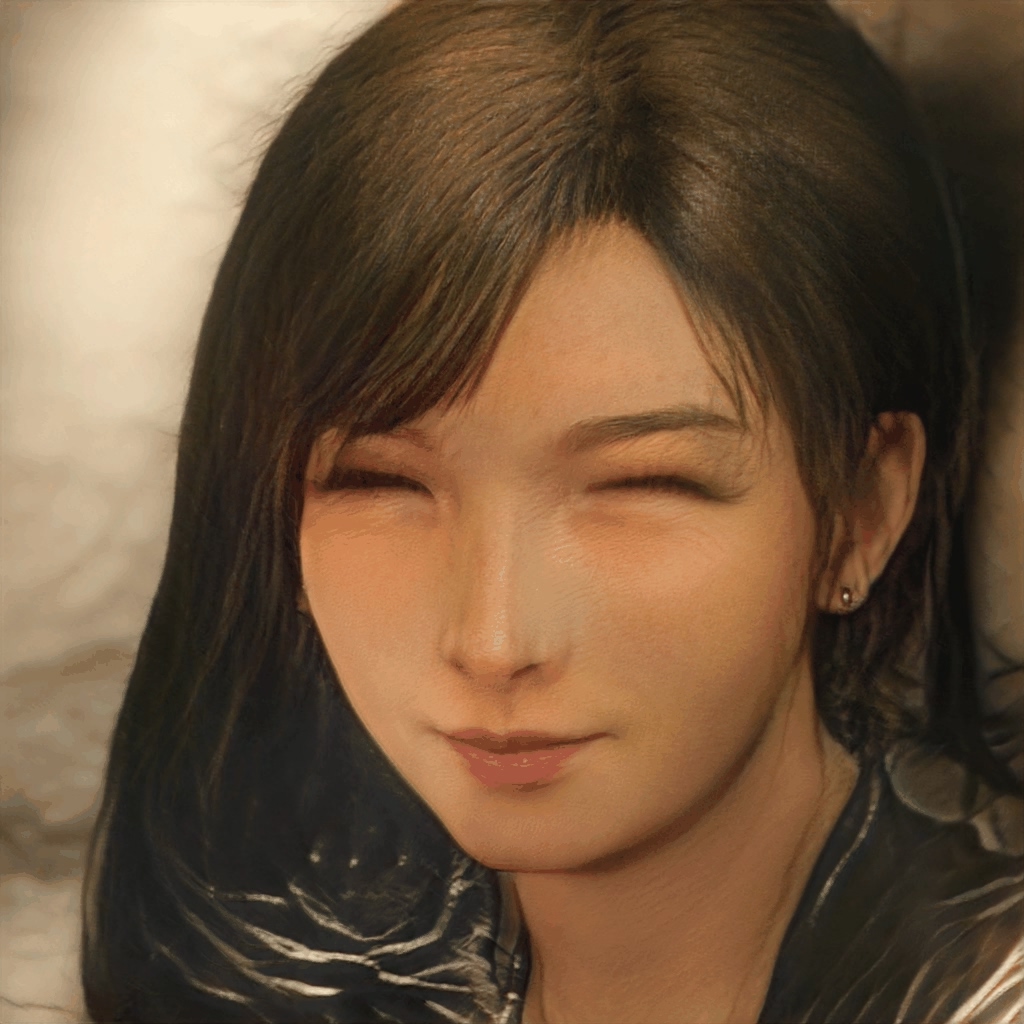}\vspace{-0.6em}
  \caption*{``a person with \\ \{\} eyes open''}
\end{subfigure}
\begin{subfigure}{.18\textwidth}
  \centering
  \includegraphics[width=\linewidth]{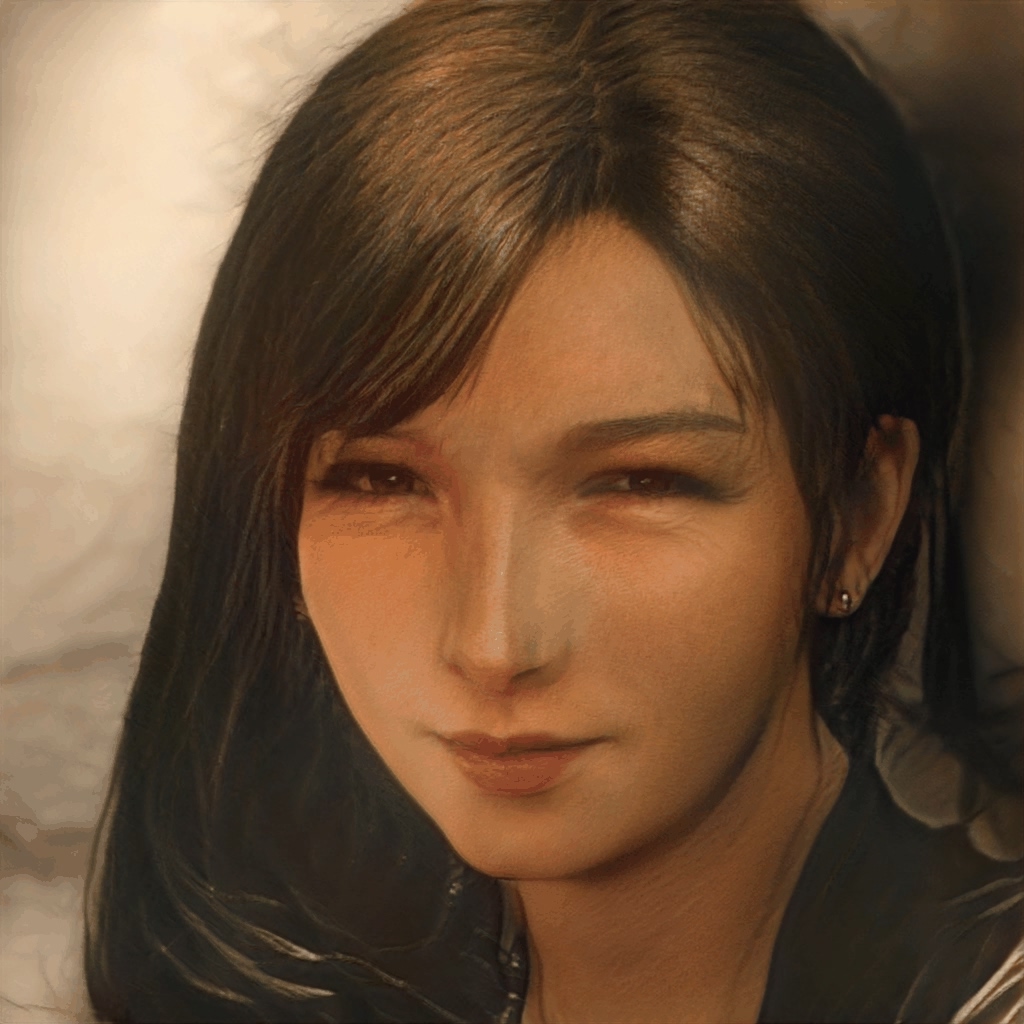}\vspace{-0.6em}
  \caption*{``a woman with \\ \{\} eyes open''}
\end{subfigure}
\begin{subfigure}{.18\textwidth}
  \centering
  \includegraphics[width=\linewidth]{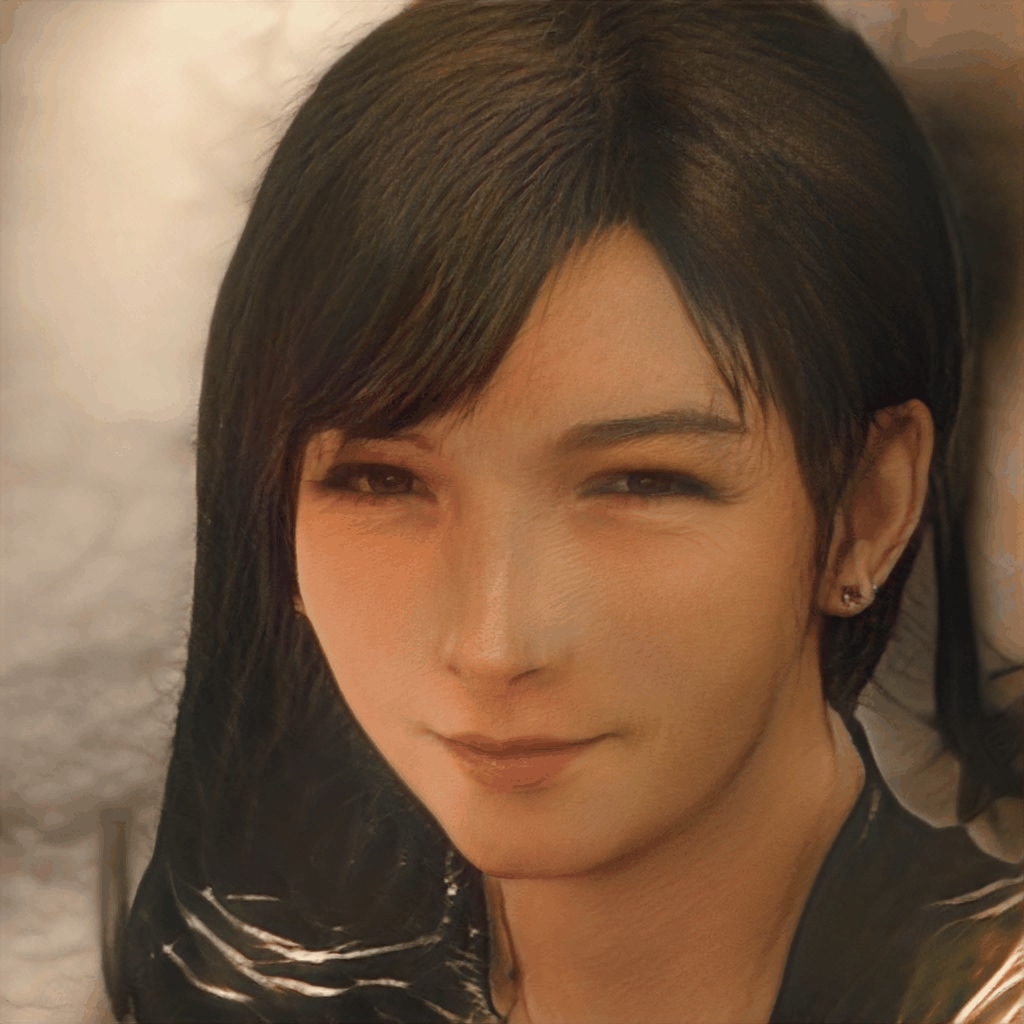}\vspace{-0.6em}
  \caption*{``a man with \\ \{\} eyes open''}
\end{subfigure}
\begin{subfigure}{.18\textwidth}
  \centering
  \hspace{-1.0em}\includegraphics[width=\linewidth]{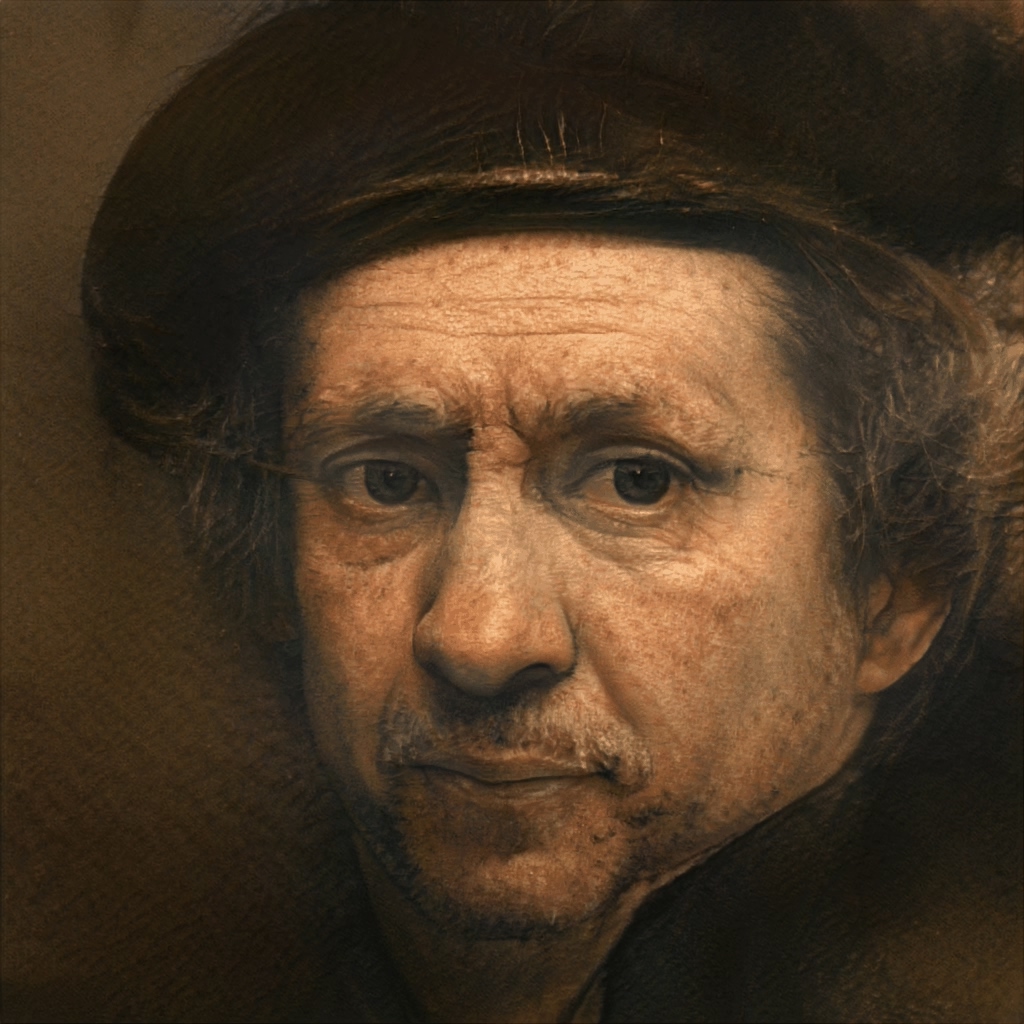}\vspace{-0.6em}
  \caption*{Original \\\hspace{0em}}
\end{subfigure}%
\begin{subfigure}{.18\textwidth}
  \centering
  \includegraphics[width=\linewidth]{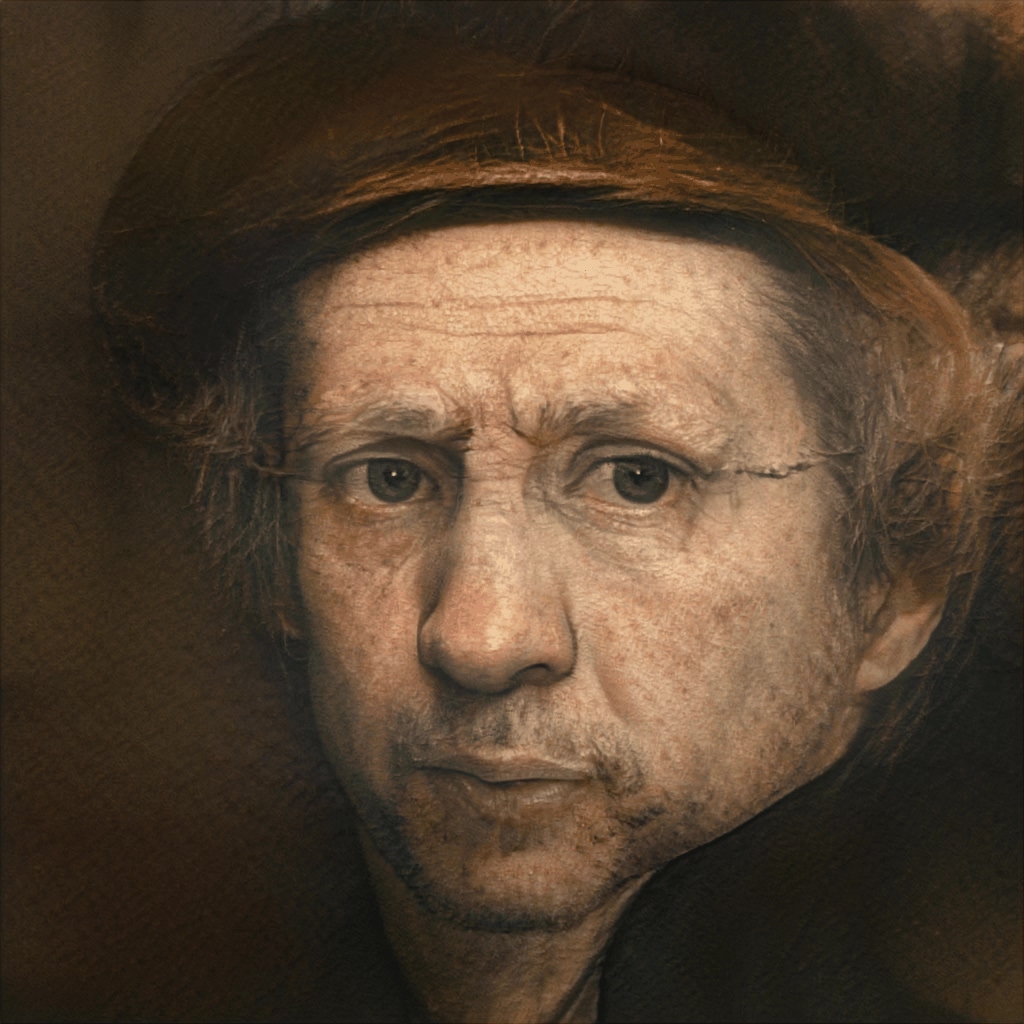}\vspace{-0.6em}
  \caption*{``a \{\} year \\old person''}
\end{subfigure}
\begin{subfigure}{.18\textwidth}
  \centering
  \includegraphics[width=\linewidth]{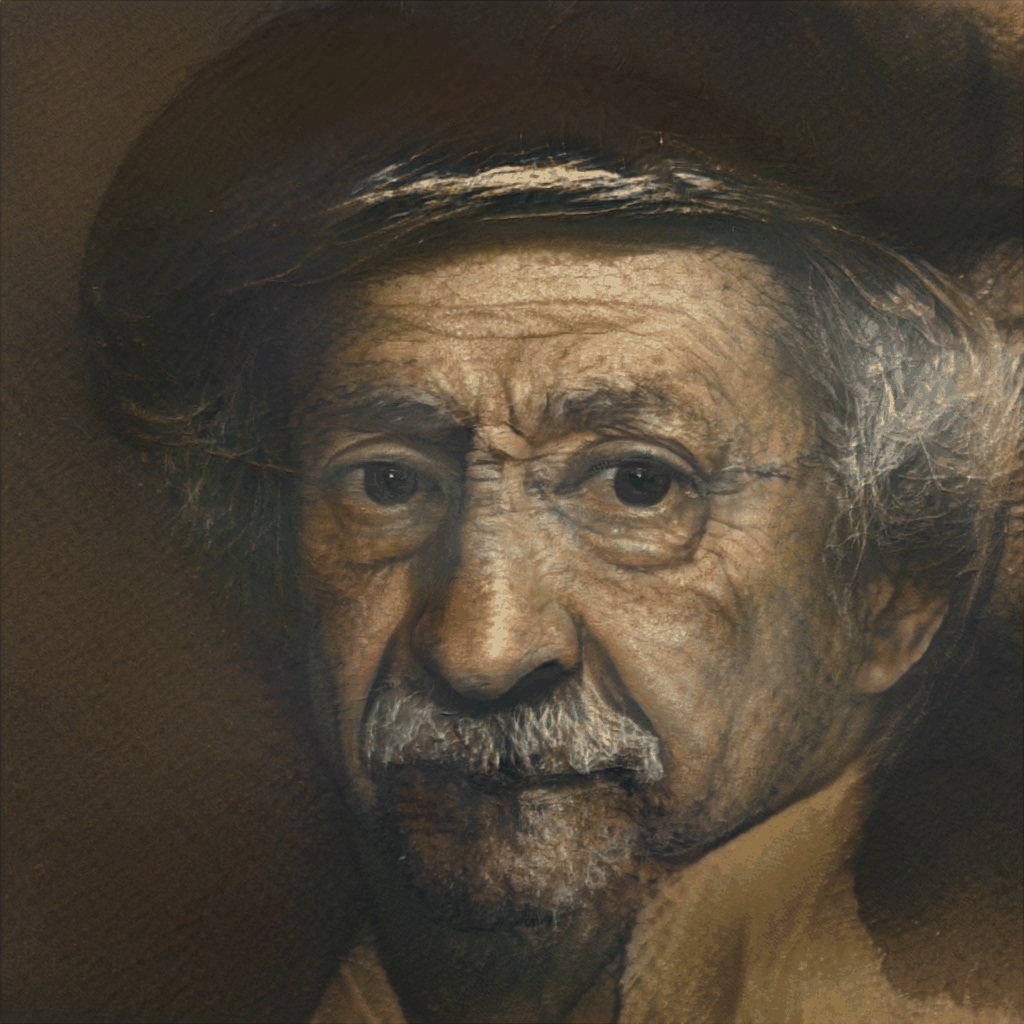}\vspace{-0.6em}
  \caption*{``a \{\} old \\person''}
\end{subfigure}
\begin{subfigure}{.18\textwidth}
  \centering
  \includegraphics[width=\linewidth]{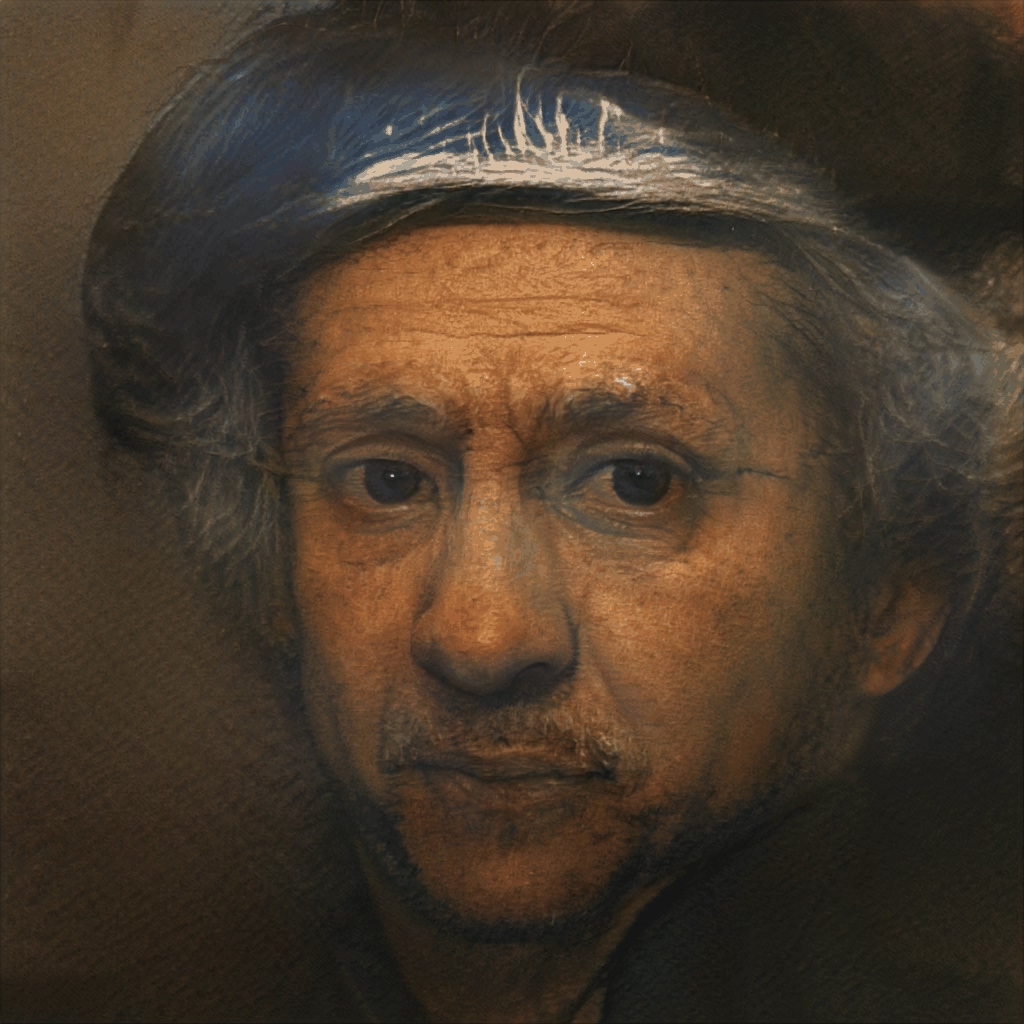}\vspace{-0.6em}
  \caption*{``a \{\} old \\woman''}
\end{subfigure}
\begin{subfigure}{.18\textwidth}
  \centering
  \includegraphics[width=\linewidth]{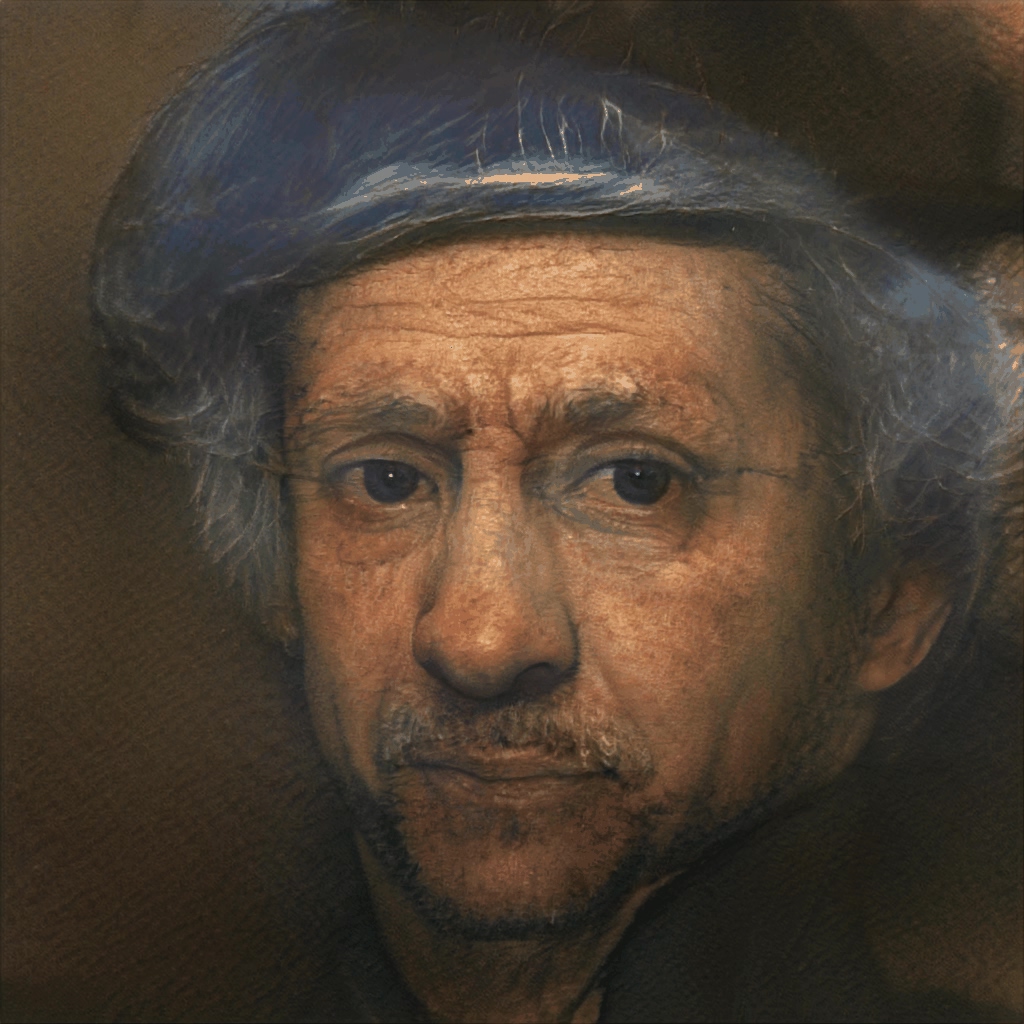}\vspace{-0.6em}
  \caption*{``a \{\} old \\man''}
\end{subfigure}
\caption{\textbf{Ablation on the choice of text template for micromotion ``opening eyes'' and ``aging face''.} For each template, we fill the wildcard ``\{\}'' using descriptive text, including \{10\%, 20\%, ..., 100\%\}, \{10, 20, ..., 60\}, and \{small, big, ...\}. For conciseness, we only show the last frame of each group; please refer to supplementary for intermediate frames. Best view when zoomed in.}
\label{fig:text-template-ablation}
\end{figure}

\subsection{Ablation Study}
\paragraph{Ablation on component decomposition in micromotion subspace}

To verify the effectiveness of the robust decomposition in our workflow, instead of doing robust PCA to decompose the low-rank micromotion space, we randomly pick two anchoring latent codes and adopt its interpolated linear space as the low-rank space. Then, we compare the qualitative results of the decoded micromotions. Results in Figure~\ref{fig:micromotions:rpca} show that synthesized videos without robust space decomposition step incur many undesired artifacts, often entangling many noisy attributes not belonging to the original and presumably mixed from other identities. Adding a robustness aware subspace decomposition, however, effectively extracts more stable and  clearly disentangled linear subspace dimensions in the presence of feature fluctuations and outliers.

\paragraph{Ablation on text templates} 
To explore the sensitivity of the micromotion subspace w.r.t the text templates, we study various text templates that describe the same micromotion.
In Figure~\ref{fig:text-template-ablation} top row, we can see that the micromotion ``closing eyes'' is agnostic to the choice of different text templates and generate similar visual results. On the other hand, In Figure~\ref{fig:text-template-ablation} bottom row, we observe the opposite where the micromotion ``face aging'' is sensitive to different text templates, which results in diverse visual patterns. This suggests the choice of text template may influence the performance of some micromotions, and a high-quality text guidance based on prompts engineering or prompts learning could be interesting future work.

\section{Conclusions} 
In this work, we analyze the latent space of StyleGAN-v2, demonstrating that although trained with static images, the StyleGAN still captures temporal micromotion representation in its feature space. We find versatile micromotions can be represented by low-dimensional subspaces of the original StyleGAN latent space, and such representations are disentangled and agnostic to the choice of identities. Based on this finding, we explore and successfully decode representative micromotion subspace by two methods: text-anchored and video-anchored reference generation, and these micromotions can be applied to arbitrary cross-domain subjects, even for the virtual figures including oil paintings, sculptures, and anime characters. Future works may study more complex motion subspace and further explore if larger-scale motion subspace is also ubiquitous, which serves as a profound step to connect discrete image manipulation with continuous video synthesis.

\bibliography{references}

\begin{thebibliography}{39}
\providecommand{\natexlab}[1]{#1}
\providecommand{\url}[1]{\texttt{#1}}
\expandafter\ifx\csname urlstyle\endcsname\relax
  \providecommand{\doi}[1]{doi: #1}\else
  \providecommand{\doi}{doi: \begingroup \urlstyle{rm}\Url}\fi

\bibitem[Abdal et~al.(2019)Abdal, Qin, and Wonka]{abdal2019image2stylegan}
R.~Abdal, Y.~Qin, and P.~Wonka.
\newblock Image2stylegan: How to embed images into the stylegan latent space?
\newblock In \emph{Proceedings of the IEEE/CVF International Conference on
  Computer Vision}, pages 4432--4441, 2019.

\bibitem[Abdal et~al.(2020)Abdal, Qin, and Wonka]{abdal2020image2stylegan++}
R.~Abdal, Y.~Qin, and P.~Wonka.
\newblock Image2stylegan++: How to edit the embedded images?
\newblock In \emph{Proceedings of the IEEE/CVF Conference on Computer Vision
  and Pattern Recognition}, pages 8296--8305, 2020.

\bibitem[Abdal et~al.(2021)Abdal, Zhu, Femiani, Mitra, and
  Wonka]{abdal2021clip2stylegan}
R.~Abdal, P.~Zhu, J.~Femiani, N.~J. Mitra, and P.~Wonka.
\newblock Clip2stylegan: Unsupervised extraction of stylegan edit directions.
\newblock \emph{arXiv preprint arXiv:2112.05219}, 2021.

\bibitem[Alaluf et~al.(2021{\natexlab{a}})Alaluf, Patashnik, and
  Cohen-Or]{alaluf2021only}
Y.~Alaluf, O.~Patashnik, and D.~Cohen-Or.
\newblock Only a matter of style: Age transformation using a style-based
  regression model.
\newblock \emph{ACM Transactions on Graphics (TOG)}, 40\penalty0 (4):\penalty0
  1--12, 2021{\natexlab{a}}.

\bibitem[Alaluf et~al.(2021{\natexlab{b}})Alaluf, Patashnik, and
  Cohen-Or]{alaluf2021restyle}
Y.~Alaluf, O.~Patashnik, and D.~Cohen-Or.
\newblock Restyle: A residual-based stylegan encoder via iterative refinement.
\newblock In \emph{Proceedings of the IEEE/CVF International Conference on
  Computer Vision}, pages 6711--6720, 2021{\natexlab{b}}.

\bibitem[Brock et~al.(2018)Brock, Donahue, and Simonyan]{brock2018large}
A.~Brock, J.~Donahue, and K.~Simonyan.
\newblock Large scale gan training for high fidelity natural image synthesis.
\newblock \emph{arXiv preprint arXiv:1809.11096}, 2018.

\bibitem[Cand{\`e}s et~al.(2011)Cand{\`e}s, Li, Ma, and
  Wright]{candes2011robust}
E.~J. Cand{\`e}s, X.~Li, Y.~Ma, and J.~Wright.
\newblock Robust principal component analysis?
\newblock \emph{Journal of the ACM (JACM)}, 58\penalty0 (3):\penalty0 1--37,
  2011.

\bibitem[Chefer et~al.(2021)Chefer, Benaim, Paiss, and Wolf]{chefer2021image}
H.~Chefer, S.~Benaim, R.~Paiss, and L.~Wolf.
\newblock Image-based clip-guided essence transfer.
\newblock \emph{arXiv preprint arXiv:2110.12427}, 2021.

\bibitem[Chong et~al.(2021)Chong, Chu, Kumar, and Forsyth]{chong2021retrieve}
M.~J. Chong, W.-S. Chu, A.~Kumar, and D.~Forsyth.
\newblock Retrieve in style: Unsupervised facial feature transfer and
  retrieval.
\newblock In \emph{Proceedings of the IEEE/CVF International Conference on
  Computer Vision}, pages 3887--3896, 2021.

\bibitem[Chu et~al.(2020)Chu, Xie, Mayer, Leal-Taix{\'e}, and
  Thuerey]{chu2020learning}
M.~Chu, Y.~Xie, J.~Mayer, L.~Leal-Taix{\'e}, and N.~Thuerey.
\newblock Learning temporal coherence via self-supervision for gan-based video
  generation.
\newblock \emph{ACM Transactions on Graphics (TOG)}, 39\penalty0 (4):\penalty0
  75--1, 2020.

\bibitem[Fox et~al.(2021)Fox, Tewari, Elgharib, and
  Theobalt]{fox2021stylevideogan}
G.~Fox, A.~Tewari, M.~Elgharib, and C.~Theobalt.
\newblock Stylevideogan: A temporal generative model using a pretrained
  stylegan.
\newblock \emph{arXiv preprint arXiv:2107.07224}, 2021.

\bibitem[Gal et~al.(2021)Gal, Patashnik, Maron, Chechik, and
  Cohen-Or]{gal2021stylegan}
R.~Gal, O.~Patashnik, H.~Maron, G.~Chechik, and D.~Cohen-Or.
\newblock Stylegan-nada: Clip-guided domain adaptation of image generators.
\newblock \emph{arXiv preprint arXiv:2108.00946}, 2021.

\bibitem[Gourier et~al.(2004)Gourier, Hall, and Crowley]{gourier2004estimating}
N.~Gourier, D.~Hall, and J.~L. Crowley.
\newblock Estimating face orientation from robust detection of salient facial
  features.
\newblock In \emph{ICPR International Workshop on Visual Observation of Deictic
  Gestures}. Citeseer, 2004.

\bibitem[Gu et~al.(2020)Gu, Shen, and Zhou]{gu2020image}
J.~Gu, Y.~Shen, and B.~Zhou.
\newblock Image processing using multi-code gan prior.
\newblock In \emph{Proceedings of the IEEE/CVF conference on computer vision
  and pattern recognition}, pages 3012--3021, 2020.

\bibitem[Hou et~al.(2022)Hou, Zhang, Liang, Shen, Lai, and
  Wan]{hou2022guidedstyle}
X.~Hou, X.~Zhang, H.~Liang, L.~Shen, Z.~Lai, and J.~Wan.
\newblock Guidedstyle: Attribute knowledge guided style manipulation for
  semantic face editing.
\newblock \emph{Neural Networks}, 145:\penalty0 209--220, 2022.

\bibitem[Karras et~al.(2017)Karras, Aila, Laine, and
  Lehtinen]{karras2017progressive}
T.~Karras, T.~Aila, S.~Laine, and J.~Lehtinen.
\newblock Progressive growing of gans for improved quality, stability, and
  variation.
\newblock \emph{arXiv preprint arXiv:1710.10196}, 2017.

\bibitem[Karras et~al.(2019)Karras, Laine, and Aila]{karras2019style}
T.~Karras, S.~Laine, and T.~Aila.
\newblock A style-based generator architecture for generative adversarial
  networks.
\newblock In \emph{Proceedings of the IEEE/CVF conference on computer vision
  and pattern recognition}, pages 4401--4410, 2019.

\bibitem[Karras et~al.(2020)Karras, Laine, Aittala, Hellsten, Lehtinen, and
  Aila]{karras2020analyzing}
T.~Karras, S.~Laine, M.~Aittala, J.~Hellsten, J.~Lehtinen, and T.~Aila.
\newblock Analyzing and improving the image quality of stylegan.
\newblock In \emph{Proceedings of the IEEE/CVF conference on computer vision
  and pattern recognition}, pages 8110--8119, 2020.

\bibitem[Karras et~al.(2021)Karras, Aittala, Laine, H{\"a}rk{\"o}nen, Hellsten,
  Lehtinen, and Aila]{karras2021alias}
T.~Karras, M.~Aittala, S.~Laine, E.~H{\"a}rk{\"o}nen, J.~Hellsten, J.~Lehtinen,
  and T.~Aila.
\newblock Alias-free generative adversarial networks.
\newblock \emph{Advances in Neural Information Processing Systems}, 34, 2021.

\bibitem[Kim et~al.(2021)Kim, Choi, Kim, Yoo, and Uh]{kim2021exploiting}
H.~Kim, Y.~Choi, J.~Kim, S.~Yoo, and Y.~Uh.
\newblock Exploiting spatial dimensions of latent in gan for real-time image
  editing.
\newblock In \emph{Proceedings of the IEEE/CVF Conference on Computer Vision
  and Pattern Recognition}, pages 852--861, 2021.

\bibitem[Lewis et~al.(2021)Lewis, Varadharajan, and
  Kemelmacher-Shlizerman]{lewis2021vogue}
K.~M. Lewis, S.~Varadharajan, and I.~Kemelmacher-Shlizerman.
\newblock Vogue: Try-on by stylegan interpolation optimization.
\newblock \emph{arXiv e-prints}, pages arXiv--2101, 2021.

\bibitem[Park et~al.(2020)Park, Zhu, Wang, Lu, Shechtman, Efros, and
  Zhang]{park2020swapping}
T.~Park, J.-Y. Zhu, O.~Wang, J.~Lu, E.~Shechtman, A.~Efros, and R.~Zhang.
\newblock Swapping autoencoder for deep image manipulation.
\newblock \emph{Advances in Neural Information Processing Systems},
  33:\penalty0 7198--7211, 2020.

\bibitem[Patashnik et~al.(2021)Patashnik, Wu, Shechtman, Cohen-Or, and
  Lischinski]{patashnik2021styleclip}
O.~Patashnik, Z.~Wu, E.~Shechtman, D.~Cohen-Or, and D.~Lischinski.
\newblock Styleclip: Text-driven manipulation of stylegan imagery.
\newblock In \emph{Proceedings of the IEEE/CVF International Conference on
  Computer Vision}, pages 2085--2094, 2021.

\bibitem[Radford et~al.(2021)Radford, Kim, Hallacy, Ramesh, Goh, Agarwal,
  Sastry, Askell, Mishkin, Clark, et~al.]{radford2021learning}
A.~Radford, J.~W. Kim, C.~Hallacy, A.~Ramesh, G.~Goh, S.~Agarwal, G.~Sastry,
  A.~Askell, P.~Mishkin, J.~Clark, et~al.
\newblock Learning transferable visual models from natural language
  supervision.
\newblock In \emph{International Conference on Machine Learning}, pages
  8748--8763. PMLR, 2021.

\bibitem[Richardson et~al.(2021)Richardson, Alaluf, Patashnik, Nitzan, Azar,
  Shapiro, and Cohen-Or]{richardson2021encoding}
E.~Richardson, Y.~Alaluf, O.~Patashnik, Y.~Nitzan, Y.~Azar, S.~Shapiro, and
  D.~Cohen-Or.
\newblock Encoding in style: a stylegan encoder for image-to-image translation.
\newblock In \emph{Proceedings of the IEEE/CVF Conference on Computer Vision
  and Pattern Recognition}, pages 2287--2296, 2021.

\bibitem[Roich et~al.(2021)Roich, Mokady, Bermano, and
  Cohen-Or]{roich2021pivotal}
D.~Roich, R.~Mokady, A.~H. Bermano, and D.~Cohen-Or.
\newblock Pivotal tuning for latent-based editing of real images.
\newblock \emph{arXiv preprint arXiv:2106.05744}, 2021.

\bibitem[Sauer et~al.(2022)Sauer, Schwarz, and Geiger]{sauer2022stylegan}
A.~Sauer, K.~Schwarz, and A.~Geiger.
\newblock Stylegan-xl: Scaling stylegan to large diverse datasets.
\newblock \emph{arXiv preprint arXiv:2202.00273}, 2022.

\bibitem[Shen et~al.(2020)Shen, Gu, Tang, and Zhou]{shen2020interpreting}
Y.~Shen, J.~Gu, X.~Tang, and B.~Zhou.
\newblock Interpreting the latent space of gans for semantic face editing.
\newblock In \emph{Proceedings of the IEEE/CVF Conference on Computer Vision
  and Pattern Recognition}, pages 9243--9252, 2020.

\bibitem[Skorokhodov et~al.(2021)Skorokhodov, Tulyakov, and
  Elhoseiny]{skorokhodov2021stylegan}
I.~Skorokhodov, S.~Tulyakov, and M.~Elhoseiny.
\newblock Stylegan-v: A continuous video generator with the price, image
  quality and perks of stylegan2.
\newblock \emph{arXiv preprint arXiv:2112.14683}, 2021.

\bibitem[Suzuki et~al.(2018)Suzuki, Koyama, Miyato, Yonetsuji, and
  Zhu]{suzuki2018spatially}
R.~Suzuki, M.~Koyama, T.~Miyato, T.~Yonetsuji, and H.~Zhu.
\newblock Spatially controllable image synthesis with internal representation
  collaging.
\newblock \emph{arXiv preprint arXiv:1811.10153}, 2018.

\bibitem[Tewari et~al.(2020{\natexlab{a}})Tewari, Elgharib, Bernard, Seidel,
  P{\'e}rez, Zollh{\"o}fer, and Theobalt]{tewari2020pie}
A.~Tewari, M.~Elgharib, F.~Bernard, H.-P. Seidel, P.~P{\'e}rez,
  M.~Zollh{\"o}fer, and C.~Theobalt.
\newblock Pie: Portrait image embedding for semantic control.
\newblock \emph{ACM Transactions on Graphics (TOG)}, 39\penalty0 (6):\penalty0
  1--14, 2020{\natexlab{a}}.

\bibitem[Tewari et~al.(2020{\natexlab{b}})Tewari, Elgharib, Bharaj, Bernard,
  Seidel, P{\'e}rez, Zollhofer, and Theobalt]{tewari2020stylerig}
A.~Tewari, M.~Elgharib, G.~Bharaj, F.~Bernard, H.-P. Seidel, P.~P{\'e}rez,
  M.~Zollhofer, and C.~Theobalt.
\newblock Stylerig: Rigging stylegan for 3d control over portrait images.
\newblock In \emph{Proceedings of the IEEE/CVF Conference on Computer Vision
  and Pattern Recognition}, pages 6142--6151, 2020{\natexlab{b}}.

\bibitem[Tov et~al.(2021)Tov, Alaluf, Nitzan, Patashnik, and
  Cohen-Or]{tov2021designing}
O.~Tov, Y.~Alaluf, Y.~Nitzan, O.~Patashnik, and D.~Cohen-Or.
\newblock Designing an encoder for stylegan image manipulation.
\newblock \emph{ACM Transactions on Graphics (TOG)}, 40\penalty0 (4):\penalty0
  1--14, 2021.

\bibitem[Wei et~al.(2021)Wei, Chen, Zhou, Liao, Tan, Yuan, Zhang, and
  Yu]{wei2021hairclip}
T.~Wei, D.~Chen, W.~Zhou, J.~Liao, Z.~Tan, L.~Yuan, W.~Zhang, and N.~Yu.
\newblock Hairclip: Design your hair by text and reference image.
\newblock \emph{arXiv preprint arXiv:2112.05142}, 2021.

\bibitem[Wright et~al.(2009)Wright, Ganesh, Rao, Peng, and
  Ma]{wright2009robust}
J.~Wright, A.~Ganesh, S.~Rao, Y.~Peng, and Y.~Ma.
\newblock Robust principal component analysis: Exact recovery of corrupted
  low-rank matrices via convex optimization.
\newblock \emph{Advances in neural information processing systems}, 22, 2009.

\bibitem[Wu et~al.(2021)Wu, Lischinski, and Shechtman]{wu2021stylespace}
Z.~Wu, D.~Lischinski, and E.~Shechtman.
\newblock Stylespace analysis: Disentangled controls for stylegan image
  generation.
\newblock In \emph{Proceedings of the IEEE/CVF Conference on Computer Vision
  and Pattern Recognition}, pages 12863--12872, 2021.

\bibitem[Zhang and Pollett(2021)]{zhang2021facial}
L.~Zhang and C.~Pollett.
\newblock Facial expression video synthesis from the stylegan latent space.
\newblock In \emph{Thirteenth International Conference on Digital Image
  Processing (ICDIP 2021)}, volume 11878, page 118781M. International Society
  for Optics and Photonics, 2021.

\bibitem[Zhu et~al.(2020)Zhu, Abdal, Qin, Femiani, and Wonka]{zhu2020improved}
P.~Zhu, R.~Abdal, Y.~Qin, J.~Femiani, and P.~Wonka.
\newblock Improved stylegan embedding: Where are the good latents?
\newblock \emph{arXiv preprint arXiv:2012.09036}, 2020.

\bibitem[Zhu et~al.(2021)Zhu, Abdal, Femiani, and Wonka]{zhu2021barbershop}
P.~Zhu, R.~Abdal, J.~Femiani, and P.~Wonka.
\newblock Barbershop: Gan-based image compositing using segmentation masks.
\newblock \emph{arXiv preprint arXiv:2106.01505}, 2021.

\end{thebibliography}
\bibliographystyle{abbrvnat}

\end{document}